\documentclass[runningheads]{llncs}
\usepackage{graphicx}
\usepackage{amsmath,amssymb} % define this before the line numbering.
\usepackage{color,xcolor}
\begin{document}
\pagestyle{headings}
\mainmatter
\def\ECCV16SubNumber{6007}  % Insert your submission number here

\title{Table Structure Recognition using Top-Down and Bottom-Up Cues} % Replace with your title

\titlerunning{Table Structure Recognition using Top-Down and Bottom-Up Cues}
\author{Sachin Raja \and Ajoy Mondal \and C. V. Jawahar}
\authorrunning{Raja et al.}
\institute{Center for Visual Information Technology,\\ International Institute of Information Technology, Hyderabad, India \\
\email{sachinraja13@gmail.com, \{ajoy.mondal,jawahar\}@iiit.ac.in}
}

\maketitle

\begin{abstract}

Tables are information-rich structured objects in document images. While significant work has been done in localizing tables as graphic objects in document images, only limited attempts exist on table structure recognition. Most existing literature on structure recognition depends on extraction of meta-features from the {\sc pdf} document or on the optical character recognition ({\sc ocr}) models to extract low-level layout features from the image. However, these methods fail to generalize well because of the absence of meta-features or errors made by the {\sc ocr} when there is a significant variance in table layouts and text organization. In our work, we focus on tables that have complex structures, dense content, and varying layouts with no dependency on meta-features and/or {\sc ocr}.

We present an approach for table structure recognition that combines cell detection and interaction modules to localize the cells and predict their row and column associations with other detected cells. We incorporate structural constraints as additional differential components to the loss function for cell detection. We empirically validate our method on the publicly available real-world datasets - {\sc icdar}-2013, {\sc icdar}-2019 (c{\sc td}a{\sc r}) archival, {\sc unlv}, {\sc s}ci{\sc tsr}, {\sc s}ci{\sc tsr-comp}, {\sc t}able{\sc b}ank, and {\sc p}ub{\sc t}ab{\sc n}et. Our attempt opens up a new direction for table structure recognition by combining top-down (table cells detection) and bottom-up (structure recognition) cues in visually understanding the tables.

\end{abstract}

%%%%%%%%%%%%%%%%%%%%%%%%%%%%%%%%%%%%%%%%%%%
%%image for introduction  
\begin{figure}[ht!]
\begin{center}
\fbox{
\includegraphics[width=0.95\linewidth,height=0.35\linewidth]{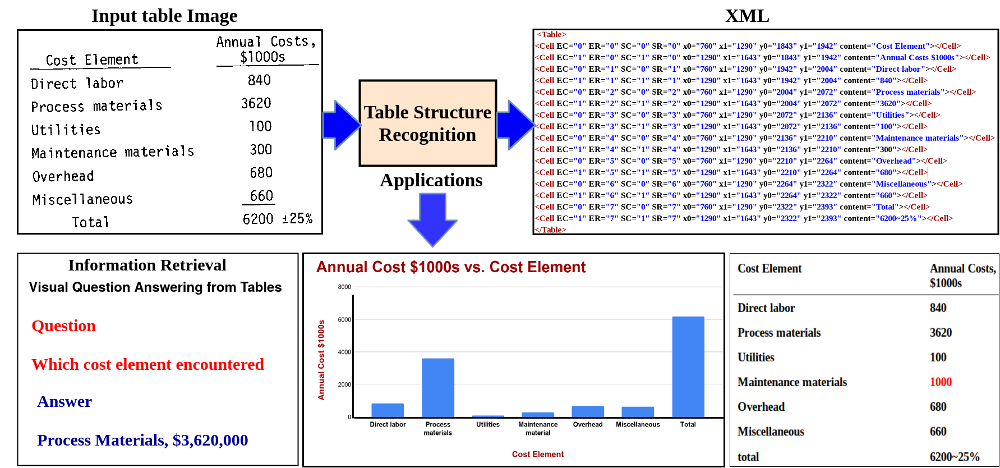}}
\end{center}
\caption{The figure depicts the problem of recognizing table structure from it's image. This opens up many applications including information retrieval, graphical representation and digitizing for editing.} \label{fig_introduction}
\end{figure}

\section{Introduction} \label{introduction}

Deep neural networks have shown promising results in understanding document layouts~\cite{yang2017learning,augusto2017fast,icdar_2017}. However, more needs to be done for structural and semantic understanding. Among these, the problem of table structure recognition has been of high interest in the community~\cite{hu1999medium,wang2004table,nishida2017understanding,schreiber2017deepdesrt,bao2018table,qasim2019rethinking,table_splitting,li2019tablebank,paliwal2019tablenet,zhong2019image,chi2019complicated,Khan_2019,Siddiqui2019Rethinking,xue2019res2tim,gobel2013icdar,gao2019icdar,ajoy2020_das}. Table structure recognition refers to representation of a table in a machine-readable format, where its layout is encoded according to a pre-defined standard~\cite{table_splitting,li2019tablebank,paliwal2019tablenet,zhong2019image,chi2019complicated,xue2019res2tim}. It can be represented in the form of either physical~\cite{table_splitting,paliwal2019tablenet,chi2019complicated,xue2019res2tim} or logical formats~\cite{li2019tablebank,zhong2019image}. While logical structure contains every cells' row and column spanning information, physical structure additionally contains bounding box coordinates. Table structure recognition is a precursor to contextual table understanding, which has a myriad of applications in business document analysis, information retrieval, visualization, and human-document interactions, as motivated in Figure~\ref{fig_introduction}. 

\begin{figure*}[t]
\begin{center}
\fbox{
\includegraphics[width=0.95\linewidth,height=0.32\linewidth]{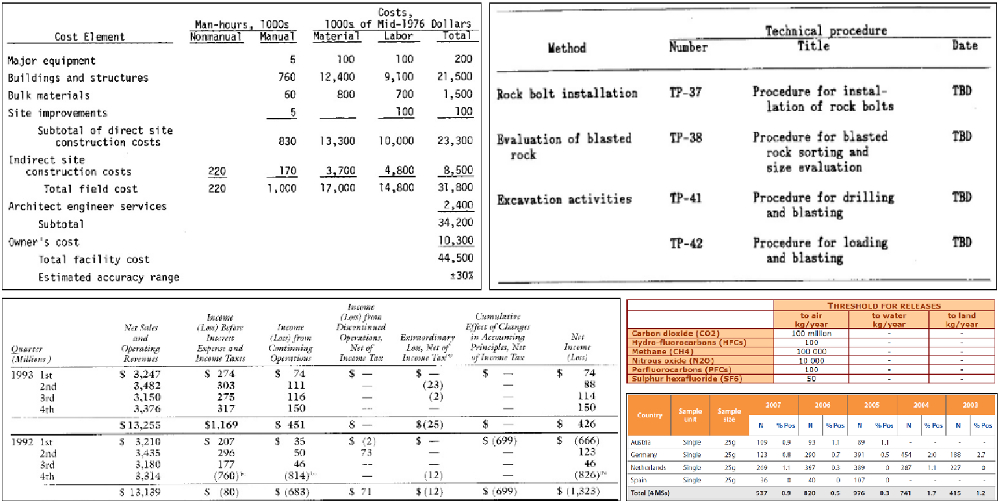}}
\end{center}
\caption{Examples of complex table images from {\sc unlv} and {\sc icdar}-2013 datasets. Complex tables are ones which contain partial or no ruling lines, multi-row/column spanning cells, multi-line content, many empty dense cells.} \label{fig_complex_table}
\end{figure*}

Table structure recognition is a challenging problem due to complex structures and high variability in table layouts~\cite{hu1999medium,wang2004table,nishida2017understanding,schreiber2017deepdesrt,bao2018table,qasim2019rethinking,table_splitting,li2019tablebank,paliwal2019tablenet,zhong2019image,chi2019complicated,Khan_2019,Siddiqui2019Rethinking,xue2019res2tim}. Early attempts in this space are dependent on extraction of hand-crafted features and meta-data extracted from the {\sc pdf}s on top of heuristic/rule-based algorithms~\cite{itonori1993table,green1995recognition,kieninger1998table,tupaj1996extracting} to locate tables and understanding tables by predicting/recognizing structures. These methods, however, fail to extend to scanned documents as they rely on meta-data information contained in the {\sc pdf}s. They also make strong assumptions about the structure of the tables. Some of these methods are also dependent on textual information analysis which make them domain dependent. While textual features are useful, visual analysis becomes imperative for analysis of complex page objects. Inconsistency of size and density of tables, presence and location of table cell borders, variation in table cells' shapes and sizes, table cells spanning multiple rows and/or columns and multi-line content are some challenges (refer
Figure~\ref{fig_complex_table} for some examples) that need to be addressed to solve the problem using visual cues~\cite{hu1999medium,wang2004table,itonori1993table,green1995recognition,kieninger1998table,tupaj1996extracting}. 

%%%%%%%%%%%%%%%%%%%%%%%%%%%%%%%%%%%%%%%%%%%%%%%%%%%%%%
%figure for complex table 

We pose the table structure recognition problem as the generation of {\sc xml} containing table's physical structure in terms of bounding boxes along with spanning information and, additionally, digitized content for every cell (see Figure~\ref{fig_introduction}). Since our method aims to predict this table structure given the table image only (without using any meta-information), we employ a two-step process --- (a) \textit{top-down:} where we decompose the table image into fundamental table objects, which are table cells using a cell detection network and (b) \textit{bottom-up:} where we re-build the entire table as a collection of all the table cells localized from the top-down process, along with their row and column associations with every other cell. We represent row and column associations of table cells using row and column adjacency matrices.

Though table detection has observed significant success ~\cite{li2019tablebank,gilani2017table,dong2019tablesense,kavasidis2019saliency,ranajit_2019}, detection of table cells remains a challenging problem. This is because of (i) large variation in sizes and aspect ratios of different cells present in the same table, (ii) cells' inherent alignment despite high variance in text amount and text justification, (iii) lack of linguistic context in cells' content, (iv) presence of empty cells and (v) presence of cells with multi-line content. To overcome these challenges, we introduce a novel loss function that models the inherent alignment of cells in the cell detection network; and a graph-based problem formulation to build associations between the detected cells. Moreover, as detection of cells and building associations between them depend highly on one another, we present a novel end-to-end trainable architecture, termed as {\sc t}ab{\sc s}truct-{\sc n}et, for cell detection and structure recognition. We evaluate our model for physical structure recognition on benchmark datasets: {\sc s}ci{\sc tsr}~\cite{chi2019complicated}, {\sc s}ci{\sc tsr-comp}~\cite{chi2019complicated}, {\sc icdar}-2013 table recognition~\cite{gobel2013icdar}, {\sc icdar}-2019 (c{\sc td}a{\sc r}) archival~\cite{gao2019icdar}, and {\sc unlv}~\cite{shahab2010open}. Further, we extend the comparative analysis of the proposed work for logical structure recognition on {\sc t}able{\sc b}ank~\cite{li2019tablebank} dataset. Our method sets up a new direction for table structure recognition as a collaboration of cell detection, establishing an association between localized cells and, additionally, cells' content extraction.

Our main contributions can be summarised as follows:
\begin{itemize}
\item We demonstrate how the top-down (cell detection) and bottom-up (structure recognition) cues can be combined visually to recognize table structures in document images.
\item We present an end-to-end trainable network, termed as {\sc t}ab{\sc s}truct-{\sc n}et for training cell detection and structure recognition networks in a joint manner.
\item We formulate a novel loss function (i.e., alignment loss) to incorporate structural constraints between every pair of table cells and modify Feature Pyramid Network ({\sc fpn}) to capture better low-level and long-range features for cell detection.
\item We enhance the visual features representation for structure recognition (built on top of model~\cite{qasim2019rethinking}) through {\sc lstm}.
\item We unify results from previously published methods on table structure recognition for a thorough comparison study.
\end{itemize}

%%%%%%%%%%%%%%%%%%%%%%%%%%%%%%%%%%%%%%%%%%%%%%%%
%figure for block diagram of proposed approach
\begin{figure*}[t]
\begin{center}
\fbox{
\includegraphics[width=1.0\linewidth, height=0.15\linewidth]{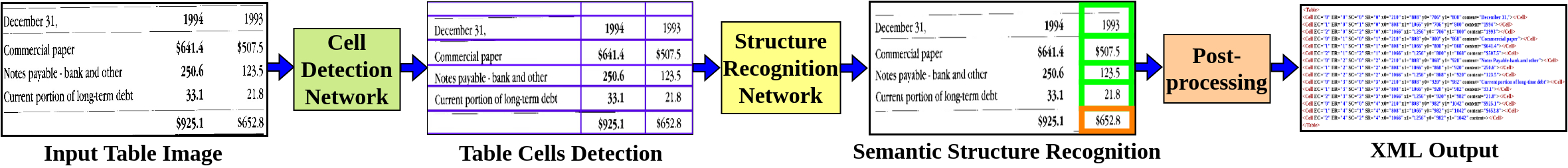}}
\end{center}
\caption{Block diagram of our approach. Table detection is a precursor to table structure recognition and our method assumes that table is already localized from the input document image. The end-to-end architecture predicts cell bounding boxes and their associations jointly. From the outputs of cell detection and association predictions, {\sc xml} is generated using a post-processing heuristic.}
\label{fig_block}
\end{figure*}

\section{Related Work} \label{related_work}

In the space of document images, researchers have been working on understanding equations~\cite{zanibbi2002recognizing,zhang2018multi}, figures~\cite{siegel2016figureseer,deepchart} and tables~\cite{nishida2017understanding,schreiber2017deepdesrt,bao2018table,qasim2019rethinking,table_splitting,li2019tablebank,paliwal2019tablenet,zhong2019image,chi2019complicated,Khan_2019,Siddiqui2019Rethinking,xue2019res2tim}. Diverse table layouts, tables with many empty cells and multi-row/column spanning cells are some challenges that make table structure recognition difficult. 
Research in the domain of table understanding through its structure recognition from document images dated back to the early 1990s when algorithms based on heuristics were proposed~\cite{itonori1993table,green1995recognition,kieninger1998table,tupaj1996extracting,harit2012table,gatos2005automatic,ohta2019cell}. These methods were primarily dependent on hand-crafted features and heuristics (horizontal and vertical ruling lines, spacing and geometric analysis). To avoid heuristics, Wang et al.~\cite{wang2004table} proposed a method for table structure analysis using optimization methods similar to the {\sc x}-{\sc y} cut algorithm. Another technique based on column segmentation, header detection, and row segmentation to identify the table structure was proposed by Hu et al.~\cite{hu1999medium}. These methods make strong assumptions about table layouts for a domain agnostic algorithm.

Many cognitive methods \cite{nishida2017understanding,schreiber2017deepdesrt,bao2018table,qasim2019rethinking,table_splitting,li2019tablebank,paliwal2019tablenet,chi2019complicated,Khan_2019,Siddiqui2019Rethinking,deng2019challenges,adiga2019table,table_invoice,holevcek2019line,deng2019table2vec,le2019extracting,sage2019recurrent} have also been presented to understand table structures as they are robust to the input type (whether being scanned images or native digital). These also do not make any assumptions about the layouts, are data-driven, and are easy to fine-tune across different domains. Minghao et al.~\cite{li2019tablebank} proposed one class of deep learning methods to directly predict an {\sc xml} from the table image using the image-to-markup model. Though this method worked well for small tables, it was not robust enough to dense and complex tables. Another set of methods is invoice specific table extraction ~\cite{table_invoice,holevcek2019line}, which were not competent for a more generic use-cases. To overcome this challenge, a combination of heuristics and cognitive methods has also been presented in~\cite{paliwal2019tablenet}. Chris et al.~\cite{table_splitting} presented another interesting deep model, called~{\sc splerge}, which is based on the fundamental idea of first splitting the table into sub-cells, and then merging semantically connected sub-cells to preserve the complete table structure. Though this algorithm showed considerable improvements over earlier methods, it was still not robust to skew present in the table images. Another interesting direction was presented by Vine et al.~\cite{le2019extracting}, where they used conditional generative adversarial networks to obtain table skeleton and then fit a latent table structure into the skeleton using a genetic algorithm. Khan et al.~\cite{Khan_2019}, through their {\sc gru} based sequential models, showed improvements over several {\sc cnn} based methods for table structure extraction. Recently, many works have preferred a graph-based formulation of the problem as the graph is inherently an ideal data structure to model structural associativity. Qasim et al.~\cite{qasim2019rethinking} proposed a solution where they used graph neural networks to model table-level associativity between words. The authors validate their method on synthetic table images. Chi et al.~\cite{chi2019complicated} proposed another graph-based problem formulation and solution using a graph attention mechanism. While these methods made significant progress towards understanding complex structured tables, they made certain assumptions like availability of accurate word bounding boxes, accurate document text, etc. as additional inputs~\cite{nishida2017understanding,qasim2019rethinking,chi2019complicated}. Our method does not make any such assumptions. We use the table image as the input and produce {\sc xml} output without any other information. We demonstrate results on complex tables present in {\sc unlv}, {\sc icdar}-2013, {\sc icdar}-2019 c{\sc td}a{\sc r} archival, {\sc s}ci{\sc tsr}, {\sc s}ci{\sc tsr-comp} {\sc t}able{\sc b}ank, and {\sc p}ub{\sc t}ab{\sc n}et datasets.

%%%%%%%%%%%%%%%%%%%%%%%%%%%%%%%%%%%%%%%%%%%%%%%%%%%%%%%%%%
%figure forstructure information 
\begin{figure*}[t]
\begin{center}
\fbox{
\includegraphics[width=0.95\linewidth, height=0.3\linewidth]{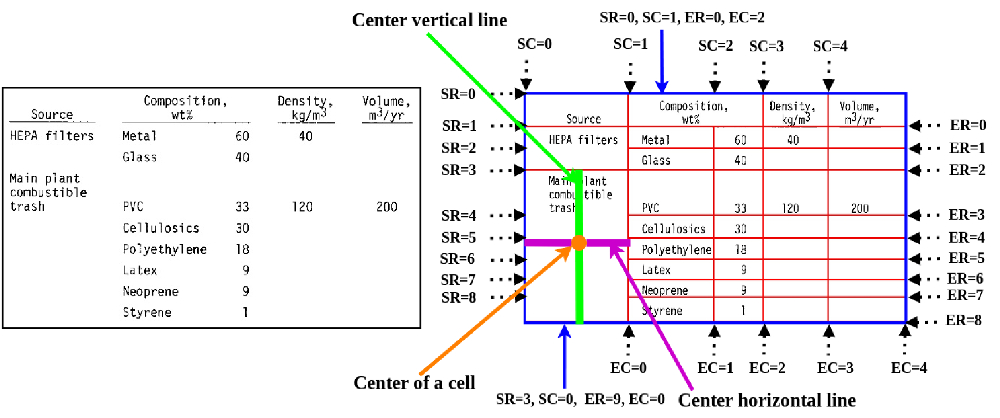}}
\end{center}
\caption{Visual illustration of cell spanning information along rows and columns of a table from {\sc unlv} dataset. \textbf{Left Image:} shows original table image in {\sc unlv} and \textbf{Right Image:} illustrates ground-truth cell spanning information. } \label{fig_ts}
\end{figure*}

\section{TabStruct-Net}\label{proposed_approach}

%%%%%%%%%%%%%%%%%%%%%%%%%%%%%%%%%%%%%%%%%%%%%%
%%figure for GNN for structure recognition
\begin{figure*}[t]
\begin{center}
\fbox{
\includegraphics[width=0.95\linewidth, height=0.35\linewidth]{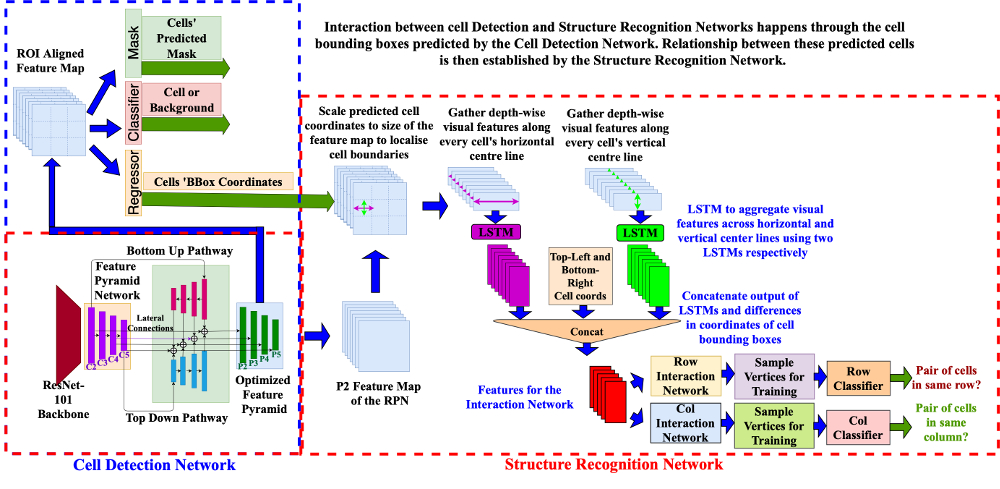}}
\end{center}
\caption{Our {\sc t}ab{\sc s}truct-{\sc n}et. Modified {\sc rpn} in cell detection network, which consists of both top-down and bottom-up pathways to better capture low-level visual features. P2 layer of the optimized feature pyramid is used in the structure recognition network to extract visual features.}\label{fig_gnn}
\end{figure*}

Our solution for table structure recognition progresses in three steps --- (a) detection of table cells; (b) establishing row/column relationships between the detected cells, and (c) post-processing step to produce the {\sc xml} output as desired. Figure~\ref{fig_block} depicts the block diagram of our approach. 

\subsection{Top-Down: Cell Detection}

The first step of our solution for table structure recognition is localization of individual cells in a table image, for which we use the popular object detection paradigm. The difference from natural scene images, however, is an inherent association between table cells.  Recent success of {\sc r-cnn}s~\cite{girshick2014rich} and its improved modifications (Fast {\sc r-cnn}~\cite{girshick2015fast}, Faster {\sc r-cnn}~\cite{ren2015faster}, Mask {\sc r-cnn}~\cite{he2017mask}) have shown significant success in object detection in natural scene images. Hence, we employ Mask {\sc r-cnn}~\cite{he2017mask} for our solution with additional enhancements --- (a) we augment the Region Proposal Network ({\sc rpn}) with dilated convolutions~\cite{yu2015multi,chen2017deeplab} to better capture long-range row and column visual features of the table. This improves detection of multi-row/column spanning and multi-line cells ; (b) inspired by~\cite{woo2019gated}, we append the feature pyramid network with a top-down pathway, which propagates high-level semantic information to low-level feature maps. This allows the network to work better for cells with varying scales; and (c) we append additional losses during the training phase in order to model the inherent structural constraints. We formulate two ways of incorporating this information --- (i) through an end-to-end training of cell detection and the structure recognition networks (explained next), and (ii) through a novel alignment loss function. For the latter, we make use of the fact that every pair of cells is aligned horizontally if they span the same row and aligned vertically if they span the same column. For the ground truth, where tight bounding boxes around the cells' content are provided~\cite{gobel2013icdar,chi2019complicated,zhong2019image}, we employ an additional ground truth pre-processing step to ensure that bounding boxes of cells in the same row and same column are aligned vertically and horizontally, respectively. We model these constraints during the training in the following manner:
\begin{equation*}
\begin{array}{l}
L_{1} = \sum_{r\in SR}\sum_{c_i, c_j\in r} ||y1_{c_i} - y1_{c_j}||_2^2 \text{,}\  
L_{2} = \sum_{r\in ER}\sum_{c_i, c_j\in r} ||y2_{c_i} - y2_{c_j}||_2^2 \\
L_{3} = \sum_{c\in SC}\sum_{c_i, c_j\in c} ||x1_{c_i} - x1_{c_j}||_2^2\ \text{and} \
L_{4} = \sum_{c\in EC}\sum_{c_i, c_j\in c} ||x2_{c_i} - x2_{c_j}||_2^2
\end{array}
\end{equation*}
Here, $SR$, $SC$, $ER$ and $EC$ represent starting row, starting column, ending row and ending column indices as shown in Figure~\ref{fig_ts}. Also, $c_i$ and $c_j$ denote two cells in a particular row $r$ or column $c$; $x1_{c_i}$, $y1_{c_i}$, $x2_{c_i}$ and $y2_{c_i}$ represent bounding box coordinates X-start, Y-start, X-end and Y-end respectively of the cell $c_i$. These losses ($L_{1}$, $L_{2}$, $L_{3}$, $L_{4}$) can be interpreted as constraints that enforce proper alignment of cells beginning from same row, ending on same row, beginning from same column and ending on same column respectively. Alignment loss is defined as
\begin{equation}
\begin{array}{l}
L_{align} = L_{1}+L_{2}+L_{3}+L_{4}. \label{equation_structural_loss}
\end{array}
\end{equation}

\subsection{Bottom-Up: Structure Recognition}
                     
We formulate the table structure recognition using graphs similar to~\cite{qasim2019rethinking}. We consider each cell of the table as a vertex and construct two adjacency matrices - a row matrix $M_{row}$ and a column matrix $M_{col}$ which describe the association between cells with respect to rows and columns. $M_{row}, M_{col}\in \mathbb{R}^{N_{cells} \times N_{cells}}$. $M_{row_{i,j}} = 1$ or $M_{col_{i,j}} = 1$ if cells $i,j$ belong to the same row or column, else $0$.

The structure recognition network aims to predict row and column relationships between the cells predicted by the cell detection module during training and testing. During training, only those predicted table cells are used for structure recognition which overlap with the ground truth table cells having an IoU greater than or equal to 0.5. This network has three components:
\begin{itemize}
    \item \textit{Visual Component:} We use visual features from P2 layer (refer Figure  \ref{fig_gnn}) of the feature pyramid based on the linear interpolation of cell bounding boxes predicted by the cell detection module. In order to encode cells' visual characteristics across their entire height and width, we pass the gathered P2 features for every cell along their centre horizontal and centre vertical lines using {\sc lstm}~\cite{hochreiter1997long} to obtain the final visual features (refer Figure~\ref{fig_gnn}) (as opposed to visual features corresponding to cells' centroids only as in~\cite{qasim2019learning}).
    \item \textit{Interaction Component:} We use the {\sc dgcnn} architecture based on graph neural networks used in~\cite{qasim2019learning} to model the interaction between geometrically neighboring detected cells. It's output, termed as interaction features, is a fixed dimensional vector for every cell that has information aggregated from its neighbouring table cells.
    \item \textit{Classification Component:} For a pair of table cells, the interaction features are concatenated and appended with difference between cells' bounding box coordinates. This is fed as an input to the row/column classifiers to predict row/column associations. Please note that we use the same~\cite{qasim2019learning} Monte Carlo based sampling to ensure efficient training and class balancing. During testing time, however, predictions are made for every unique pair of table cells.
\end{itemize}

We train the cell detection and structure recognition networks in a joint manner (termed as {\sc t}ab{\sc s}truct-{\sc n}et) to collectively predict cell bounding boxes along with row and column adjacency matrices. Further, the two structure recognition pathways for row and column adjacency matrices are put together in parallel. The visual features prepared using {\sc lstm}s for every vertex are duplicated for both the pathways, after which they work in a parallel manner. The overall empirical loss of {\sc t}ab{\sc s}truct-{\sc n}et is given by:
\begin{equation}
\begin{array}{l}
L = L_{box} + L_{cls} + L_{mask} + L_{align} + L_{{\sc gnn}}, \label{equation_modified_loss} 
\end{array}
\end{equation}
where $L_{box}$, $L_{cls}$ and $L_{mask}$ are bounding box regression loss, classification loss and mask loss, respectively defined in Mask {\sc r-cnn}~\cite{he2017mask}, $L_{align}$ is alignment loss which is modeled as a regularizer (defined in Eq.~\ref{equation_structural_loss}) and $L_{{\sc gnn}}$ is the cross-entropy loss back propagated from the structure recognition module of {\sc t}ab{\sc s}truct-{\sc n}et. The additional loss components help the model in better alignment of cells belonging to same rows/columns during training, and in a sense fine-tunes the predicted bounding boxes that makes it easier for post-processing and structure recognition in the subsequent step. 

\subsection{Post-Processing}

Once all the cells and their row/column adjacency matrices are predicted, we create the {\sc xml} interpretable output as a post-processing step. From the cell coordinates along with row and column adjacency matrix, $SR$, $SC$, $ER$ and $EC$ indexes are assigned to each cell, which indicate spanning of that cell along rows and columns. We use Tesseract~\cite{smith2007overview} to extract the content of every predicted cell. The {\sc xml} output for every table image finally contains coordinates of predicted cell bounding boxes and along with cell spanning information and its content. 

\section{Experiments} \label{experiments}

\subsection{Datasets} \label{dataset}

We use various benchmark datasets - {\sc s}ci{\sc tsr}~\cite{chi2019complicated}, {\sc s}ci{\sc tsr}-{\sc comp}~\cite{chi2019complicated}, {\sc icdar}-2013 table recognition~\cite{gobel2013icdar}, {\sc icdar}-2019 c{\sc td}a{\sc r} archival~\cite{gao2019icdar}, {\sc unlv}~\cite{shahab2010open}, Marmot extended~\cite{paliwal2019tablenet}, {\sc t}able{\sc b}ank~\cite{li2019tablebank} and {\sc p}ub{\sc t}ab{\sc n}et~\cite{zhong2019image} datasets for extracting structure information of tables. Statistics of these datasets are listed in Table~\ref{table_dataset_statistics}. 
%%%%%%%%%%%%%%%%%%%%%%%%%%%%%%%%%%%%%%%%%%%%%%%%
%table for dataset statistics
\begin{table*}[ht!]
%\addtolength{\tabcolsep}{-0.5pt}
\begin{center}
\begin{tabular}{|r |r |r |r |r |r |r |r |r |r|r|} \hline
   &{\sc s}ci{\sc tsr} &{\sc s}ci{\sc tsr} &{\sc icdar} &{\sc icdar}-2013 &{\sc icdar} &{\sc unlv} &{\sc unlv-} &Marmot &{\sc t}able &{\sc p}ub{\sc t}ab{\sc n}et \\
   &   &{\sc comp} &2013  &-partial &2019 &  &partial &extended &{\sc b}ank  & \\ \hline
\textbf{Train} &12000 &12000 &-  &124 &600 &- &446 &1016 &145K &339K \\
\textbf{Test}  &3000  &716   &158 &34 &150 &558 &112 &- &1000 &114K \\ \hline
\end{tabular}
\end{center}
\caption{Statistics of the datasets used for our experiments.} \label{table_dataset_statistics}
\end{table*}

\subsection{Baseline Methods}

We compare the performance of our {\sc t}ab{\sc s}truct-{\sc n}et against seven benchmark methods --- {\sc d}eep{\sc d}e{\sc srt}~\cite{schreiber2017deepdesrt}, {\sc t}able{\sc n}et~\cite{paliwal2019tablenet}, {\sc g}raph{\sc tsr}~\cite{chi2019complicated}, {\sc splerge}~\cite{table_splitting}, {\sc dgcnn}~\cite{qasim2019rethinking}, Bi-directional {\sc gru}~\cite{Khan_2019} and Image-to-Text~\cite{li2019tablebank}.

\subsection{Implementation Details} \label{implementation_details}

{\sc t}ab{\sc s}truct-{\sc n}et\footnote{Our code is available at \url{ https://github.com/sachinraja13/TabStructNet.git}} has been trained and evaluated with table images scaled to a fixed size of 1536$\times$1536 while maintaining the original aspect ratio as the input. While training, cell-level bounding boxes along with row and column adjacency matrices (prepared from start-row, start-column, end-row and end-column indices) are used as the ground truth. We use {\sc nvidia titan x gpu} with 12 {\sc gb} memory for our experiments and a batch-size of 1. Instead of using 3$\times$3 convolution on the output feature maps from the {\sc fpn}, we use a dilated convolution with filter size of 2$\times$2 and dilation parameter of 2. Also, we use the {\sc r}es{\sc n}et-101 backbone that is pre-trained on {\sc ms-coco}~\cite{ms_coco} dataset. Dilated convolution blocks of filter size 7 are used in the {\sc fpn}. To compute region proposals, we use 0.5, 1 and 2 as the anchor scale and anchor box sizes of 8, 16, 32, 64 and 128. {\sc lstm}s used to gather visual features have a depth of 128. The final memory state of the {\sc lstm} layers is concatenated with the cell's coordinates to prepare features for the interaction network.  Further, for generation of the row/column adjacency matrices, we use 2400 as the maximum number of vertices keeping in mind dense tables. Next, features from 40 neighboring vertices are aggregated using an edge convolution layer followed by a dense layer of size 64 with ReLu activation. Since every input table may contain hundreds of table cells, training can be a time consuming process. To achieve faster training, we employ a two-stage training process. In the first stage, we use 2014 anchors and 512 {\sc r}o{\sc i}s, and in the second stage, we use with 3072 anchors and 2048 {\sc r}o{\sc i}s. During both the stages, we use 0.001 as the learning rate, 0.9 as the momentum and 0.0001 as the weight decay regularisation.

%%%%%%%%%%%%%%%%%%%%%%%%%%%%%%%%%%%%%%%%%%%%%
%quantitative results of physical structure recognition using our method for all datasets
\begin{table}[t!]
%\addtolength{\tabcolsep}{-1.3pt}
\begin{center}
\begin{tabular}{|l|l|l l l |l l l|} \hline
\textbf{Test Dataset} &\textbf{Train Dataset} &\multicolumn{3}{c|}{\textbf{S-A}} &\multicolumn{3}{c|}{\textbf{S-B}} \\ \cline{3-8}
  &  &\textbf{P}$\uparrow$ &\textbf{R}$\uparrow$ &\textbf{F1}$\uparrow$ &\textbf{P}$\uparrow$ &\textbf{R}$\uparrow$ &\textbf{F1}$\uparrow$ \\ \hline
{\sc icdar-2013} &{\sc s}ci{\sc tsr}          &0.915 &0.897 &0.906 &0.976 &0.985 &0.981 \\  
{\sc icdar-2013}-partial &{\sc s}ci{\sc tsr}  &0.930 &0.908 &0.919 &0.991 &0.993 &0.992 \\  
{\sc s}ci{\sc tsr} &{\sc s}ci{\sc tsr}        &0.927 &0.913	&0.920 &0.989 &0.993 &0.991 \\
{\sc s}ci{\sc tsr}-comp	&{\sc s}ci{\sc tsr}	  &0.909 &0.882	&0.895 &0.981 &0.987 &0.984 \\
{\sc unlv}-partial	&{\sc s}ci{\sc tsr}       &0.849 &0.828	&0.839 &0.992 &0.994 &0.993 \\
{\sc icdar-2019}    &{\sc s}ci{\sc tsr}       &0.595 &0.572	&0.583 &0.924 &0.899 &0.911 \\
{\sc icdar-2019}    &{\sc icdar-2019}	      &0.803 &0.768	&0.785 &0.975 &0.957 &0.966 \\
{\sc icdar-2019}	&{\sc s}ci{\sc tsr}+{\sc icdar-2019} &0.822 &0.787	&0.804 &0.975 &0.958 &0.966 \\ \hline
\end{tabular}
\end{center}
\caption{shows the performance of our {\sc t}ab{\sc s}truct-{\sc n}et for physical table structure recognition on various benchmark datasets. \label{physical_structure_recognition_all}}
\vspace{-1em}
\end{table}

%%%%%%%%%%%%%%%%%%%%%%%%%%%%%%%%%%%%%%%%%%%%%
%quantitative results of logical structure recognition using our method for all datasets
\begin{table}[ht!]
%\addtolength{\tabcolsep}{-1.3pt}
\begin{center}
\begin{tabular}{|l|l|l|l|} \hline
\textbf{Test Dataset} &\textbf{Train Dataset} &\textbf{Metric}  &\textbf{Score} \\ \hline
{\sc t}able{\sc b}ank-{\sc w}ord &{\sc s}ci{\sc tsr} &{\sc bleu} &0.914 \\ 
{\sc t}able{\sc b}ank-{\sc l}a{\sc t}e{\sc x} &{\sc s}ci{\sc tsr} &{\sc bleu} &0.937 \\
{\sc t}able{\sc b}ank-{\sc w}ord+{\sc l}a{\sc t}e{\sc x} &{\sc s}ci{\sc tsr} &{\sc bleu} &0.916 \\
{\sc p}ub{\sc t}ab{\sc n}et &{\sc s}ci{\sc tsr} &{\sc teds} &0.901 \\ \hline 
\end{tabular}
\end{center}
\caption{shows the performance of our {\sc t}ab{\sc s}truct-{\sc n}et for logical table structure recognition on various benchmark datasets. \label{logical_structure_recognition_all}}
\vspace{-1em}
\end{table}

\subsection{Evaluation Measures} \label{evaluation_measure}

We use various existing measures --- precision, recall and F1~\cite{chi2019complicated,gobel2013icdar,shahab2010open} to evaluate the performance of our model for recognition of physical structure of tables. For recognition of logical structure of tables, we use {\sc bleu}~\cite{papineni2002bleu} score as used in~\cite{li2019tablebank} and Tree-Edit-Distance-based similarity ({\sc teds})~\cite{zhong2019image}. Since {\sc xml} is our final output for table structure recognition, we also use {\sc bleu}~\cite{papineni2002bleu}, {\sc cide}r~\cite{vedantam2015cider} and {\sc rouge}~\cite{lin2004rouge} scores to compare generated {\sc xml} and ground truth {\sc xml} on spanning information and content of every cell. We first calculate these scores separately on each table and then compute averaged score as the final result. We consistently use an IoU threshold of 0.6 to compute the confusion matrix. Please note that only non-empty table cells are considered similar to~\cite{gobel2013icdar} for the evaluation. It is also important to note that in our evaluation, we do not take into account content of the cell and the confusion matrix is computed solely on the basis of IoU overlap with the ground truth cell box.

\subsection{Experimental Setup} \label{experimental_setup}

One major challenge in the comparison study with the existing methods is the inconsistent use of additional information (e.g., meta-features extracted from the {\sc pdf}s~\cite{table_splitting}, content-level bounding boxes from ground truths~\cite{paliwal2019tablenet,chi2019complicated} and cell's location features generated from synthetic dataset~\cite{qasim2019rethinking}). Hence, we do experiments in two different setups 
\begin{itemize}
    \item \textbf{Setup-A (S-A):} using only table image as the input
    \item \textbf{Setup-B (S-B):} using table image along with additional information (e.g., cell bounding boxes) as the input. For this, instead of removing the cell detection component from the network, we ignore the predicted boxes and use the ground truth ones as input for structure recognition.
\end{itemize}

\section{Results on Table Structure Recognition} \label{result_analysis_table_structure_recognition}

Tables~\ref{physical_structure_recognition_all} and~\ref{logical_structure_recognition_all} summarize the performance of our model on standard datasets used in the space of table structure recognition.

\subsection{Analysis of Results}

Table~\ref{table_physical_icdar_2013} presents results on {\sc icdar}-2013 dataset. In S-A, we observe that our model outperforms {\sc d}eep{\sc d}e{\sc srt}~\cite{schreiber2017deepdesrt} method by a 27.5\% F1 score. This is because cell coordinates for the latter are obtained by row and column intersections, making it unable to recognize cells that span multiple rows/columns. For dense tables (small inter-row spacing), row segmentation results of {\sc d}eep{\sc d}e{\sc srt} combined multiple rows into one in several instances. {\sc s}plit+{\sc h}euristic~\cite{table_splitting} method outperforms {\sc t}ab{\sc s}truct-{\sc n}et by a small margin, however, it requires {\sc icdar}-2013 dataset-specific cell merging heuristics and is trained on a considerably larger set of images. Therefore, a direct comparison of ({\sc s}plit+{\sc h}euristic) with our method is not fair. Nevertheless, comparable results of {\sc t}ab{\sc s}truct-{\sc n}et indicates its robustness to {\sc icdar}-2013 dataset, without using any kind of dataset-specific post-processing. However, if compared under the same training environment and no post-processing, our model outperforms {\sc splerge} with a 3\% average F1 score. {\sc splerge} works well for datasets where ground truth bounding boxes are annotated at the content-level instead of cell-level. This is because it allows for a wider area for a prospective prediction of a row/column separator. Further, since it is based on cell detection through the row and column separators, it is not agnostic to input image noise such as skew and rotations. This method is susceptible to dataset-specific post-processing as opposed to ours, where no post-processing is needed.

%%%%%%%%%%%%%%%%%%%%%%%%%%%%%%%%%%%%%%%%%%%%%
%quantitative results on icdar-2013 complete dataset
\begin{table}[ht!]
%\addtolength{\tabcolsep}{-1.3pt}
\begin{center}
\begin{tabular}{|l | l| r| l |c c c|} \hline
\textbf{Method} &\multicolumn{2}{|c|}{\textbf{Training}} &\textbf{Experimental} &\textbf{P}$\uparrow$ &\textbf{R}$\uparrow$ &\textbf{F1}$\uparrow$ \\ \cline{2-3}
  &\textbf{Dataset} &\textbf{\#Images} &\textbf{Setup} & & & \\ \hline
{\sc d}eep{\sc d}e{\sc srt}~\cite{schreiber2017deepdesrt} &{\sc s}ci{\sc tsr} &12K &S-A &0.631 &0.619 &0.625 \\
{\sc splerge}~\cite{table_splitting} &{\sc s}ci{\sc tsr} &12K &S-A &0.883 &0.875 &0.879 \\
{\sc s}plit+{\sc h}euristic~\cite{table_splitting} &Private~\cite{table_splitting} &83K &S-A &\textbf{0.938} &\textbf{0.922} &\textbf{0.930} \\
{\sc t}ab{\sc s}truct-{\sc n}et (our) &{\sc s}ci{\sc tsr} &12K &S-A &0.915 &0.897 &0.906 \\ \hline
{\sc t}able{\sc n}et~\cite{paliwal2019tablenet} &Marmot Extended &1K &S-B &0.922 &0.899 &0.910 \\ 
{\sc g}raph{\sc tsr}~\cite{chi2019complicated} &{\sc s}ci{\sc tsr} &12K &S-B &0.885 &0.860 &0.872 \\ 
{\sc s}plit-{\sc pdf}~\cite{table_splitting} &Private~\cite{table_splitting} &83K &S-B &0.920 &0.913 &0.916 \\ 
{\sc s}plit-{\sc pdf} & & & & & & \\
+{\sc h}euristic~\cite{table_splitting} &Private~\cite{table_splitting} &83K &S-B &0.959 &0.946 &0.953 \\ 
{\sc dgcnn}~\cite{qasim2019rethinking} &{\sc s}ci{\sc tsr} &12K &S-B &0.972 &0.983 &0.977 \\
{\sc t}ab{\sc s}truct-{\sc n}et (our) &{\sc s}ci{\sc tsr} &12K &S-B &\textbf{0.976} &\textbf{0.985} &\textbf{0.981} \\ \hline 
\end{tabular}
\end{center}
\caption{Comparison of results for physical structure recognition on {\sc icdar}-2013 dataset. \textbf{\#Images:} indicates number of table images in the training set. \textbf{Heuristic:} indicates dataset specific cell merging rules for various models in~\cite{table_splitting}.\label{table_physical_icdar_2013}}
\vspace{-1em}
\end{table}

In S-B, {\sc t}ab{\sc s}truct-{\sc n}et sets up a state-of-the-art benchmark on the {\sc icdar}-2013 dataset, outperforming all the existing methods~\cite{qasim2019rethinking,table_splitting,chi2019complicated,paliwal2019tablenet}. It is further interesting to note that our technique outperforms {\sc s}plit-{\sc pdf}+{\sc h}euristic model also without needing any post-processing. It is because our enhancements to the {\sc dgcnn}~\cite{qasim2019rethinking} model can capture the visual characteristics of a cell across a larger span through {\sc lstm}s. We observe that our model achieves significantly improved performance when content-level bounding boxes are used instead of cell-level, which are much easier to obtain with the help of {\sc ocr} tools and {\sc pdf} meta-information.

%%%%%%%%%%%%%%%%%%%%%%%%%%%%%%%%%%%%%%%%%%%%%
%robustness of method using varying IoU on icdar-2013 complete dataset
\begin{table}[ht!]
\addtolength{\tabcolsep}{-1.0pt}
\begin{center}
\begin{tabular}{|l |l| l| l l l| l l l|} \hline
\textbf{CD Network} & \textbf{SR Network} &\textbf{IoU} &\multicolumn{3}{l|}{\textbf{CD Scores}} &\multicolumn{3}{l|}{\textbf{SR Scores}} \\ \cline{4-9}
 &  &{TH} &\textbf{P}$\uparrow$ &\textbf{R}$\uparrow$ &\textbf{F1}$\uparrow$ &\textbf{P}$\uparrow$ &\textbf{R}$\uparrow$ &\textbf{F1}$\uparrow$ \\ \hline
 &  &0.5	&\textbf{0.935} &\textbf{0.942} &\textbf{0.938} &\textbf{0.927} &\textbf{0.911}	&\textbf{0.919} \\  
 &      &0.6 &0.921 &0.926	&0.923 &0.915 &0.897 &0.906 \\
Mask {\sc r-cnn}+{\sc td}+{\sc bu}+{\sc al} &{\sc dgcnn}+{\sc p}2+{\sc lstm}  &0.7 &0.815 &0.820	&0.817 &0.797 &0.785 &0.791  \\
 &      &0.8 &0.638 &0.653	&0.645 &0.629 &0.615 &0.622 \\
 &      &0.9 &0.275 &0.312	&0.292 &0.247 &0.236 &0.241 \\ \hline
 \end{tabular}
\end{center}
\caption{Physical structure recognition results on {\sc icdar}-2013 dataset for varying IoU thresholds to demonstrate {\sc t}ab{\sc s}truct-{\sc n}et's robustness. \textbf{ES:} Experimental Setup, \textbf{CD:} Cell Detection, \textbf{TH:} IoU threshold value, \textbf{SR:} Structure Recognition, \textbf{P2:} using visual features from P2 layer of the {\sc fpn} instead of using separate convolution blocks, \textbf{{\sc lstm}:} use of {\sc lstm}s to model visual features along center-horizontal and center-vertical lines for every cell, \textbf{{\sc td}+{\sc bu}:} use of Top-Down and Bottom-Up pathways in the {\sc fpn}, \textbf{AL:} addition of alignment loss as a regularizer to {\sc t}ab{\sc s}truct-{\sc n}et. \label{table_varying_ious_physical_icdar_2013}}
\vspace{-1.0em}
\end{table}

\begin{figure}[ht!]
\begin{center}
\fbox{
\includegraphics[width=0.29\linewidth, height=0.16\linewidth]{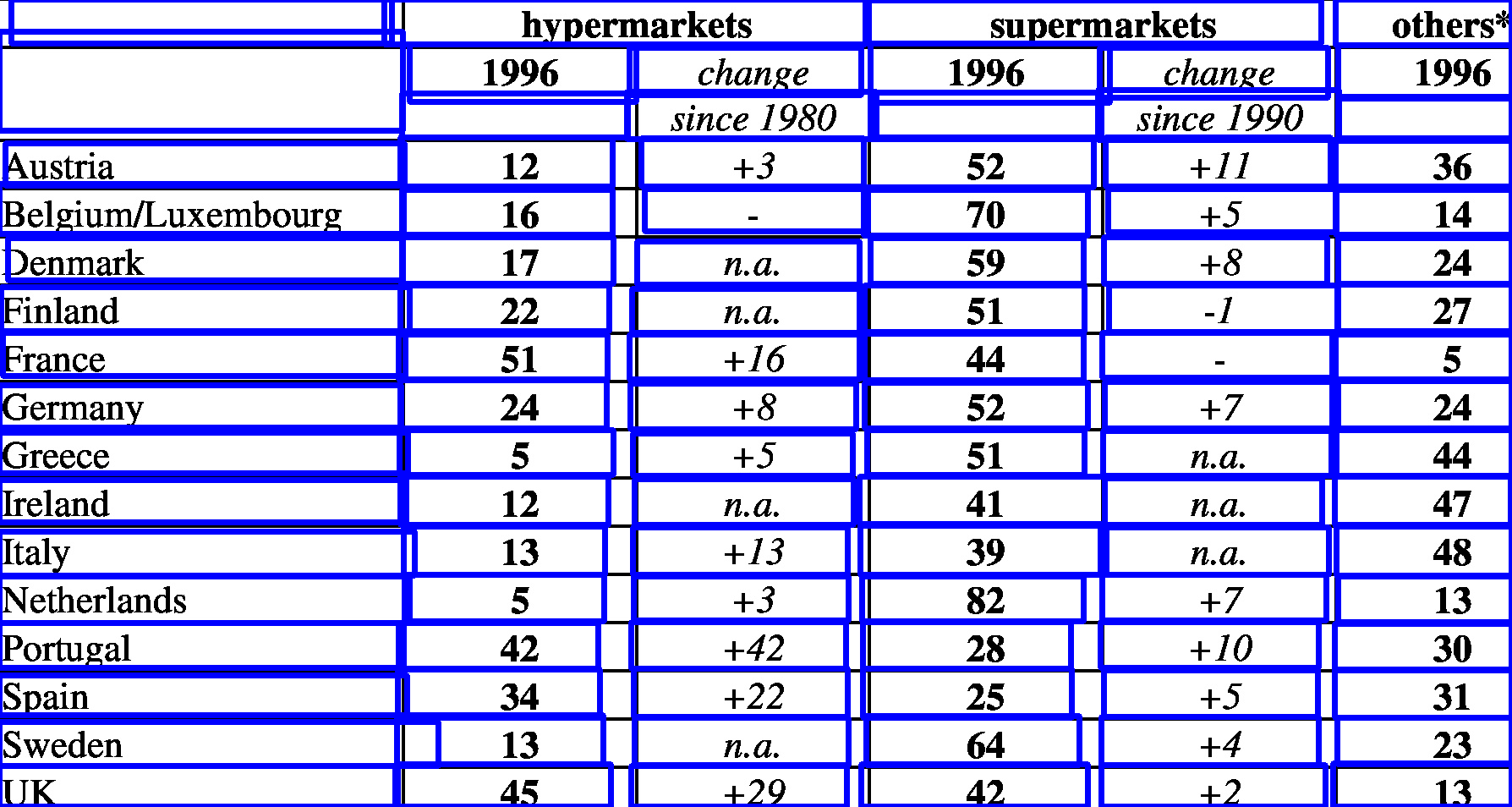}}
\hspace{-0.01\textwidth}
\fbox{
\includegraphics[width=0.29\linewidth, height=0.16\linewidth]{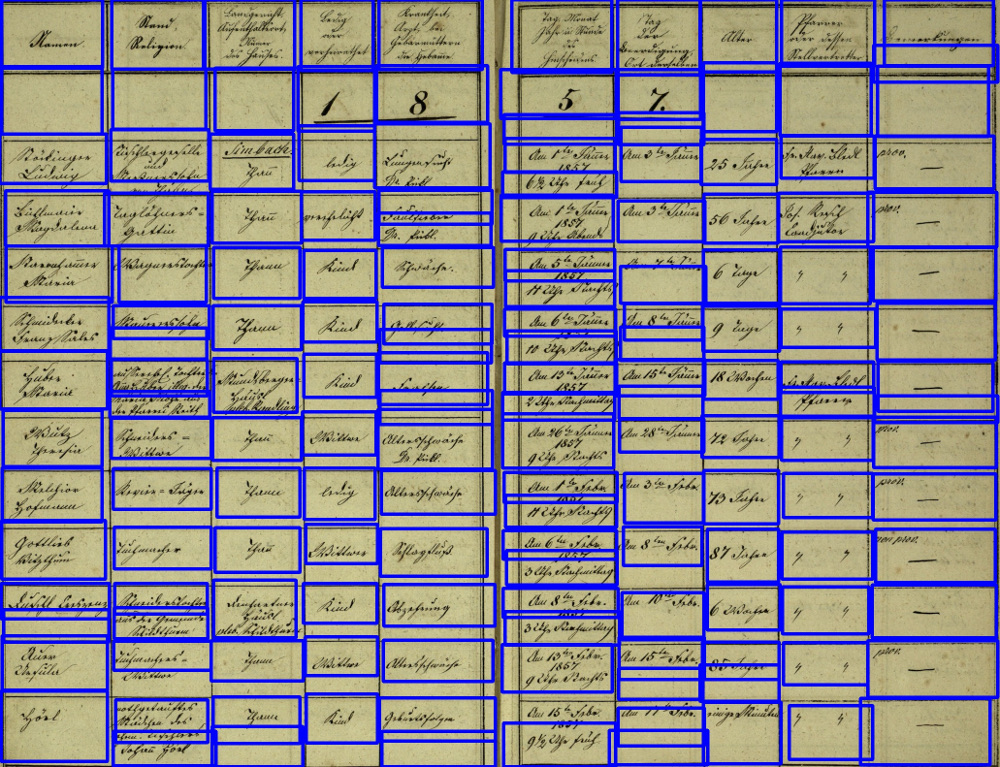}}
\hspace{-0.01\textwidth}
\fbox{
\includegraphics[width=0.29\linewidth, height=0.16\linewidth]{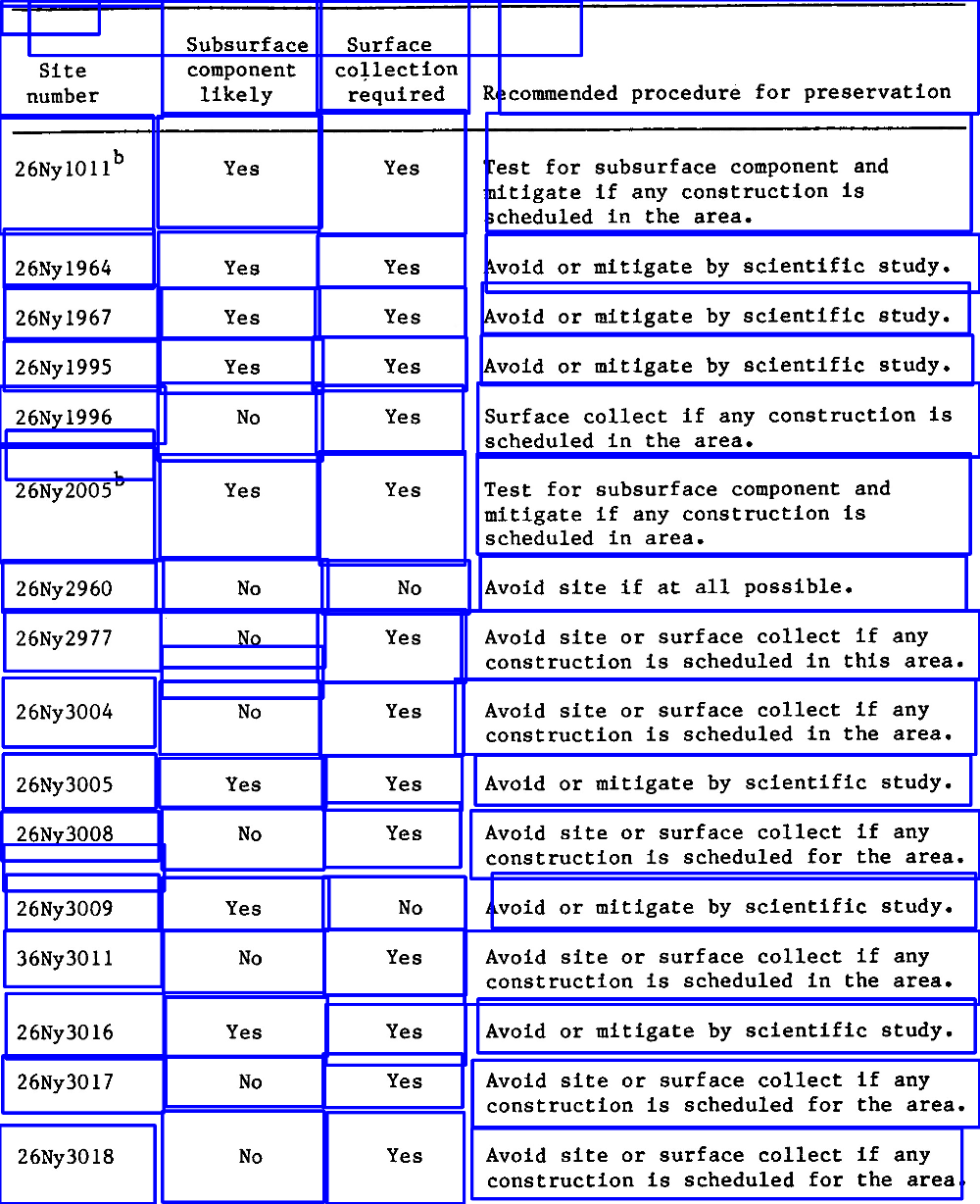}}
\vspace{0.001\textwidth}
\fbox{
\includegraphics[width=0.29\linewidth, height=0.16\linewidth]{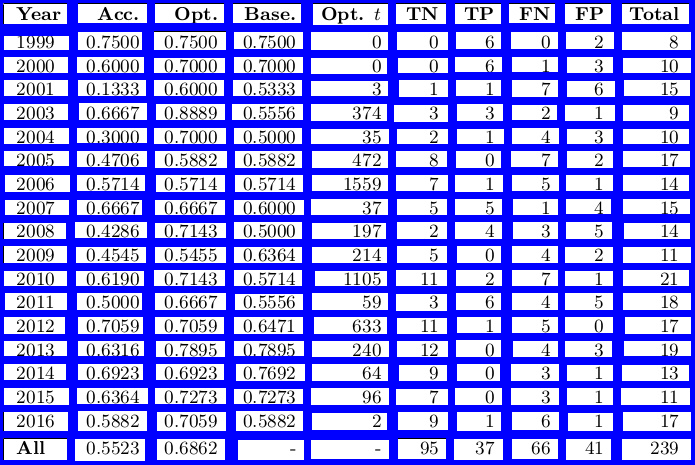}}
\hspace{-0.01\textwidth}
\fbox{
\includegraphics[width=0.29\linewidth, height=0.16\linewidth]{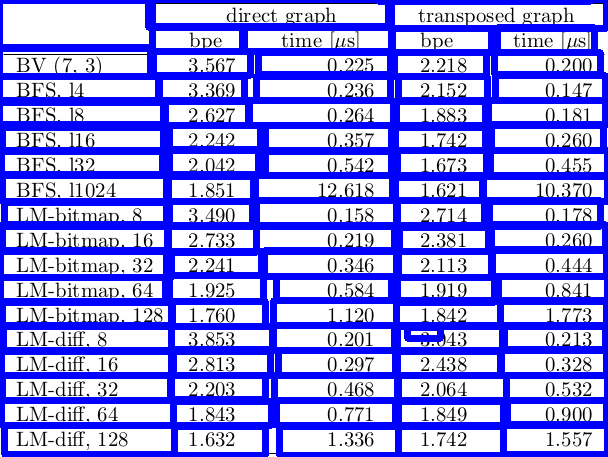}}
\hspace{-0.01\textwidth}
\fbox{
\includegraphics[width=0.29\linewidth, height=0.16\linewidth]{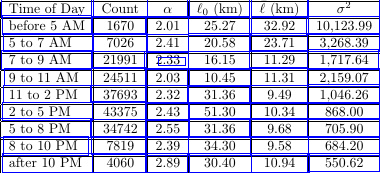}}
\end{center}
\caption{Sample intermediate cell detection results of {\sc t}ab{\sc s}truct-{\sc n}et on table images of {\sc icdar-2013}, {\sc icdar-2019} c{\sc td}a{\sc r} and {\sc unlv}, {\sc s}ci{\sc tsr}, {\sc s}ci{\sc tsr-comp} and {\sc t}able{\sc b}ank datasets.}
\label{fig_cell_detection_result}
\end{figure}

\begin{figure}[ht!]
\begin{center}
\fbox{
\includegraphics[width=0.29\linewidth, height=0.16\linewidth]{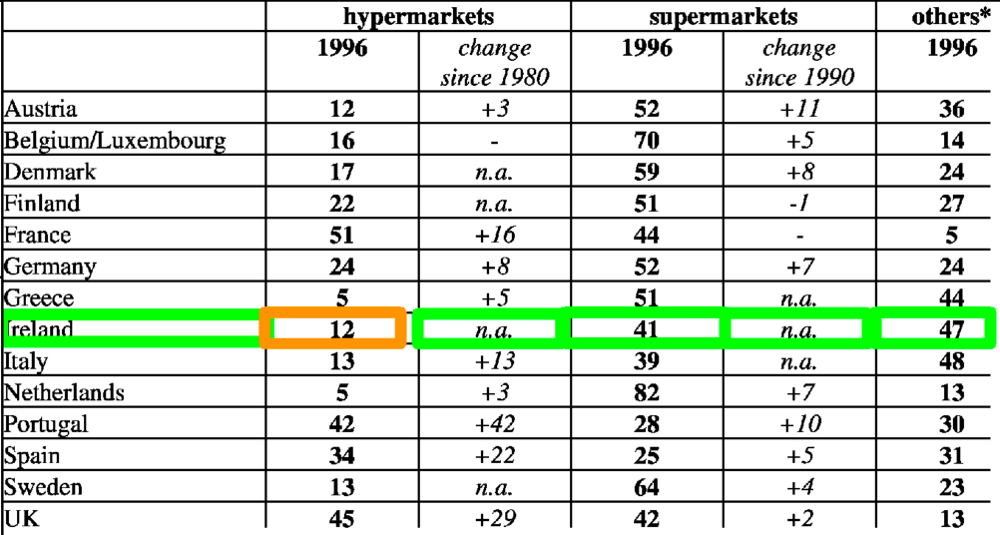}}
\hspace{-0.01\textwidth}
\fbox{
\includegraphics[width=0.29\linewidth, height=0.16\linewidth]{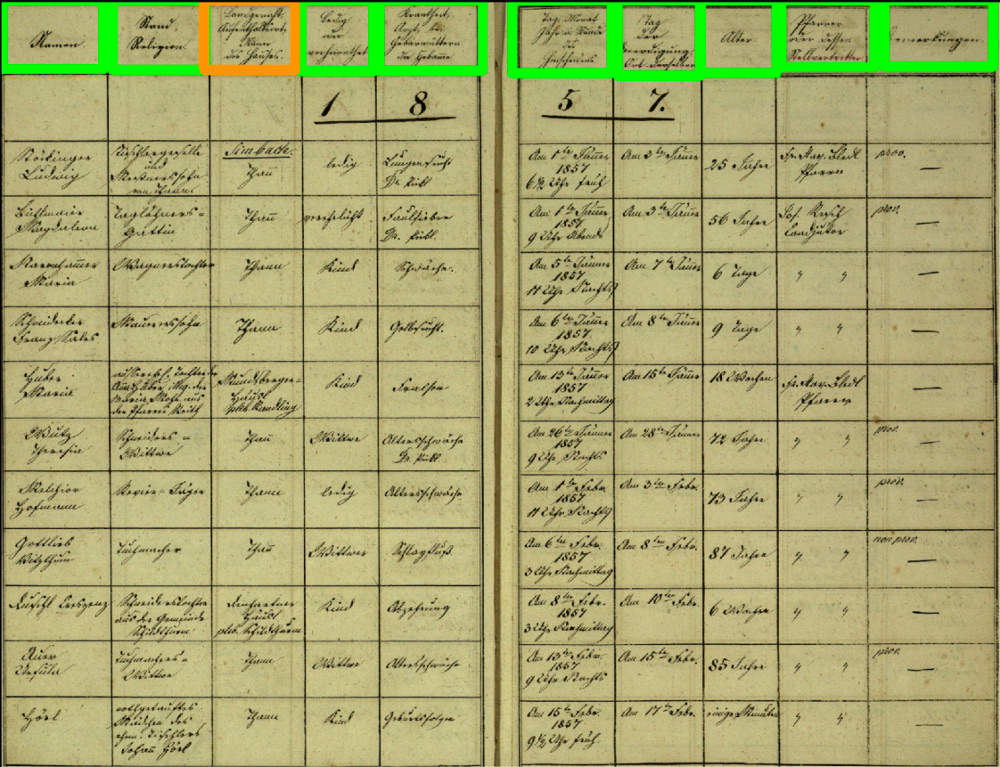}}
\hspace{-0.01\textwidth}
\fbox{
\includegraphics[width=0.29\linewidth, height=0.16\linewidth]{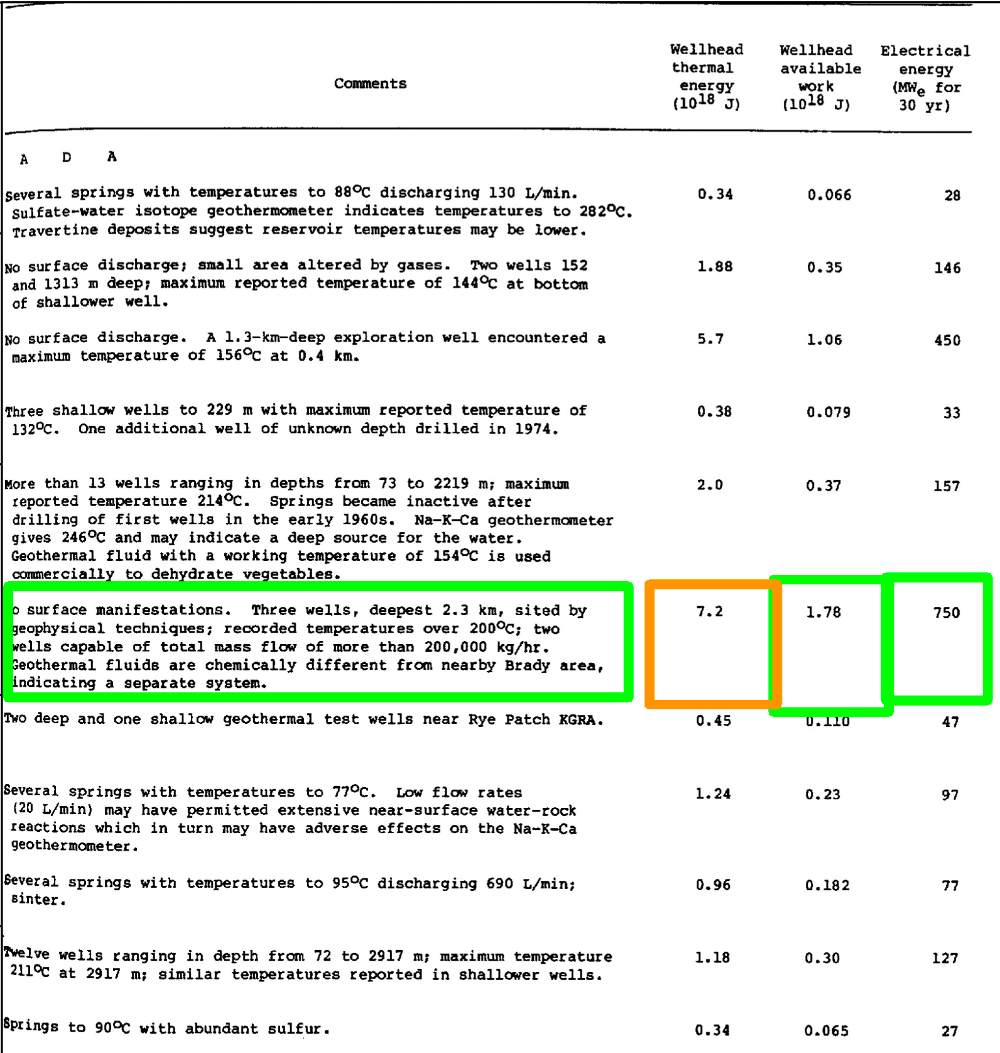}}
\vspace{0.003\textwidth}
\fbox{
\includegraphics[width=0.29\linewidth, height=0.16\linewidth]{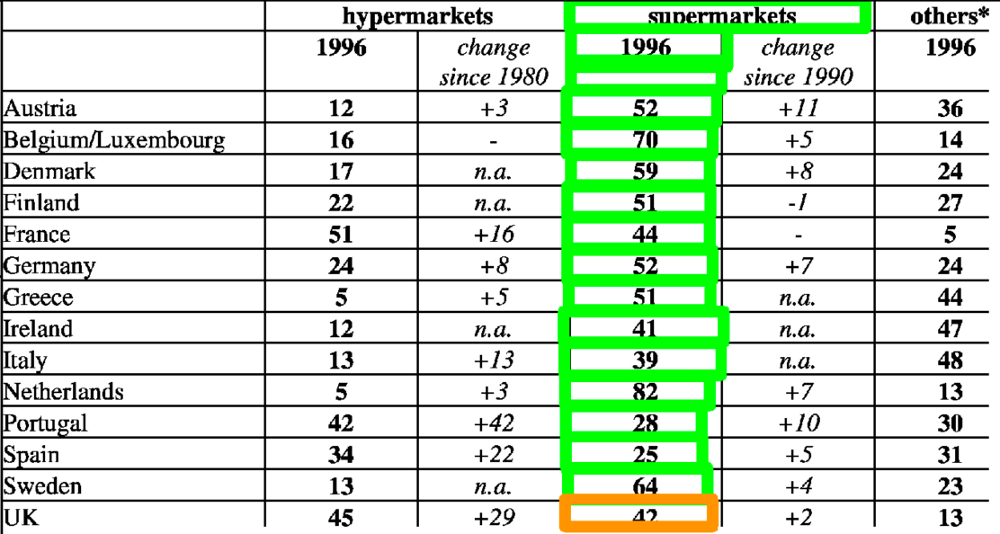}}
\hspace{-0.01\textwidth}
\fbox{
\includegraphics[width=0.29\linewidth, height=0.16\linewidth]{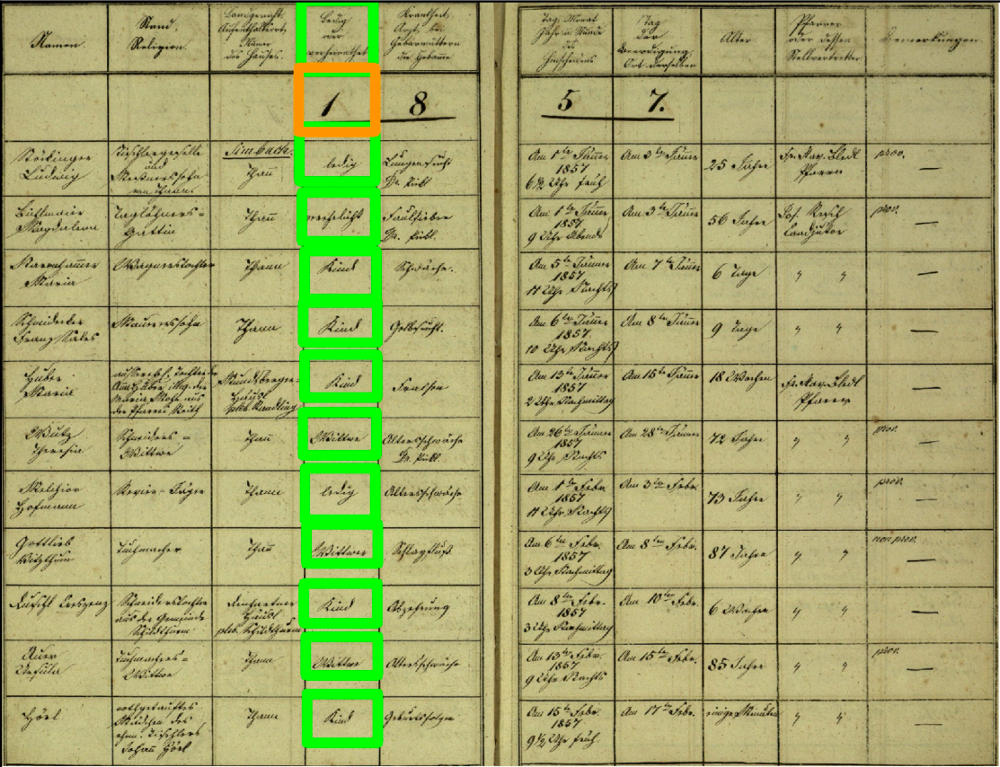}}
\hspace{-0.01\textwidth}
\fbox{
\includegraphics[width=0.29\linewidth, height=0.16\linewidth]{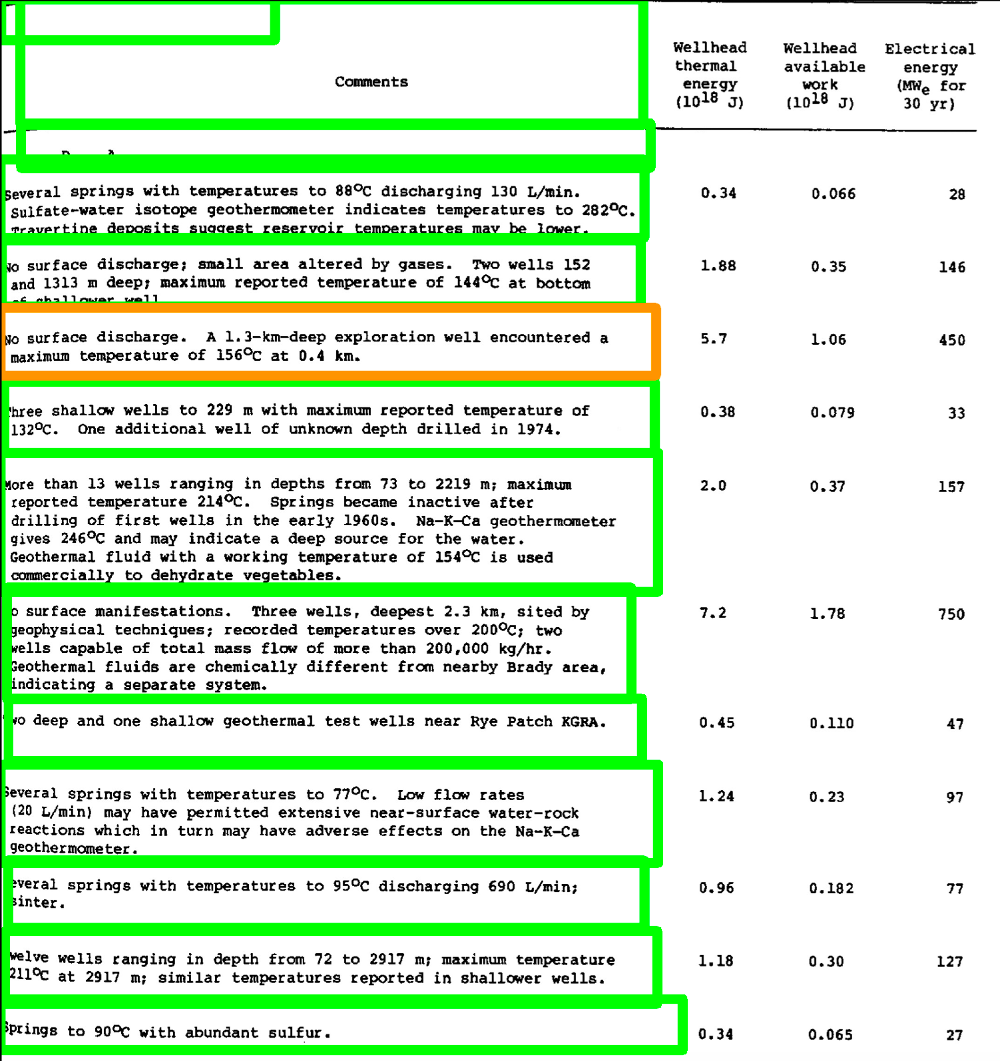}}
\end{center}
\caption{Sample structure recognition output of {\sc t}ab{\sc s}truct-{\sc n}et on table images of {\sc icdar-2013}, {\sc icdar-2019} c{\sc td}a{\sc r} archival and {\sc unlv} datasets. \textbf{First Row:} prediction of cells which belong to the same row. \textbf{Second Row:} prediction of cells which belong to the same column. Cells marked with orange colour represent the examine cells and cells marked with green colour represent those which belong to the same row/column of the examined cell.}
\label{fig_struture_recognition_result}
\end{figure}

\begin{figure}[ht!]
\begin{center}
\fbox{
\includegraphics[width=0.29\linewidth, height=0.16\linewidth]{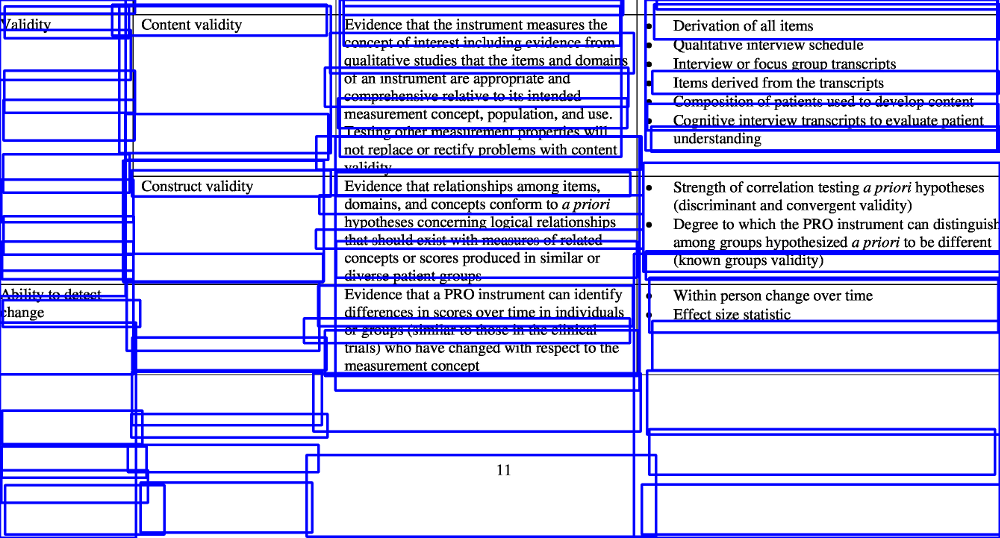}}
\hspace{-0.01\textwidth}
\fbox{
\includegraphics[width=0.29\linewidth, height=0.16\linewidth]{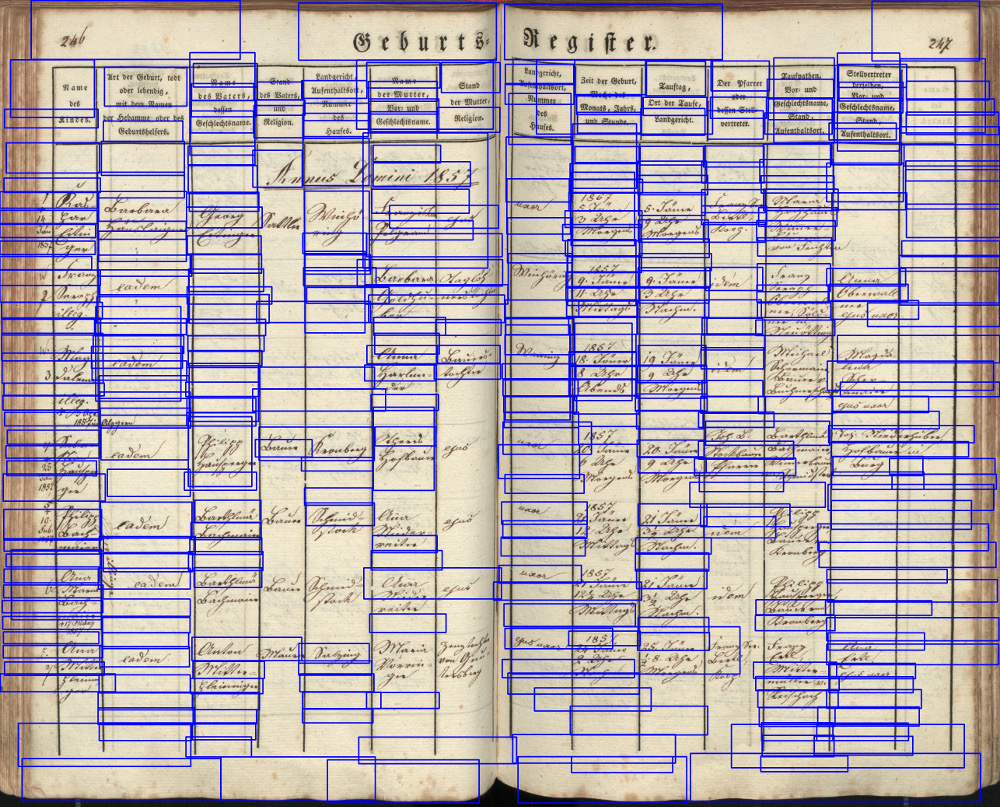}}
\hspace{-0.01\textwidth}
\fbox{
\includegraphics[width=0.29\linewidth, height=0.16\linewidth]{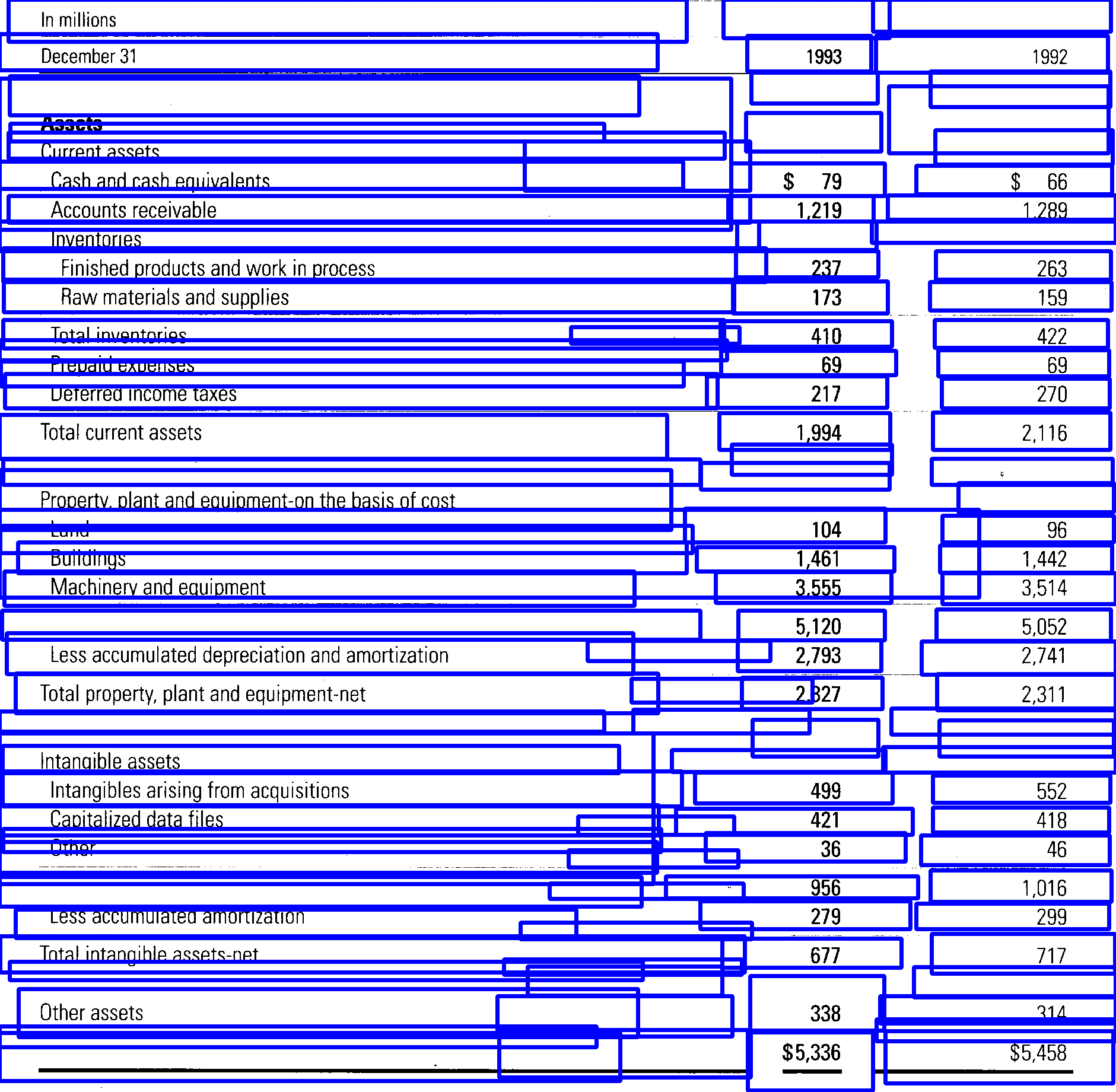}}
\vspace{0.001\textwidth}
\fbox{
\includegraphics[width=0.29\linewidth, height=0.16\linewidth]{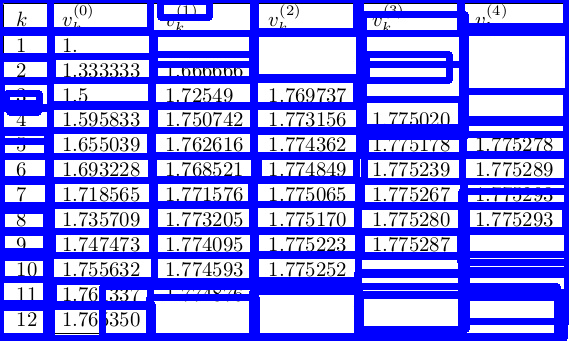}}
\hspace{-0.01\textwidth}
\fbox{
\includegraphics[width=0.29\linewidth, height=0.16\linewidth]{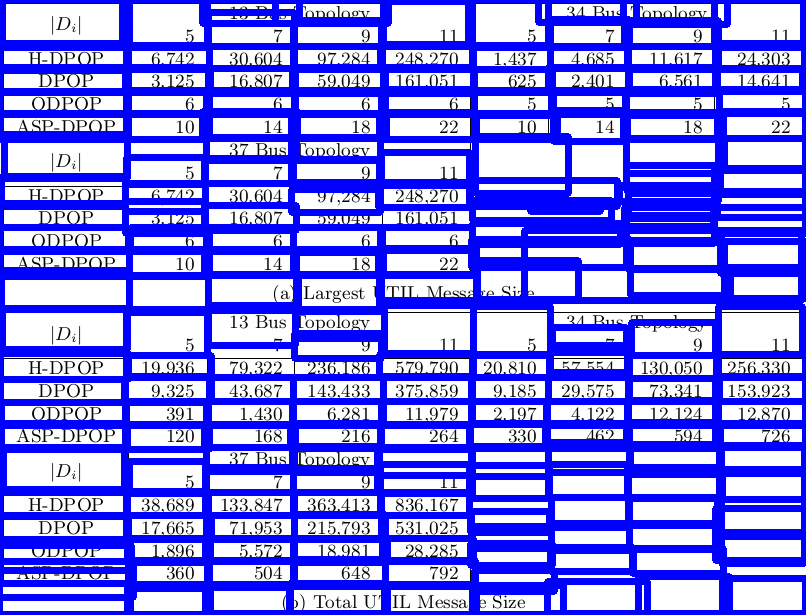}}
\hspace{-0.01\textwidth}
\fbox{
\includegraphics[width=0.29\linewidth, height=0.16\linewidth]{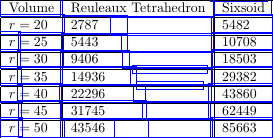}}
\end{center}
\caption{Sample intermediate cell detection results of {\sc t}ab{\sc s}truct-{\sc n}et on table images of {\sc icdar-2013}, {\sc icdar-2019} c{\sc td}a{\sc r}, {\sc unlv}, {\sc s}ci{\sc tsr}, {\sc s}ci{\sc tsr-comp} and {\sc t}able{\sc b}ank datasets illustrate failure of {\sc t}ab{\sc s}truct-{\sc n}et.}
\label{fig_failure_result}
\end{figure}

Table~\ref{table_varying_ious_physical_icdar_2013} shows the performance of our technique under the varying IoU thresholds. It can be inferred from the table that our model achieves an F1 score of 79.1\% on structure recognition with an IoU threshold value of as high as 0.7. For the IoU values of 0.5 and 0.6, our model's performance is 91.9\% and 90.6\%, respectively. It demonstrates the robustness of our model. Figures~\ref{fig_cell_detection_result} and~\ref{fig_struture_recognition_result} display some qualitative outputs of our method on the datasets discussed in Section~\ref{dataset}. Figure~\ref{fig_failure_result} shows some of the failure cases of cell detection by our method. It can be seen that our model fails for table images that have large amounts of empty spaces. 

%%%%%%%%%%%%%%%%%%%%%%%%%%%%%%%%%%%%%%
%figure images of cell detection

%%%%%%%%%%%%%%%%%%%%%%%%%%%%%%%%%%%%%%
%figure images of structure recognition output

%%%%%%%%%%%%%%%%%%%%%%%%%%%%%%%%%%%%%%
%images of failure of cell detection

\subsection{Ablation Study}

Table~\ref{table_ablation_study_physical_icdar_2013} shows the outcome of our enhancements to Mask {\sc r-cnn}~\cite{he2017mask} and {\sc dgcnn}~\cite{qasim2019rethinking} models for both cell detection and structure recognition networks under S-A and S-B. From the table, it can be observed that our additions to the networks result in a significant increase of 4\% average F1 scores on cell detection and structure recognition tasks. The novel alignment loss, along with the use of top-down and bottom-up pathways in the {\sc fpn} results in an improvement of 2.3\% F1 score for cell detection and 2.4\% on structure recognition. Use of {\sc lstm}s and P2 layer output to prepare visual features for structure recognition results in a 2.1\% improvement of F1 scores. Interestingly, because both models are trained together in an end-to-end fashion, cell detection's effect is also observed in the form of a 1.5\% average F1 score. This empirically bolsters our claim of using an end-to-end architecture for cell detection and, in turn, structure recognition.

%%%%%%%%%%%%%%%%%%%%%%%%%%%%%%%%%%%%%%%%%%%%%
%Ablation Study results on icdar-2013 complete dataset
\begin{table}[ht!]
\addtolength{\tabcolsep}{-1.0pt}
\begin{center}
\begin{tabular}{|l|l |l |l l l| l l l|} \hline
\textbf{ES} &\textbf{CD Network} & \textbf{SR Network} &\multicolumn{3}{l|}{\textbf{CD Scores}} &\multicolumn{3}{l|}{\textbf{SR Scores}} \\ \cline{4-9}
 &  &   &\textbf{P}$\uparrow$ &\textbf{R}$\uparrow$ &\textbf{F1}$\uparrow$ &\textbf{P}$\uparrow$ &\textbf{R}$\uparrow$ &\textbf{F1}$\uparrow$ \\ \hline
&Mask {\sc r-cnn} &{\sc dgcnn} &0.885 &0.890	&0.887 &0.871 &0.860 &0.865 \\
&Mask {\sc r-cnn} &{\sc dgcnn}+P2 &0.886	&0.892 &0.889 &0.877 &0.863	&0.870 \\
&Mask {\sc r-cnn} &{\sc dgcnn}+P2+{\sc lstm} &0.898	&0.904 &0.901 &0.885 &0.879 &0.882 \\ \cline{2-9}
&Mask {\sc r-cnn}+{\sc td}+{\sc bu} &{\sc dgcnn} &0.895 &0.899 &0.897 &0.883	&0.867 &0.875 \\
S-A &Mask {\sc r-cnn}+{\sc td}+{\sc bu} &{\sc dgcnn}+{\sc p}2 &0.895 &0.901 &0.898 &0.886 &0.870 &0.878 \\ 
&Mask {\sc r-cnn}+{\sc td}+{\sc bu}	&{\sc dgcnn}+{\sc p}2+{\sc lstm} &0.904 &0.910 &0.907	&0.892 &0.884 &0.888 \\ \cline{2-9}
&Mask {\sc r-cnn}+{\sc td}+{\sc bu}+{\sc al} &{\sc dgcnn} &0.905	&0.911 &0.908 &0.891 &0.879&	0.885 \\	
&Mask {\sc r-cnn}+{\sc td}+{\sc bu}+{\sc al} &{\sc dgcnn}+{\sc p}2 &0.914 &0.920 &0.917 &0.906	&0.885 &0.895 \\
&Mask {\sc r-cnn}+{\sc td}+{\sc bu}+{\sc al} &{\sc dgcnn}+{\sc p}2+{\sc lstm} &\textbf{0.921} &\textbf{0.926} &\textbf{0.924} &\textbf{0.915} &\textbf{0.897} &\textbf{0.906} \\ \hline
 & -{\sc na}- &{\sc dgcnn} & -{\sc na}- & -{\sc na}-  & -{\sc na}- &0.972 &0.983	&0.977 \\ 
S-B & -{\sc na}- &{\sc dgcnn}+{\sc p}2  & -{\sc na}- & -{\sc na}- & -{\sc na}- &0.973 &0.983	&0.978 \\
  & -{\sc na}-  &{\sc dgcnn}+{\sc p}2+{\sc lstm} & -{\sc na}- & -{\sc na}- & -{\sc na}- &\textbf{0.976}	&\textbf{0.985} &\textbf{0.981} \\ \hline 
\end{tabular}
\end{center}
\caption{Ablation study for physical structure recognition on {\sc icdar}-2013 dataset. \textbf{ES:} Experimental Setup, \textbf{CD:} Cell Detection, \textbf{SR:} Structure Recognition, \textbf{P2:} using visual features from P2 layer of the {\sc fpn} instead of using separate convolution blocks, \textbf{{\sc lstm}:} use of {\sc lstm}s to model visual features along center-horizontal and center-vertical lines for every cell, \textbf{{\sc td}+{\sc bu}:} use of Top-Down and Bottom-Up pathways in the {\sc fpn}, \textbf{AL:} addition of alignment loss as a regularizer to {\sc t}ab{\sc s}truct-{\sc n}et. \label{table_ablation_study_physical_icdar_2013}}
\vspace{-1em}
\end{table}

\section{Summary} \label{conclusion}

We formulate the problem of table structure recognition as a combination of cell detection (top-down) and structure recognition (bottom-up) tasks. For cell detection, we make a modification to the {\sc rpn} of original Mask {\sc r-cnn} and introduce a novel alignment loss function (formulated for every pair of table cells) to enforce structural constraints. For structure recognition, we improve input representation for the {\sc dgcnn} network by using {\sc lstm}, pre-trained ResNet-101 backbone and {\sc rpn} of cell detection network. Further, we propose an end-to-end trainable architecture to collectively predict cell bounding boxes along with their row and column adjacency matrices to predict structure. We demonstrate our results on multiple public datasets on both digital scanned as well as archival handwritten table images. We observe that our approach fails to handle tables containing a large number of empty cells along both horizontal and vertical directions. In conclusion, we encourage further research in this direction.

\section*{Acknowledgment}

This work is partly supported by {\sc meity}, Government of India.

\bibliographystyle{splncs}
\bibliography{egbib}

\section*{Appendix A: TabStruct-Net}

Our {\sc t}ab{\sc s}truct-{\sc n}et is a data-driven and an end-to-end trainable architecture for the prediction of table structure from a given table image, that combines top-down and bottom-up methods. As a first step, the input table image is broken down into individual cells using the cell detection network of the {\sc t}ab{\sc s}truct-{\sc n}et. We call this as the top-down step of the process. After detecting individual cells, the next step is to obtain the entire table structure by building relevant row and column associations between the detected cells. This is done using the structure recognition network of the {\sc t}ab{\sc s}truct-{\sc n}et and we call this as the bottom-up step of the process.

\subsection*{Post-processing to Get XML Output}

After the cell bounding boxes along with the row and column adjacency matrices are obtained, an {\sc xml} file is generated using an heuristic based algorithm. It works as follows:
\begin{itemize}
\item For row assignments, sort all bounding boxes by their $start_y$ coordinates, and initialize a row belonging list for every cell.
\item Assign a row belonging index (starting from 0) to the cell $c_i$ and assign the same row index to all other cells that are connected to $c_i$ in the row adjacency matrix.
\item Increment the row index and repeat the above step until all the cells are assigned at least one row belonging index.
\item For each cell, $SR$ is the minimum of indexes in the row belonging list, and $ER$ is the maximum of indexes in the row belonging list.
\item Similarly, for column assignments, sort all bounding boxes by their $start_x$ coordinates, and initialize a column belonging list for every cell.
\item Assign a column belonging index (starting from 0) to the cell $c_i$ and assign the same column index to all other cells that are connected to $c_i$ in the column adjacency matrix.
\item Increment the column index and repeat the above step until all the cells are assigned at least one column belonging index.
\item For each cell, $SC$ is the minimum of indexes in the row belonging list, and $EC$ is the maximum of indexes in the column belonging list.
\end{itemize}

We use Tesseract~\cite{smith2007overview} to extract the content of every predicted cell. Once $SR$, $ER$, $SC$ and $EC$ values (referred to as cell spanning values) and its content are obtained for every predicted cell, an {\sc xml} file is created with these cell spanning values along with bounding box coordinates (top-left and bottom-right of the cell) and its content.

\section*{Appendix B: Experiments}

\subsection*{Dataset}

We use various benchmark table structure recognition datasets --- {\sc s}ci{\sc tsr}~\cite{chi2019complicated}, {\sc s}ci{\sc tsr}-{\sc comp}~\cite{chi2019complicated}, {\sc icdar}-2013 table recognition~\cite{gobel2013icdar}, {\sc icdar}-2019 c{\sc td}a{\sc r} archival~\cite{gao2019icdar}, {\sc unlv}~\cite{shahab2010open}, Marmot extended~\cite{paliwal2019tablenet}, {\sc t}able{\sc b}ank~\cite{li2019tablebank} and {\sc p}ub{\sc t}ab{\sc n}et~\cite{zhong2019image} datasets for extracting structure information of tables. Statistics of these datasets are listed in Table 1.

Our {\sc t}ab{\sc s}truct-{\sc n}et makes an assumption that all cells belonging to the same column are aligned with respect to x coordinates and cells belonging to the same row are aligned with respect to y coordinates. {\sc s}ci{\sc tsr}~\cite{chi2019complicated}, {\sc s}ci{\sc tsr}-{\sc comp}~\cite{chi2019complicated} and {\sc icdar}-2013~\cite{gobel2013icdar} datasets have ground truth bounding boxes at the level of cell's content (box is the smallest rectangular block that encapsulates the cell's content). To handle this, we expand the bounding boxes of every cell in a row and column to get maximum sized content-level box in a particular row and column. 

%%%%%%%%%%%%%%%%%%%%%%%%%%%%%%%%%%%%%%%%%%%%%%%%%%%%
%%%%%%%%%% figure for unify ground truth %%%%%%%%%%%
\begin{figure}[ht!]
\begin{center}
\fbox{
\includegraphics[width=0.29\linewidth, height=0.2\linewidth]{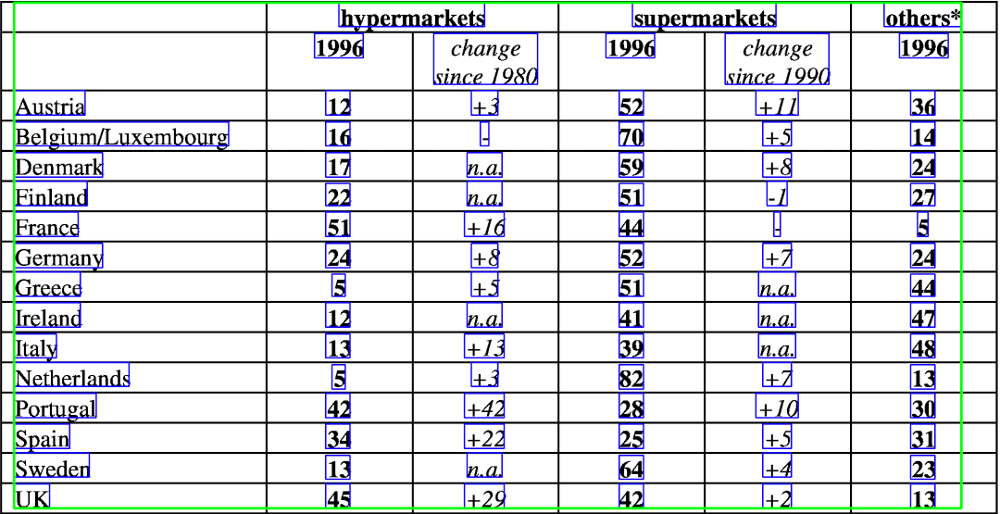}}
\hspace{-0.01\textwidth}
\fbox{
\includegraphics[width=0.29\linewidth, height=0.2\linewidth]{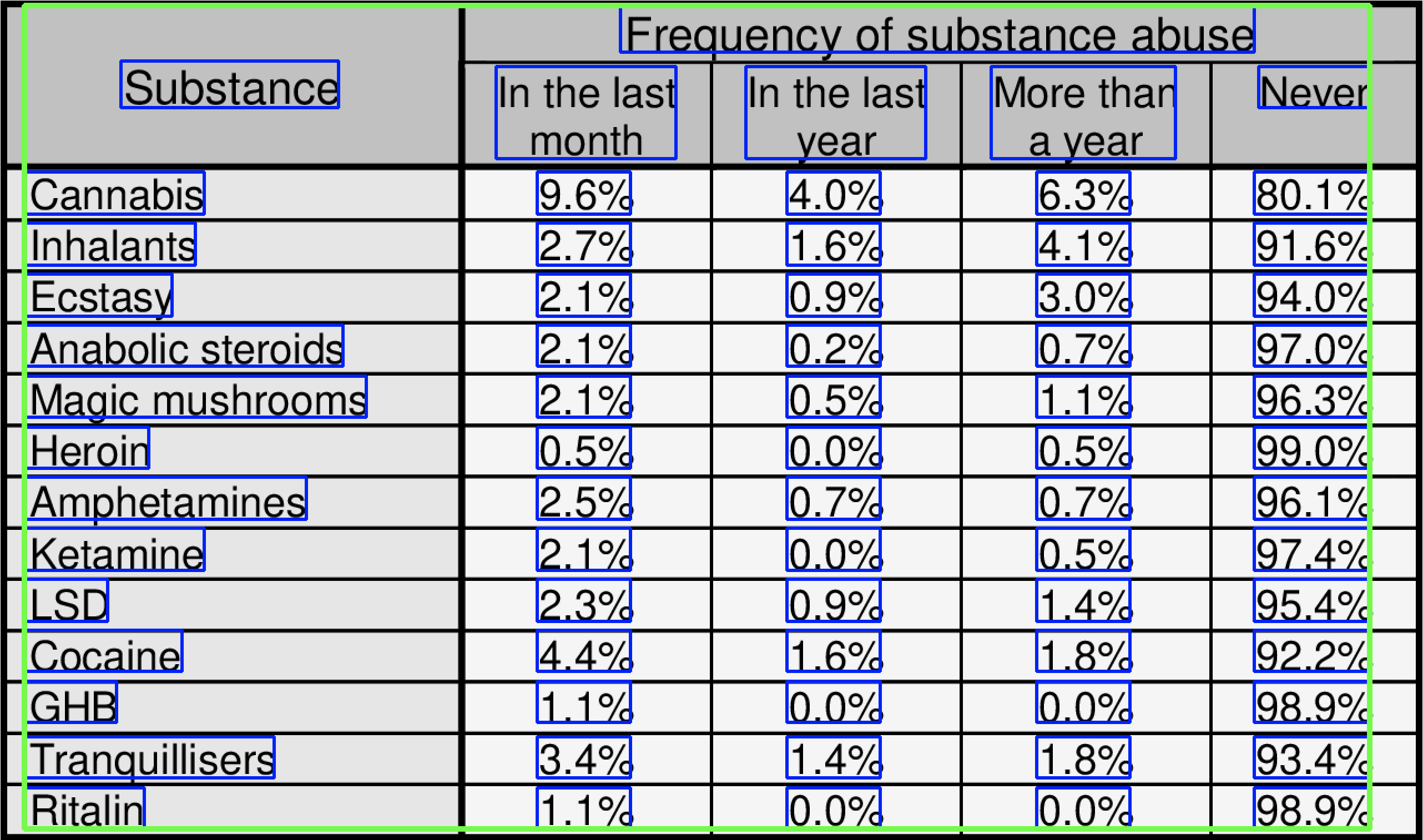}}
\hspace{-0.01\textwidth}
\fbox{
\includegraphics[width=0.29\linewidth, height=0.2\linewidth]{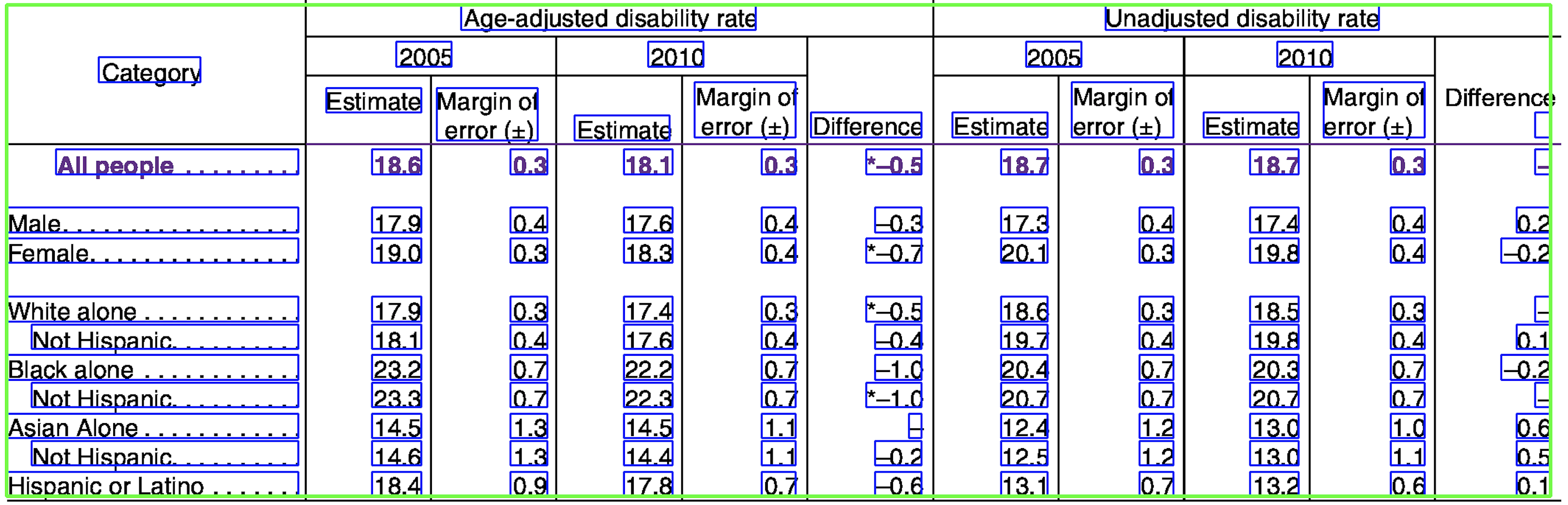}}
\vspace{0.001\textwidth}
\fbox{
\includegraphics[width=0.29\linewidth, height=0.2\linewidth]{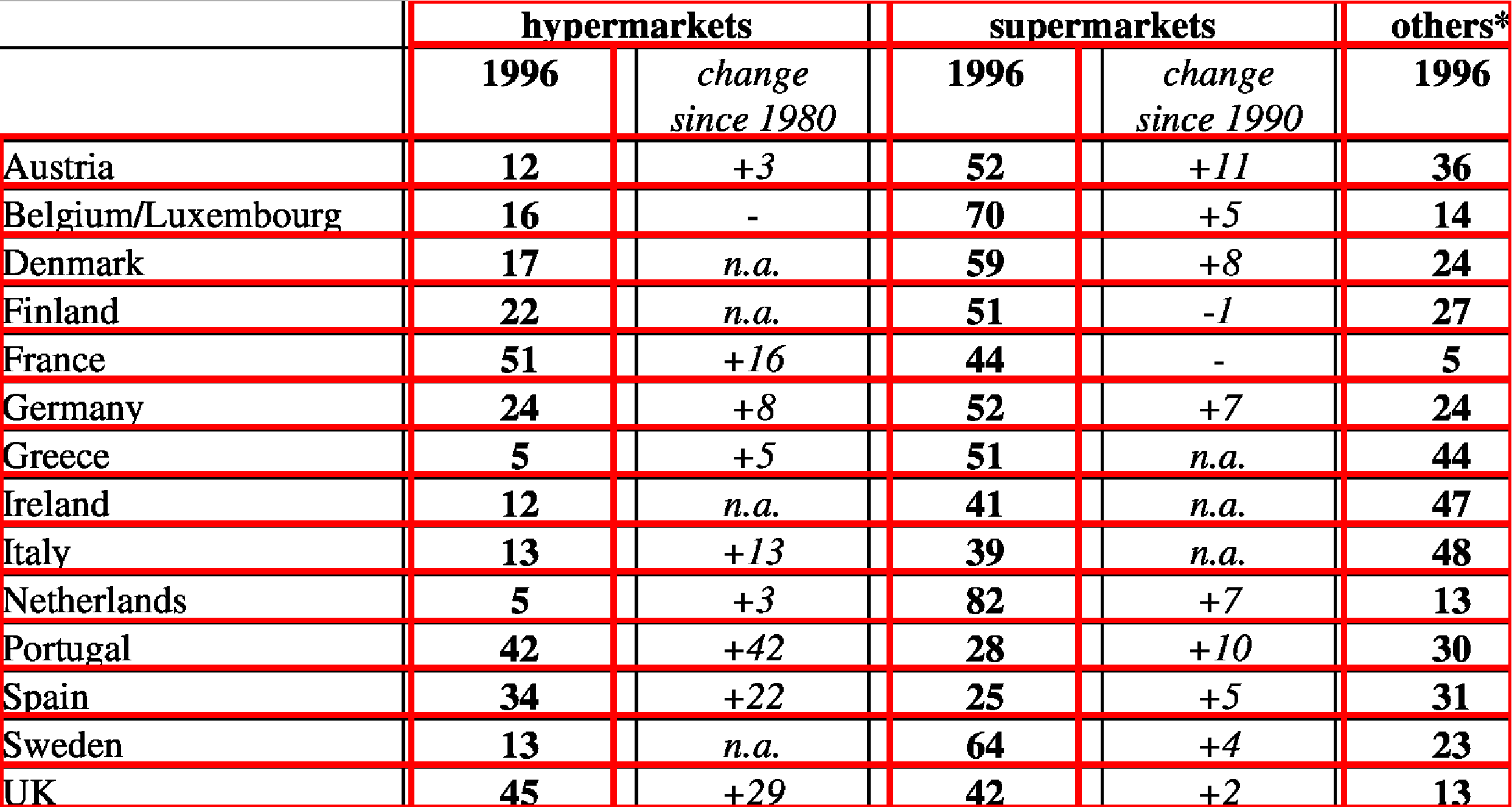}}
\hspace{-0.01\textwidth}
\fbox{
\includegraphics[width=0.29\linewidth, height=0.2\linewidth]{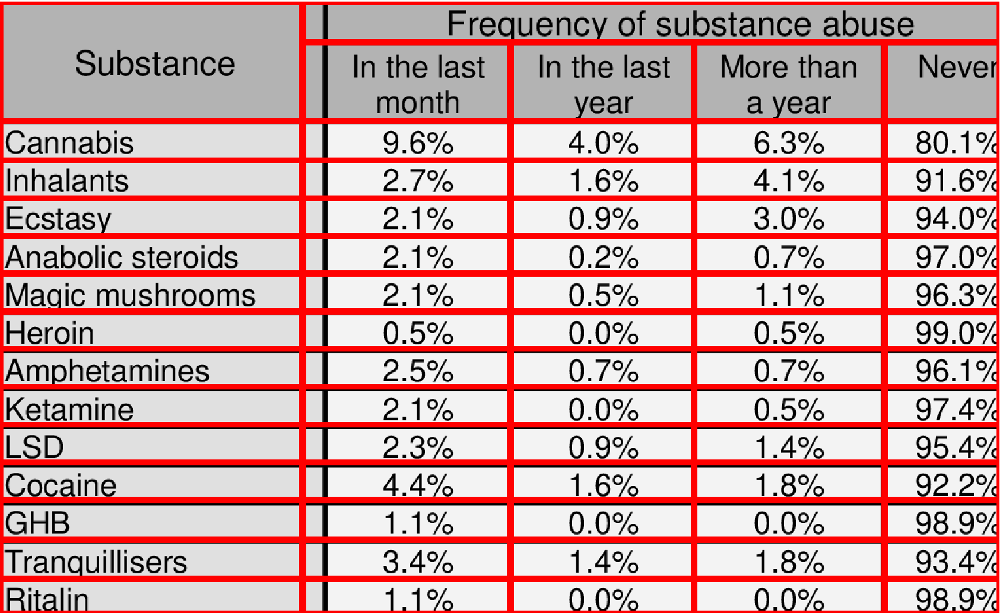}}
\hspace{-0.01\textwidth}
\fbox{
\includegraphics[width=0.29\linewidth, height=0.2\linewidth]{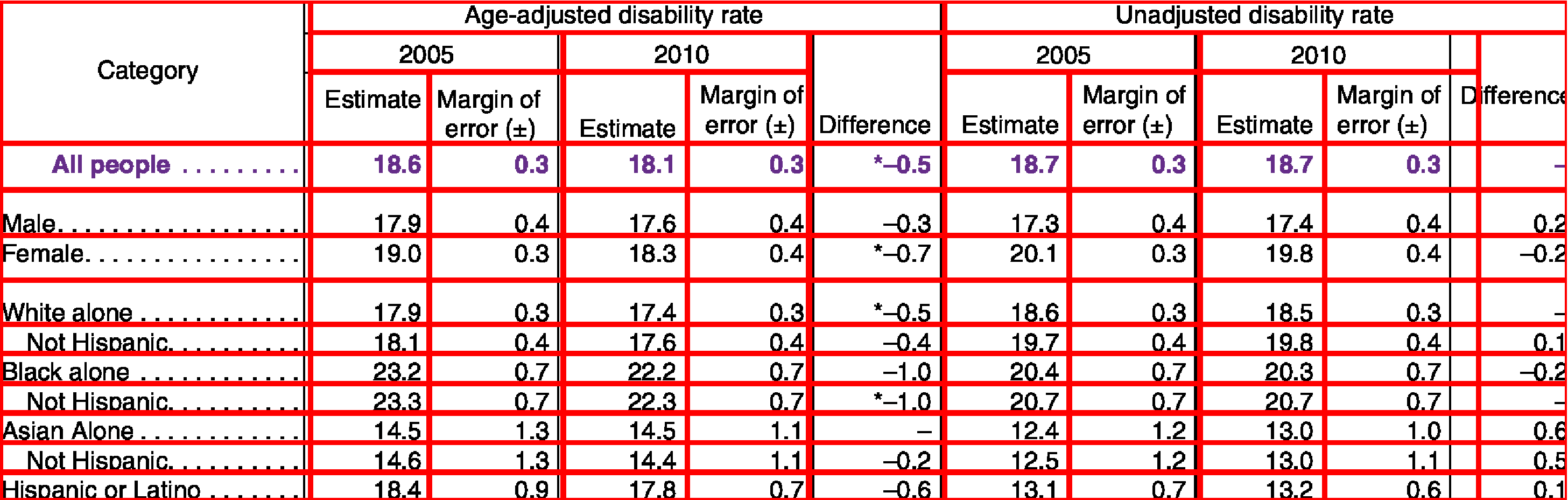}}
\end{center}
\caption{Ground truth unification. content-level bounding boxes are given in ground truth as shown in \textbf{First Row}. We make content-level bounding boxes into cell-level bounding boxes as shown in \textbf{Second Row}.} \label{fig_unify_gt}
\end{figure}

\subsection*{Baseline Methods} \label{baseline_methods_extra}
We compare the performance of our {\sc t}ab{\sc s}truct-{\sc n}et against seven following benchmark methods.

\paragraph{\textbf{DeepDeSRT~\cite{schreiber2017deepdesrt}:}} This method leverages semantic segmentation approach to localize each row and column from the table image. This method outputs table as a grid-like structure and fails to identify multiple row and multiple column spanning cells. Since no code is available, we implement our own version of this method. Since this method works by predicting every row and column, the {\sc s}ci{\sc tsr} training dataset is pre-processed to obtain row and column level coordinates before training.
    
\paragraph{\textbf{TableNet~\cite{paliwal2019tablenet}:}} This method uses semantic segmentation approach to extract table and column masks, and segments rows by identifying words present in different columns (extracted using {\sc t}esseract {\sc ocr}~\cite{smith2007overview}) that occur at the same horizontal level. For comparison against other methods, we directly use the results reported by the authors.
    
\paragraph{\textbf{GraphTSR~\cite{chi2019complicated}:}} This method consists of edge-to-vertex and vertex-to-edge graph attention blocks to compute vertex and edge representations in a latent space, and finally classify each edge as `horizontal', `vertical' or `no-relation'. It uses absolute and relative positions of every cell extracted from the {\sc pdf} to compute initial vertex and edge features.
    
\paragraph{\textbf{SPLERGE~\cite{table_splitting}:}} This method comprises of two deep learning networks that perform splitting and merging operations in sequence to predict fine grid-like table structure and to predict merged cells which span multiple rows/columns. Split method shows an improved performance when additional {\sc pdf} extracted meta-features are provided along with the table image. For the split model (to obtain the basic grid of the table), authors pre-process the ground truth by maximizing the row and column separator regions without intersecting any non multiple row or column spanning cell. For the merge model (to identify cells that span multiple rows or columns), the authors prepare the ground truth by identifying grid elements that span multiple cells and set the merging probability in the respective directions. Further, for evaluating this method on {\sc icdar}-2013~\cite{gobel2013icdar} dataset, the authors realized that merge method did not work with a good performance, and hence, introduced the following specific heuristics to merge cells instead:
\begin{itemize}
    \item Merge cells where separators pass through text.
    \item Merge adjacent blank columns with cells that have been formed by merging many cells.
    \item Merge adjacent blank rows with content cells.
    \item Split columns that have a consistent white-space gap between vertically aligned text.
\end{itemize}

\paragraph{\textbf{DGCNN$^*$~\cite{qasim2019rethinking}:}} Authors formulate the problem as a graph learning problem to predict whether every pair of words belongs to the same cell, row and/or column or not. Their architecture consists of a visual network, an interaction network and a classification network. For evaluation purposes, table image along with word-level bounding boxes is provided as inputs.
    
\paragraph{\textbf{Bi-directional GRU~\cite{Khan_2019}:}} Given the table image, this method uses two bi-directional {\sc gru}s to establish row and column boundaries in a context driven manner. This method however fails to localize multiple row and/or multiple column spanning cells.
    
\paragraph{\textbf{Image-to-Text~\cite{paliwal2019tablenet}:}} This method utilizes an Image-to-Markup model to predict a markup-like output of a given table image. It consists of a {\sc cnn} based encoder to compute visual features and an {\sc lstm} based decoder to produce markup output.

\subsection*{Implementation Details} \label{implementation_detail}

Our {\sc t}ab{\sc s}truct-{\sc n}et model has been trained and evaluated with table images scaled to a fixed size of 1536$\times$1536 while maintaining the original aspect ratio as the input. While training, cell-level bounding boxes along with row and column adjacency matrices (prepared from start-row, start-column, end-row and end-column indices) are used as the ground truth. We use {\sc nvidia titan x gpu} with 12 {\sc gb} memory for our experiments and a batch-size of 1. Instead of using 3$\times$3 convolution on the output feature maps from the {\sc fpn}, we use a dilated convolution with filter size of 2$\times$2 and dilation parameter of 2. Also, we use the {\sc r}es{\sc n}et-101 backbone that is pre-trained on {\sc ms-coco}~\cite{smith2007overview} dataset. To compute region proposals, we use 0.5, 1 and 2 as the anchor scale and anchor box sizes of 8, 16, 32, 64 and 128. Further, for generation of the row/column adjacency matrices, we use 2400 as the maximum number of vertices keeping in mind dense tables. Since every input table may contain hundreds of table cells, training can be a time consuming process. 

To achieve faster training, we employ a 2-stage training process. In the first step, we use 2014 anchors and 512 {\sc r}o{\sc i}s. With this setting, the model is able to learn high and low level features but resulted in a large number of false negatives. To combat this, network is trained with 3072 anchors and 2048 {\sc r}o{\sc i}s. This significantly reduces the number of false negatives. For the first step, we train a total of 30 epochs, for the first 8, we train all {\sc fpn} and subsequent layers, for the next 15, we train {\sc fpn} + last 4 layers of {\sc r}es{\sc n}et-101 and for the last 7 epochs, we train all the layers of the model. For the second step, we train a total of 10 epochs, for the first 3, we train all {\sc fpn} and subsequent layers, for the next 4, we train {\sc fpn} + last 4 layers of {\sc r}es{\sc n}et-101 and for the last 3 epochs, we train all the layers of the model. During both the stages, we use 0.001 as the learning rate, 0.9 as the momentum and 0.0001 as the weight decay regularisation.

\subsection*{Evaluation Measures}

\paragraph{\textbf{Details of Evaluation Criteria:}} For comparison against most of the existing methods, we use the precision, recall and F1 score~\cite{chi2019complicated,gobel2013icdar,shahab2010open}. Before evaluating performance of structure recognition, it is important to understand the specific cases in which detected cells are taken into consideration for structure recognition:
\begin{itemize}
    \item We consider a detected cell to be a true positive if it overlaps with the ground truth cell bounding box is more than 0.5.
    %having an IoU score that exceeds 0.5.
    \item During evaluation of structure recognition, cells that have no content (i.e., empty or blank cells) are discarded. It means that adjacency relations that involve a blank cell are not taken into consideration.
\end{itemize}

To evaluate the performance of structure recognition, adjacency relations between every cell (with content) are generated with their horizontal and vertical neighbors. This predicted relation list is then compared with the ground truth list to generate precision, recall and F1 measures.

As per~\cite{gobel2013icdar}, this method accounts for the standard evaluation measures for table structure recognition for the following reasons:
\begin{itemize}
    \item It provides for a simple way to account for errors in the scenarios of complex table layouts containing blank cells, and cells that span multiple rows and/or columns.
    \item It accounts for evaluation of both physical as well as logical structure prediction methods as it is not dependent on the bounding box coordinates information.
\end{itemize}

\subsection*{Experimental Setup} \label{experimental_setup_extra}

One major challenge in the comparison study with the existing methods is the inconsistent use of additional information (e.g., meta-features extracted from the {\sc pdf}s~\cite{table_splitting}, content-level bounding boxes from ground truths~\cite{paliwal2019tablenet,chi2019complicated} and cell's location features generated from synthetic dataset~\cite{qasim2019rethinking}). 

For the unification of fair comparison for table structure recognition, we divide all inconsistencies into several levels - (i) inconsistency with respect to input modalities, (ii) inconsistency with respect to annotation levels, (iii) inconsistency with respect to representation of table structure, (iv) inconsistency with respect to evaluation methods, and (v) inconsistency with respect to way of computing evaluation scores.

\paragraph{\textbf{Inconsistency with respect to input modalities:}} Section 3.2 describes that many methods for table structure recognition work with table images alone~\cite{schreiber2017deepdesrt,table_splitting,Khan_2019}, several other assume additional information in the form of meta-features or bounding boxes around every word or cell-content extracted from the {\sc pdf}s~\cite{qasim2019rethinking,paliwal2019tablenet,chi2019complicated}. This makes it difficult to compare these methods under a unified scenario. To take care of this problem, we define two different experimental setups - (a) \textbf{Setup-A (S-A)} where only table image is used as an input to the structure recognition model and (b) \textbf{Setup-B (S-B)} where table image along with additional meta-features such as low-level content bounding boxes are used as an input to the structure recognition model. We present our results under both the experimental setups for a thorough comparison of our work against most of the recent methods in this space. To achieve this, we train our model for cell detection as well as structure recognition collectively for S-A. For evaluation in S-B, instead of predicting cell bounding boxes from the image, we use the table image and the low-level bounding box information from {\sc ocr} or ground truth to be able to directly and fairly compare our method.
    
\paragraph{\textbf{Inconsistency with respect to annotation levels:}} It is important to note that training of {\sc t}ab{\sc s}truct-{\sc n}et assumes cell-level bounding boxes in a way that all cells that (a) have the same $SR$ indices having same $y1$ coordinates, (b) have the same $SC$ indices having same $x1$ coordinates, (c) have the same $ER$ indices having same $y2$ coordinates, and (d) have the same $EC$ indices having same $x2$ coordinates. This assumption is necessary for our alignment loss function to work properly. However, different datasets for physical table structure recognition have ground truth annotations defined in different ways. {\sc unlv} and {\sc icdar}-2019 archival datasets have ground truth annotated at the cell-level. {\sc s}ci{\sc tsr}~\cite{chi2019complicated} and {\sc icdar}-2013~\cite{gobel2013icdar} datasets have ground truth annotation defined at the content-level (cells' bounding box is the smallest rectangle that covers entire content of the cell). To be able to use those for training, we pre-process the ground-truth to obtain cell-level bounding boxes (as explained in Section 3.1). Please note that this pre-processing step is only done for the training process. Similarly, ground-truth bounding boxes of the synthetic dataset proposed in~\cite{qasim2019rethinking} are provided at the word-level. To obtain cell-level bounding boxes, we use the ground-truth cell adjacency matrix and word-level bounding boxes to obtain content-level bounding boxes. During the testing time in S-A, however, to compute if a detected cell is a true positive, we use the original ground-truth bounding boxes (either at cell-level or content-level), and not the pre-processed ones. Similarly while testing in S-B, we use the original content-level or cell-level bounding boxes as the additional input instead of the pre-processed ones. This ensures consistency while comparing our methods against previously published ones.
    
\paragraph{\textbf{Inconsistency with respect to representation of table structure:}} We broadly classify table structure methods into two categories - (a) physical structure predicting methods that predict cell-level bounding boxes along with their associations~\cite{schreiber2017deepdesrt,table_splitting,Khan_2019} and (b) logical structure predicting methods~\cite{qasim2019rethinking,li2019tablebank,paliwal2019tablenet,chi2019complicated} that predict only cell associations. In our work, we standardize our representation as described in Section 3.5, which allows us to directly compare methods in both the experimental setups. To compare the results of {\sc t}ab{\sc s}truct-{\sc n}et on logical structure prediction, we generate the mark-up string from the post-processed {\sc xml} output of {\sc t}ab{\sc s}truct-{\sc n}et in the same format as {\sc t}able{\sc B}ank~\cite{li2019tablebank} and {\sc p}ub{\sc t}ab{\sc n}et~\cite{zhong2019image} by extracting only the structure without cells' coordinates and content.

\paragraph{\textbf{Inconsistency with respect to evaluation methods:}} While most of the previously published methods for table structure recognition use precision, recall and F1 scores as described in~\cite{gobel2013icdar}, there are some inconsistencies in this aspect as well. In~\cite{qasim2019rethinking}, authors use true positive rate ({\sc tpr}), false positive rate ({\sc fpr}) and absolute accuracy on the predicted adjacency matrix to compute performance. In order to standardize evaluation with~\cite{qasim2019rethinking}, we infer neighboring cell relations from their output to ensure consistency. Further,~\cite{paliwal2019tablenet} use {\sc bleu} scores to compare their output with the ground truth. Since our method generates and {\sc xml} output from the model's predictions, we bring our output to the same format as~\cite{paliwal2019tablenet} to ensure a direct and fair comparison on the {\sc t}able{\sc b}ank dataset~\cite{li2019tablebank}.

\paragraph{\textbf{Inconsistency with respect to way of computing evaluation scores:}} To fairly compare {\sc t}ab{\sc s}truct-{\sc n}et against previous methods, we list both micro averaged results on the test datasets. However, it is important to note that for {\sc t}able{\sc B}ank~\cite{li2019tablebank} and {\sc p}ub{\sc t}ab{\sc n}et~\cite{zhong2019image} datasets, where we evaluate our results on the markup output of the model, we only consider document-averaged results.

\section*{Appendix C: Results on Table Structure Recognition}

%%%%%%%%%%%%%%%%%%%%%%%%%%%%%%%%%%%%%%%%%%%%%%%%%%
\subsection*{Micro-averaged Results}

Tables~\ref{table_physical_icdar_2013_partial}-\ref{table_physical_scistr} show the micro-averaged results of various methods for structure recognition on multiple datasets. From the tables, it can be observed that our method outperforms previously published works under multiple kinds of experimental settings. Further, it is important to note that the tables use an IoU threshold of 0.6 to identify true positive cells for experiment setup S-A. We also show the precision, recall and F1 measures on various IoU thresholds to better interpret the performance of the cell detection module of {\sc t}ab{\sc s}truct-{\sc n}et.

%%%%%%%%%%%%%%%%%%%%%%%%%%%%%%%%%%%%%%%%%%%%%
%quantitative results on icdar-2013-partial dataset
\begin{table}
%\addtolength{\tabcolsep}{-1.0pt}
\begin{center}
\begin{tabular}{|l |l| r| l |c c c|} \hline
\textbf{Method} &\multicolumn{2}{|c|}{\textbf{Training}} &\textbf{Exp.} &\textbf{P}$\uparrow$ &\textbf{R}$\uparrow$ &\textbf{F1}$\uparrow$ \\ \cline{2-3}
  &\textbf{Dataset} &\textbf{\#Images} &\textbf{Setup} & & & \\ \hline
{\sc d}eep{\sc d}e{\sc srt}~\cite{schreiber2017deepdesrt} &{\sc icdar}-2013-partial  &0.124K &S-A &0.959 &0.874 &0.914 \\ 
{\sc splerge}~\cite{table_splitting} &{\sc icdar}-2013-partial  &0.124K &S-A &0.917 &0.911 &0.914 \\ 
Bi-directional {\sc gru}~\cite{Khan_2019} &{\sc icdar}-2013-partial &0.124K &S-A &\textbf{0.969} &\textbf{0.901} &\textbf{0.934} \\ 
 {\sc t}ab{\sc s}truct-{\sc n}et (our) &{\sc icdar}-2013-partial &0.124K &S-A &0.928  &0.903 &0.915 \\   
{\sc t}ab{\sc s}truct-{\sc n}et (our) &{\sc s}ci{\sc tsr} &12.124K &S-A &0.930  &0.908  &0.919 \\ 
 &+ {\sc icdar}-2013-partial & & & & & \\ \hline 
{\sc t}able{\sc n}et~\cite{paliwal2019tablenet} &Marmot extended &1.016K &S-B &0.931 &0.900 &0.915 \\
{\sc g}raph{\sc tsr}~\cite{chi2019complicated} &{\sc s}ci{\sc tsr} &12.124K &S-B &0.854  &0.891  &0.872 \\   
  &+ {\sc icdar}-2013-partial & & & & &  \\ 
{\sc dgcnn}~\cite{qasim2019rethinking} &{\sc s}ci{\sc tsr} &12.124K &S-B &0.986  &0.990  &0.988 \\
 &+ {\sc icdar}-2013-partial & & & & &  \\
 {\sc t}ab{\sc s}truct-{\sc n}et (our) &{\sc icdar}-2013-partial &0.124K &S-B &\textbf{0.991}  &0.989 &0.990 \\   
{\sc t}ab{\sc s}truct-{\sc n}et (our) &{\sc s}ci{\sc tsr} &12.124K &S-B &\textbf{0.991}  &\textbf{0.993}  &\textbf{0.992} \\   
 &+ {\sc icdar}-2013-partial & & & & &  \\ \hline
\end{tabular}
\end{center}
\caption{Comparison of results for physical structure recognition on {\sc icdar}-2013-partial dataset. \textbf{P:} indicates precision, \textbf{R:} indicates recall, \textbf{F1:} indicates F1 Score and \textbf{\#Images:} indicates number of table images in the training set. \label{table_physical_icdar_2013_partial}}
\end{table}

%%%%%%%%%%%%%%%%%%%%%%%%%%%%%%%%%%%%%%%%%%%%%
%quantitative results on ICDAR-2019 dataset
\begin{table}
%\addtolength{\tabcolsep}{-1.3pt}
\begin{center}
\begin{tabular}{|l | l| r| l |c c c|} \hline
\textbf{Method} &\multicolumn{2}{|c|}{\textbf{Training}} &\textbf{Exp.} &\textbf{P}$\uparrow$ &\textbf{R}$\uparrow$ &\textbf{F1}$\uparrow$ \\ \cline{2-3}
  &\textbf{Dataset} &\textbf{\#Images} &\textbf{Setup} & & & \\ \hline
{\sc nlpr-pal} [19] &c{\sc td}a{\sc r} &0.6K &S-A &0.720 &0.770 &0.745 \\ 
{\sc dgcnn}~\cite{qasim2019rethinking} &c{\sc td}a{\sc r} &0.6K &S-A &0.785 &0.751 &0.768 \\
{\sc dgcnn}~\cite{qasim2019rethinking} &{\sc s}ci{\sc tsr} &12.0K &S-A &0.552 &0.519 &0.535 \\
{\sc dgcnn}~\cite{qasim2019rethinking} &c{\sc td}a{\sc r} + {\sc s}ci{\sc tsr} &12.6K &S-A &0.803 &0.778 &0.790 \\
{\sc splerge}~\cite{table_splitting} &c{\sc td}a{\sc r} &0.6K &S-A &0.774 &0.783 &0.778 \\
{\sc splerge}~\cite{table_splitting} &{\sc s}ci{\sc tsr} &12.0K &S-A &0.559 &0.572 &0.565 \\
{\sc splerge}~\cite{table_splitting} &c{\sc td}a{\sc r} + {\sc s}ci{\sc tsr} &12.6K &S-A &0.792 &0.800 &0.796 \\
{\sc t}ab{\sc s}truct-{\sc n}et (our) &c{\sc td}a{\sc r} &0.6K &S-A &0.803 &0.768 &0.785 \\
{\sc t}ab{\sc s}truct-{\sc n}et (our) &{\sc s}ci{\sc tsr} &12.0K &S-A &0.595 &0.572 &0.583 \\
{\sc t}ab{\sc s}truct-{\sc n}et (our) &c{\sc td}a{\sc r} + {\sc s}ci{\sc tsr} &12.6K &S-A &\textbf{0.822} &\textbf{0.787} &\textbf{0.804} \\ \hline 
\end{tabular}
\end{center}
\caption{Comparison of results for physical structure recognition on {\sc icdar}-2019 (c{\sc td}a{\sc r}) archival dataset. For comparison against {\sc dgcnn}~\cite{qasim2019rethinking}, we use the cell bounding boxes detected from {\sc t}ab{\sc s}truct-{\sc n}et for a fair comparison. \textbf{P:} indicates precision, \textbf{R:} indicates recall, \textbf{F1:} indicates F1 Score and \textbf{\#Images:} indicates number of table images in the training set.
\label{table_physical_icdar_2019}}
\end{table}

%%%%%%%%%%%%%%%%%%%%%%%%%%%%%%%%%%%%%%%%%%%%%
%%quantitative results on UNLV-partial dataset
\begin{table}
%\addtolength{\tabcolsep}{-1.0pt}
\begin{center}
\begin{tabular}{|l |l |l l l |} \hline
\textbf{Method} &\textbf{Exp. Setup} &\textbf{P}$\uparrow$ &\textbf{R}$\uparrow$ &\textbf{F1}$\uparrow$ \\ \hline
{\sc d}eep{\sc d}e{\sc srt}~\cite{schreiber2017deepdesrt} &S-A &0.554 &0.529 &0.541 \\
{\sc splerge}~\cite{table_splitting} &S-A &0.795 &0.776 &0.785 \\
{\sc t}ab{\sc s}truct-{\sc n}et (our) &S-A &\textbf{0.849} &\textbf{0.828} &\textbf{0.839} \\ \hline
{\sc g}raph{\sc tsr}~\cite{chi2019complicated} &S-B &0.763 &0.786 &0.774 \\
{\sc dgcnn}~\cite{qasim2019rethinking} &S-B &0.921 &0.898 &0.909 \\
{\sc t}ab{\sc s}truct-{\sc n}et (our) &S-B &\textbf{0.992} &\textbf{0.994} &\textbf{0.993} \\ \hline
\end{tabular}
\end{center}
\caption{Comparison of results for physical structure recognition on {\sc unlv}-partial dataset. \textbf{P:} indicates precision, \textbf{R:} indicates recall, \textbf{F1:} indicates F1 Score. All models are trained on {\sc s}ci{\sc tsr} and fine-tuned on {\sc unlv}-partial datasets.} \label{table_physical_unlv}
\end{table}

%%%%%%%%%%%%%%%%%%%%%%%%%%%%%%%%%%%%%%%%%%%%%
%quantitative results on SciTSR dataset
\begin{table}
%\addtolength{\tabcolsep}{-1.0pt}
\begin{center}
\begin{tabular}{|l |l|l l l|l l l|} \hline
\textbf{Method} &\textbf{Exp.} &\multicolumn{6}{|l|}{\textbf{Evaluation on}} \\ \cline{3-8}
 &\textbf{Setup} &\multicolumn{3}{|l|}{\textbf{SciTSR}} &\multicolumn{3}{|l|}{\textbf{SciTSR-COMP}} \\ \cline{3-8}
 & &\textbf{P}$\uparrow$ &\textbf{R}$\uparrow$ &\textbf{F1}$\uparrow$ &\textbf{P}$\uparrow$ &\textbf{R}$\uparrow$ &\textbf{F1}$\uparrow$ \\ \hline
{\sc d}eep{\sc d}e{\sc srt}~\cite{schreiber2017deepdesrt} &S-A &0.906 &0.887 &0.890 &0.863 &0.831 &0.846 \\ 
{\sc splerge}~\cite{table_splitting} &S-A &0.922 &\textbf{0.915} &0.918 &\textbf{0.911} &0.880 &\textbf{0.895} \\ 
{\sc t}ab{\sc s}truct-{\sc n}et (our) &S-A &\textbf{0.927} &0.913 &\textbf{0.920} &0.909 &\textbf{0.882} &\textbf{0.895} \\ \hline
{\sc g}raph{\sc tsr}~\cite{chi2019complicated} &S-B &0.959 &0.948 &0.953 &0.964 &0.945 &0.955 \\
{\sc dgcnn}~\cite{qasim2019rethinking} &S-B &0.970 &0.981 &0.976 &0.963 &0.974 &0.969 \\
{\sc t}ab{\sc s}truct-{\sc n}et (our) &S-B &\textbf{0.989} &\textbf{0.993} &\textbf{0.991} &\textbf{0.981} &\textbf{0.987} &\textbf{0.984} \\ \hline 
\end{tabular}
\end{center}
\caption{Comparison of results for physical structure recognition on {\sc s}ci{\sc tsr} and {\sc s}ci{\sc tsr-comp} datasets. \textbf{P:} indicates precision, \textbf{R:} indicates recall, \textbf{F1:} indicates F1 Score. All the models are trained on {\sc s}ci{\sc tsr} dataset.} \label{table_physical_scistr}
\vspace{-1.5em}
\end{table}

%%%%%%%%%%%%%%%%%%%%%%%%%%%%%%%%%%%%%%%%%%%%%%%%%%%%%%%
\newpage
\subsection*{Average Results on Markup Output}

Tables~\ref{table_logical_tablebank}-\ref{table_logical_pubtabnet} present compare our results for logical structure prediction from the table image on {\sc t}able{\sc b}ank and {\sc p}ub{\sc t}ab{\sc n}et dataset, respectively. The scores are obtained by averaging the score for every table across all the tables in the evaluation dataset. From the tables, it can be inferred that despite trained with a much smaller set of data, our model achieves better performance than~\cite{zhong2019image}. Direct comparison, however would not be fair because of the use of different input modalities for training.

%%%%%%%%%%%%%%%%%%%%%%%%%%%%%%%%%%%%%%%%%%%%%
%quantitative results on TableBank dataset
\begin{table}
\begin{center}
\begin{tabular}{|l | l|l| r| l |c c c|} \hline
\textbf{Method} &\multicolumn{3}{|l|}{\textbf{Training Set}}  &\textbf{Experimental} &\multicolumn{3}{|l|}{\textbf{BLEU}$\uparrow$} \\ \cline{2-4}\cline{6-8}
   &\textbf{Dataset} &\textbf{Type} &\textbf{\#Images} &\textbf{Setup} &\textbf{Word} &\textbf{Latex} &\textbf{Both} \\ \hline
Image-to-Text~\cite{paliwal2019tablenet} &{\sc t}able{\sc b}ank &Word &55.866K &S-A &0.751 &0.673 &0.7138 \\
Image-to-Text~\cite{paliwal2019tablenet} &{\sc t}able{\sc b}ank &Latex &87.597K &S-A &0.405 &0.765 &0.582 \\
Image-to-Text~\cite{paliwal2019tablenet} &{\sc t}able{\sc b}ank &Both &144.493K &S-A &0.712 &0.765 &0.738 \\
{\sc t}ab{\sc s}truct-{\sc n}et (our) &{\sc s}ci{\sc tsr} &Image &12K &S-A &\textbf{0.914} &\textbf{0.937} &\textbf{0.916} \\ \hline 
\end{tabular}
\end{center}
\caption{Comparison of results for logical structure recognition on {\sc t}able{\sc b}ank dataset.\label{table_logical_tablebank}}
\vspace{-1.5em}
\end{table}

%%%%%%%%%%%%%%%%%%%%%%%%%%%%%%%%%%%%%%%%%%%%%%%%
%quantitative results on PubLatNet dataset
\begin{table}
%\addtolength{\tabcolsep}{-1.0pt}
\begin{center}
\begin{tabular}{|l |l |l |l |l |} \hline
\textbf{Method} &\textbf{Experimental Setup} &\textbf{Training Dataset} &\textbf{\#Images}&\textbf{TEDS}$\uparrow$ \\ \hline
Acrobat Pro~\cite{zhong2019image}  &S-A & - & - &0.537 \\
{\sc wygiwys}~\cite{zhong2019image} &S-A &{\sc p}ub{\sc t}ab{\sc n}et &399K &0.786 \\
{\sc edd}~\cite{zhong2019image} &S-A &{\sc p}ub{\sc t}ab{\sc n}et &399K &0.883 \\
{\sc t}ab{\sc s}truct-{\sc n}et (our) &S-A &{\sc s}ci{\sc tsr}~\cite{chi2019complicated} &12K &\textbf{0.901} \\ \hline
\end{tabular}
\end{center}
\caption{Comparison of results for logical structure recognition on {\sc p}ub{\sc t}ab{\sc n}et dataset~\cite{zhong2019image}. \textbf{TEDS:} indicates averaged tree edit distance based similarity~\cite{zhong2019image}.} \label{table_logical_pubtabnet}
\end{table}

%table for table understanding 
\begin{table}[ht!]
\addtolength{\tabcolsep}{-1.5pt}
\begin{center}
\begin{tabular}{|l |l | l |c c c|} \hline
\textbf{Training Set} &\textbf{Evaluation Set} &\textbf{Model} &\textbf{BLEU}$\uparrow$ &\textbf{CIDEr}$\uparrow$ &\textbf{ROUGE}$\uparrow$ \\ \hline 
{\sc s}ci{\sc tsr} &{\sc s}ci{\sc tsr} &{\sc dgcnn} & 0.774 &0.8 &0.782 \\ \cline{3-6}
 &    &{\sc t}ab{\sc s}truct-{\sc n}et &0.833 &0.848 &0.839 \\ \hline  
{\sc s}ci{\sc tsr} &{\sc s}ci{\sc tsr-comp} &{\sc dgcnn} &0.769 &0.795 &0.774 \\ \cline{3-6} 
 &  &{\sc t}ab{\sc s}truct-{\sc n}et &0.826 &0.837 &0.830 \\ \hline
{\sc s}ci{\sc tsr} + {\sc unlv}-partial &{\sc unlv}-partial &{\sc dgcnn} &0.721 &0.744 &0.729 \\ \cline{3-6} 
 &  &{\sc t}ab{\sc s}truct-{\sc n}et &0.804 &0.826 &0.813 \\ \hline
 {\sc s}ci{\sc tsr}  &{\sc icdar-2013} &{\sc dgcnn} &0.756 &0.773 &0.762 \\ \cline{3-6} 
 &  &{\sc t}ab{\sc s}truct-{\sc n}et &0.815 &0.831 &0.821 \\ \hline
 {\sc s}ci{\sc tsr} + {\sc icdar-2013}-partial  &{\sc icdar-2013-} &{\sc dgcnn} &0.772 &0.801 &0.78 \\ \cline{3-6} 
 &partial  &{\sc t}ab{\sc s}truct-{\sc n}et &0.829 &0.845 &0.834 \\ \hline
 \end{tabular}
\end{center}
\caption{Results comparison of various methods for table structure recognition on various datasets.} \label{table_table_understanding}
%\vspace{-2em}
\end{table}

We present our results on the output {\sc xml} file that contains --- (a) bounding box coordinates, (b) start and end row indices, (c) start and end column indices, and (d) content for every predicted cell given the table image. To evaluate our method, we compare this {\sc xml} against the ground-truth prepared in the same format using {\sc bleu}, {\sc cidr}r and {\sc rouge} scores as presented in Table~\ref{table_table_understanding}. The table also compares our results against {\sc dgcnn}~\cite{qasim2019rethinking} when cells detected from {\sc t}ab{\sc s}truct-{\sc n}et are provided as the input to their model.

%%%%%%%%%%%%%%%%%%%%%%%%%%%%%%%%%%%%%%%%%%%%%%%%%%%%%%%

%%%%%%%%%%%%%%%%%%%%%%%%%%%%%%%%%%%%%%%%%%%%%%%%%
\newpage
\subsection*{Qualitative Results of Cell Detection}

\begin{figure}[ht!]
\begin{center}
%%%%%%%%%%%%%%%%%%%%%%%%%%%%%%%%%%%%%%
%cell detection of ICDAR-2013
\fbox{
\includegraphics[width=0.29\linewidth, height=0.2\linewidth]{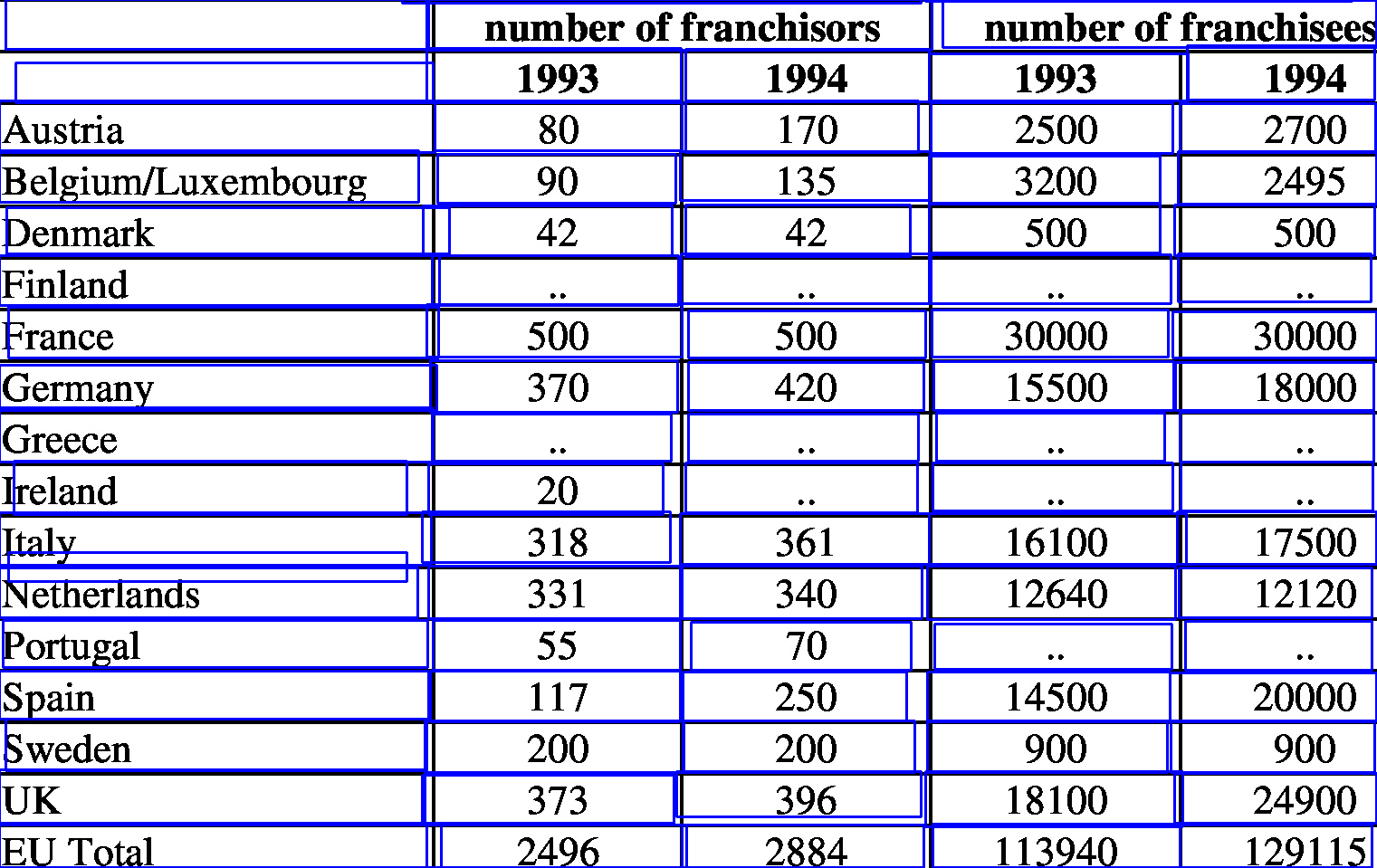}}
\hspace{-0.01\textwidth}
\fbox{
\includegraphics[width=0.29\linewidth, height=0.2\linewidth]{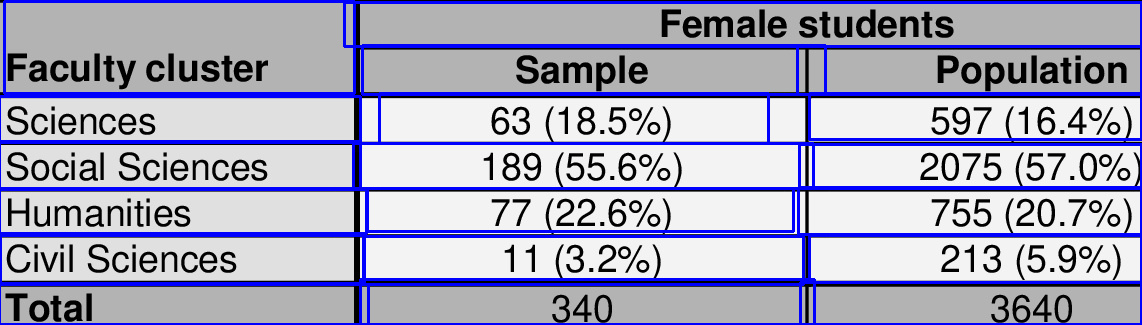}}
\hspace{-0.01\textwidth}
\fbox{
\includegraphics[width=0.29\linewidth, height=0.2\linewidth]{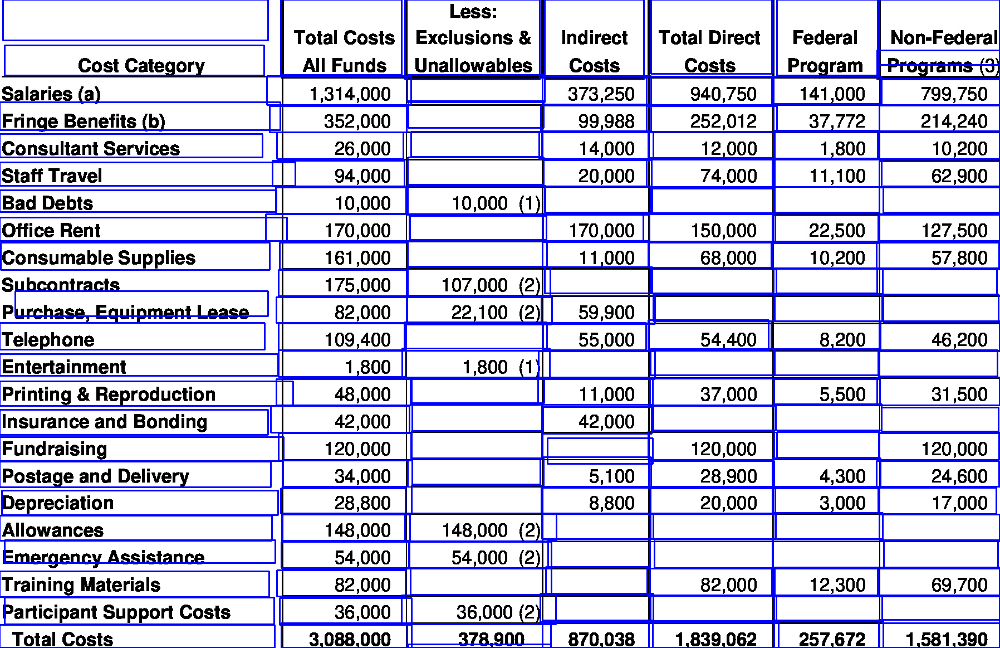}}
%%%%%%%%%%%%%%%%%%%%%%%%%%%%%%%%%%%%%%%%%%%%%%%%
%ICDAR-2019
\vspace{0.001\textwidth}
\fbox{
\includegraphics[width=0.29\linewidth, height=0.2\linewidth]{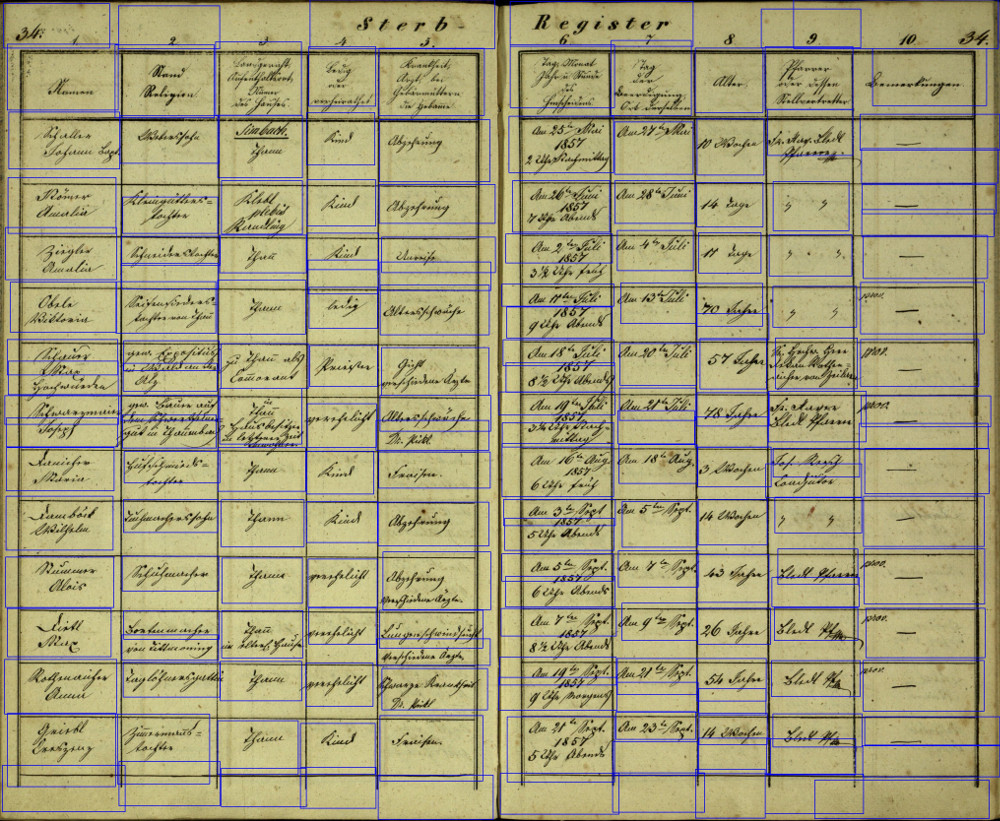}}
\hspace{-0.01\textwidth}
\fbox{
\includegraphics[width=0.29\linewidth, height=0.2\linewidth]{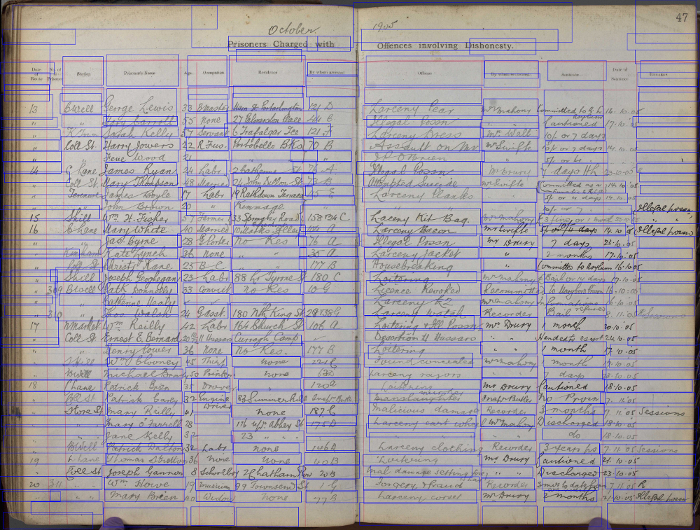}}
\hspace{-0.01\textwidth}
\fbox{
\includegraphics[width=0.29\linewidth, height=0.2\linewidth]{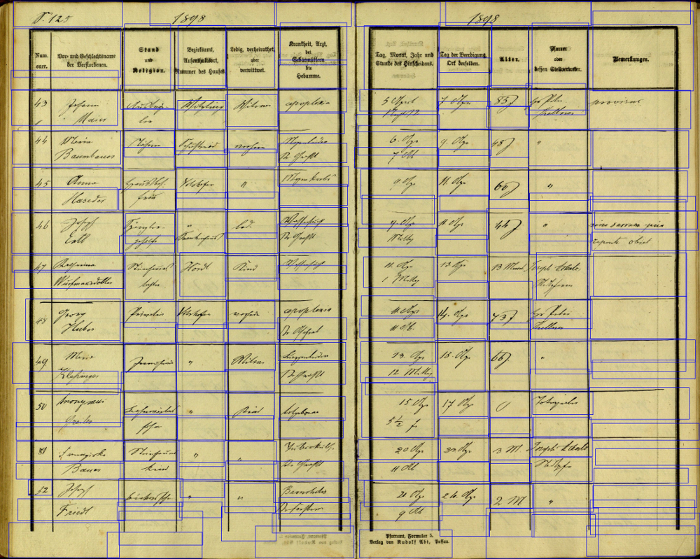}}
%%%%%%%%%%%%%%%%%%%%%%%%%%%%%%%%%%%%
%%%%%%%%%%%%%%%SciTSR%%%%%%%%%%%%%%%
\vspace{0.001\textwidth}
\fbox{
\includegraphics[width=0.29\linewidth, height=0.2\linewidth]{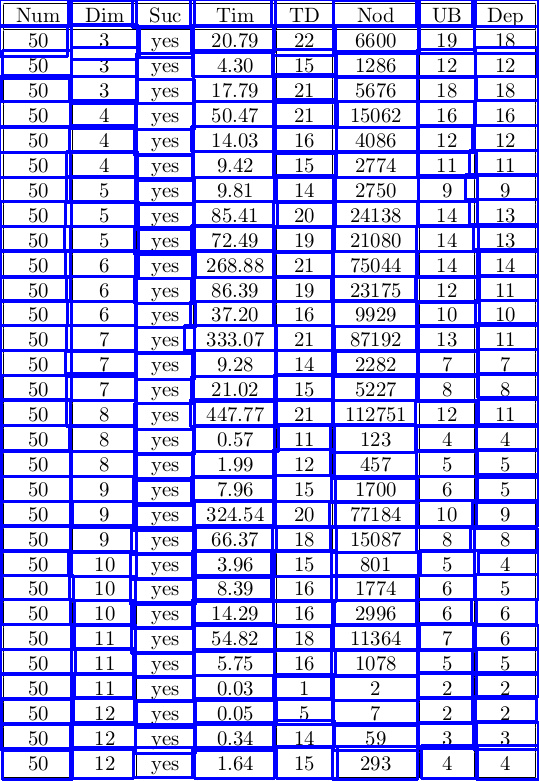}}
\hspace{-0.01\textwidth}
\fbox{
\includegraphics[width=0.29\linewidth, height=0.2\linewidth]{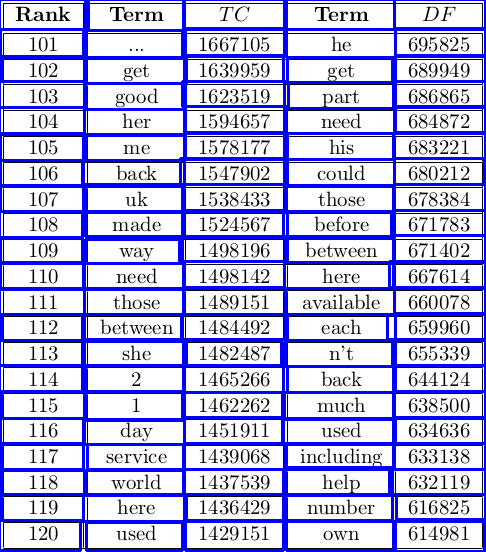}}
\hspace{-0.01\textwidth}
\fbox{
\includegraphics[width=0.29\linewidth, height=0.2\linewidth]{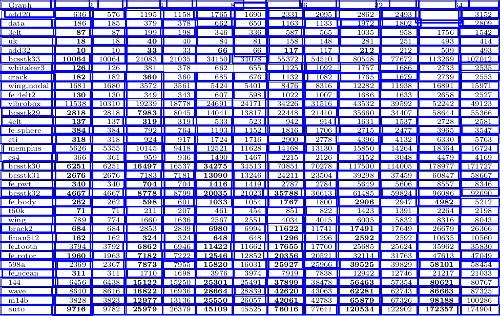}}
%%%%%%%%%%%%%%%%%%%%%%%%%%%%%%%%%%%%
%%%%%%%%%%%%%%%SciTSR-COMP%%%%%%%%%%%%%%%
\vspace{0.001\textwidth}
\fbox{
\includegraphics[width=0.29\linewidth, height=0.2\linewidth]{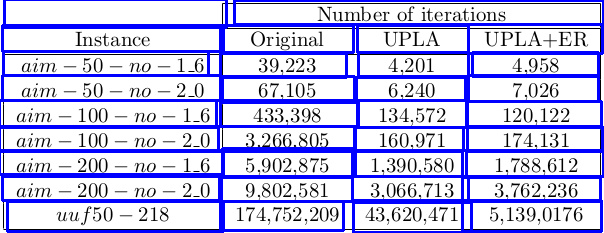}}
\hspace{-0.01\textwidth}
\fbox{
\includegraphics[width=0.29\linewidth, height=0.2\linewidth]{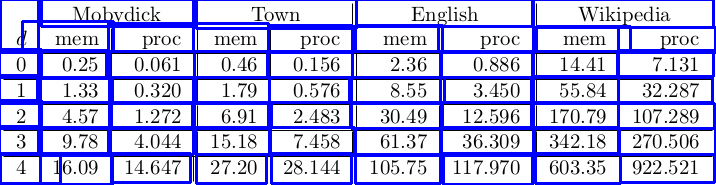}}
\hspace{-0.01\textwidth}
\fbox{
\includegraphics[width=0.29\linewidth, height=0.2\linewidth]{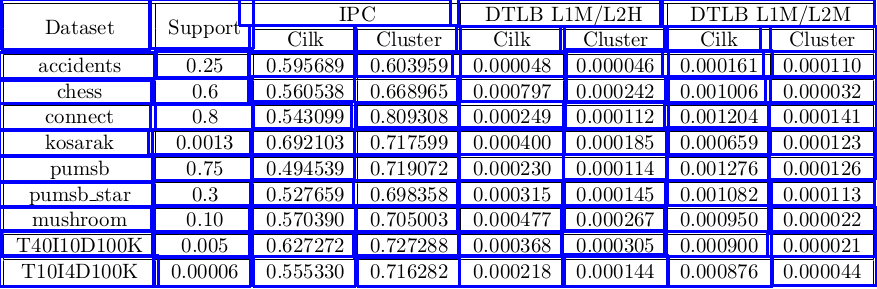}}
%%%%%%%%%%%%%%%%%%%%%%%%%%%%%%%%%%%
%%%%%%%%%%TableBank%%%%%%%%%%%%%%%%
\vspace{0.001\textwidth}
\fbox{
\includegraphics[width=0.29\linewidth, height=0.2\linewidth]{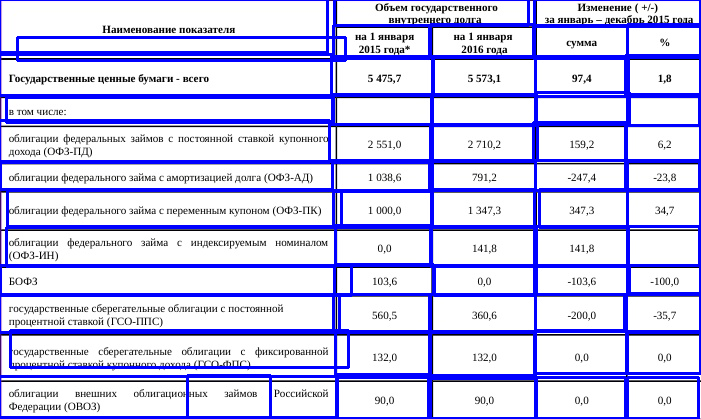}}
\hspace{-0.01\textwidth}
\fbox{
\includegraphics[width=0.29\linewidth, height=0.2\linewidth]{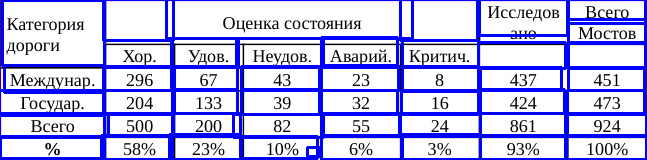}}
\hspace{-0.01\textwidth}
\fbox{
\includegraphics[width=0.29\linewidth, height=0.2\linewidth]{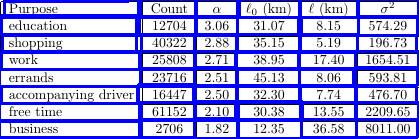}}
\end{center}
\caption{Sample intermediate cell detection results of {\sc t}ab{\sc s}truct-{\sc n}et on table images of {\sc icdar-2013} (in \textbf{First Row}), {\sc icdar-2019} (in \textbf{Second Row}), {\sc s}ci{\sc tsr} (in \textbf{Third Row}), {\sc s}ci{\sc tsr-comp} (in \textbf{Fourth Row}) and {\sc t}able{\sc b}ank (in \textbf{Fifth Row}) datasets.}
\label{fig_cell1}
\end{figure}

\begin{figure}[t]
\begin{center}
\fbox{
\includegraphics[width=0.29\linewidth, height=0.2\linewidth]{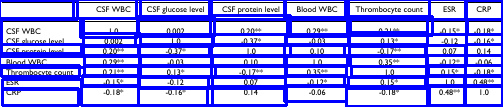}}
\hspace{-0.01\textwidth}
\fbox{
\includegraphics[width=0.29\linewidth, height=0.2\linewidth]{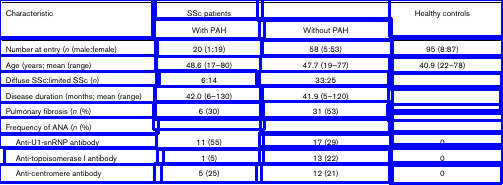}}
\hspace{-0.01\textwidth}
\fbox{
\includegraphics[width=0.29\linewidth, height=0.2\linewidth]{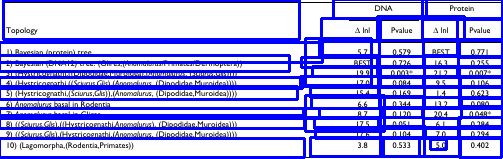}}
%%%%%%%%%%%%%%%%%%%%%%%%%%%%
%%%%%%%%%%UNLV%%%%%%%%%%%%%%
\fbox{
\includegraphics[width=0.29\linewidth, height=0.2\linewidth]{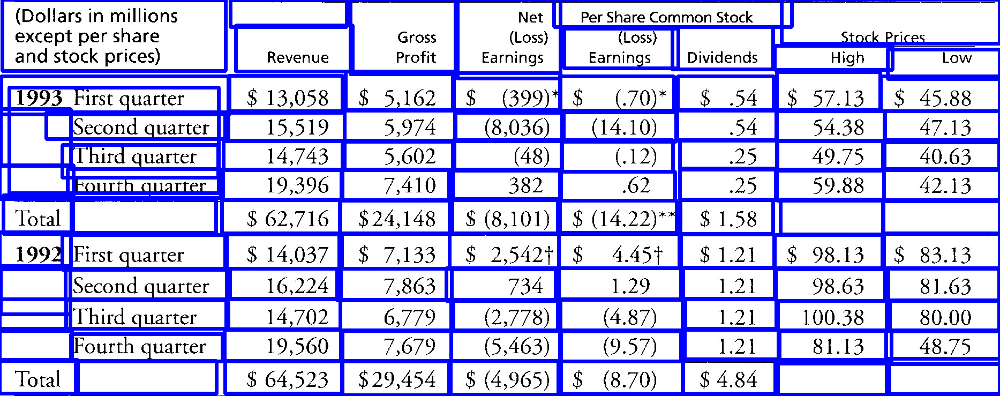}}
\hspace{-0.01\textwidth}
\fbox{
\includegraphics[width=0.29\linewidth, height=0.2\linewidth]{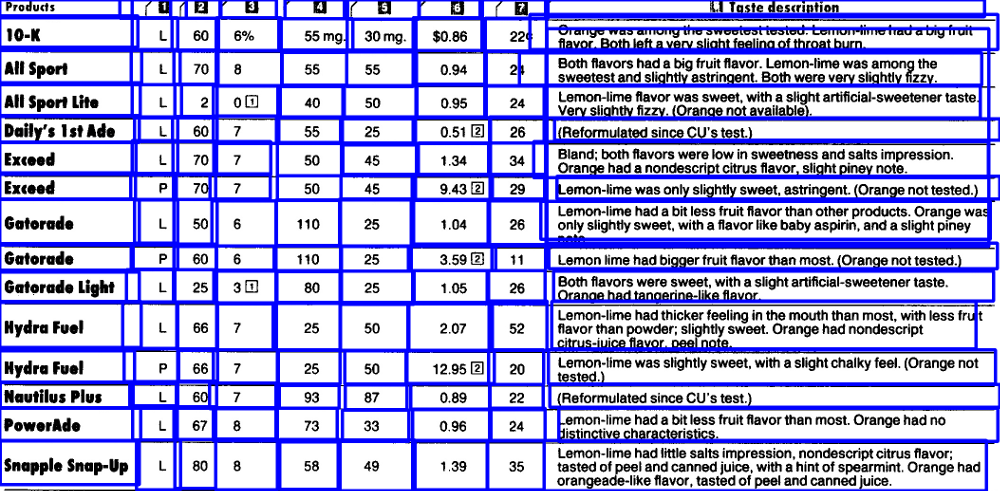}}
\hspace{-0.01\textwidth}
\fbox{
\includegraphics[width=0.29\linewidth, height=0.2\linewidth]{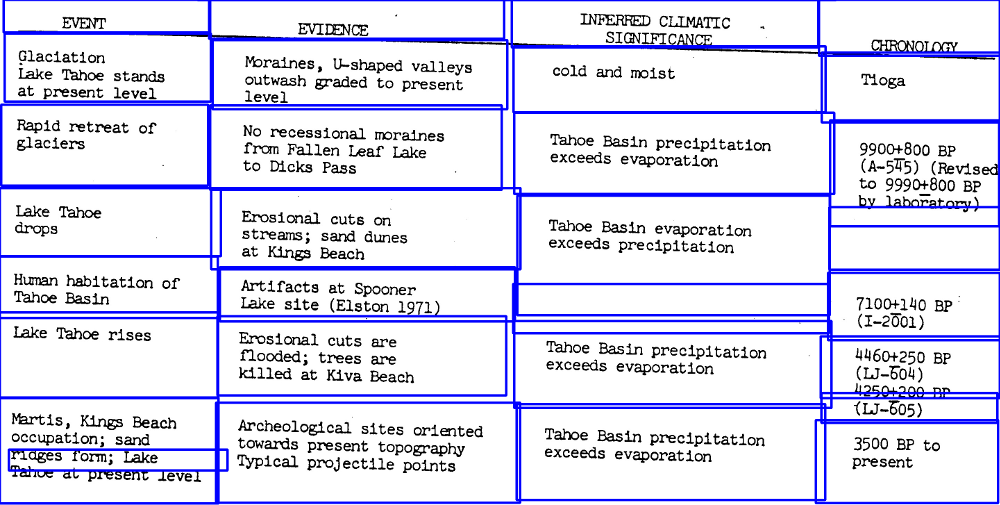}}
\end{center}
\caption{Sample intermediate cell detection results of {\sc t}ab{\sc s}truct-{\sc n}et on table images of {\sc p}ub{\sc t}ab{\sc n}et (in \textbf{First Row}) and {\sc unlv} (in \textbf{Second Row}) datasets.}
\label{fig_cell2}
\end{figure}

Figures~\ref{fig_cell1}-\ref{fig_cell2} demonstrate some qualitative results of cell detection on all the evaluation datasets. From the figures, it can be seen that our model is able to work in the presence of archival table images, multiple row/column spanning cells, varied table layouts and multiple line spanning cells. This indicates the robustness of our method under multiple kind of table images.

%%%%%%%%%%%%%%%%in new page %%%%%%%%%%%%%%%%%%%%%%
%cell detection of PubTabNet 

%%%%%%%%%%%%%%%%%%%%%%%%%%%%%%%%%%%%%%%%%%%%%%%%%%%%%
\newpage
\subsection*{Qualitative Results of Structure Recognition}

\begin{figure}
\begin{center}
\fbox{
\includegraphics[width=0.29\linewidth, height=0.2\linewidth]{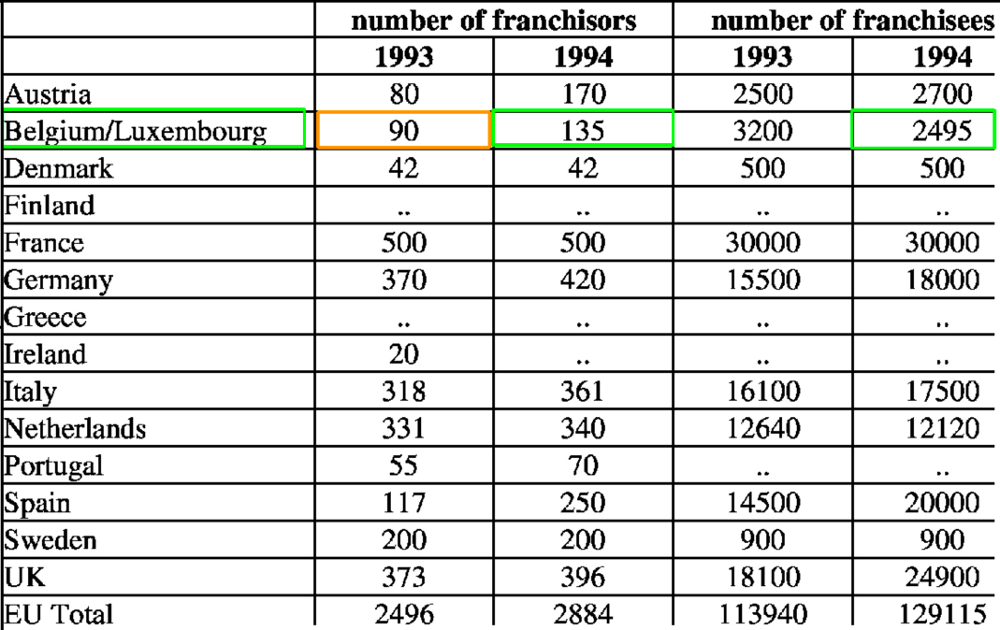}}
\hspace{-0.01\textwidth}
\fbox{
\includegraphics[width=0.29\linewidth, height=0.2\linewidth]{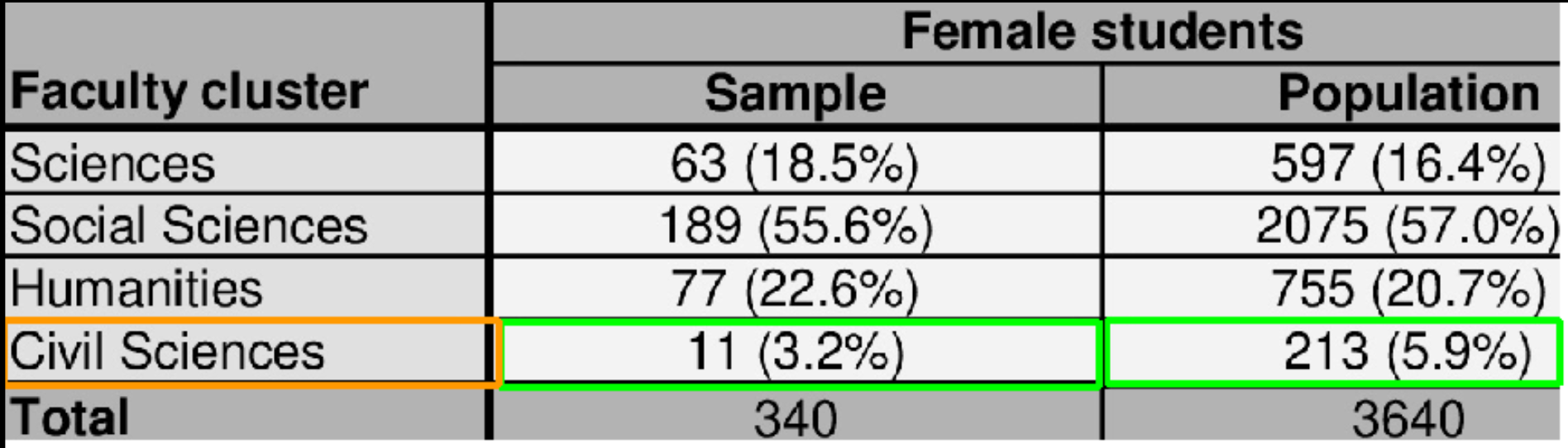}}
\hspace{-0.01\textwidth}
\fbox{
\includegraphics[width=0.29\linewidth, height=0.2\linewidth]{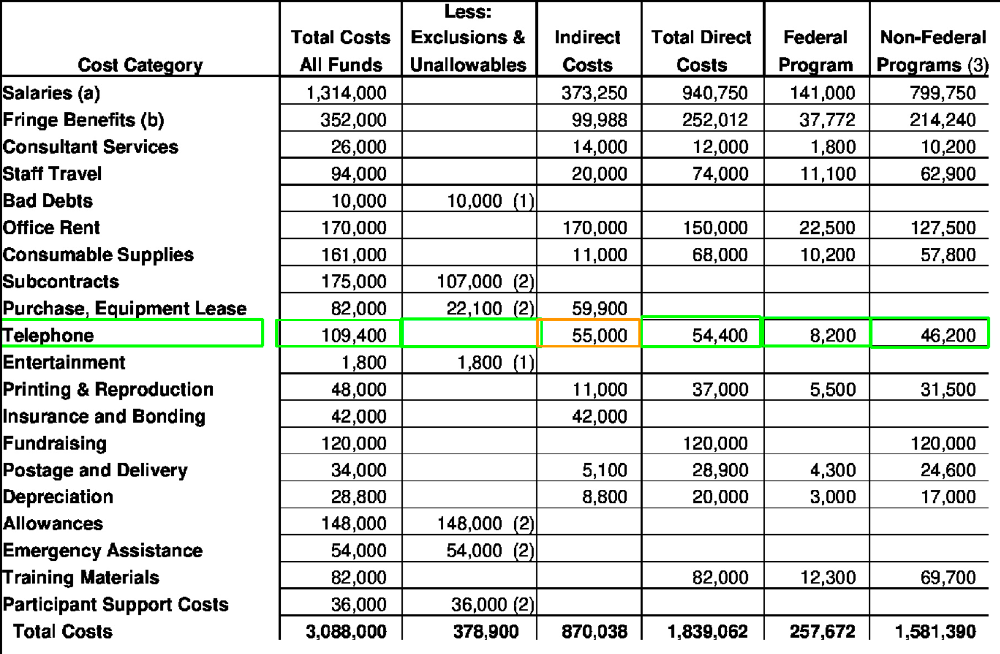}}
\vspace{0.001\textwidth}
\fbox{
\includegraphics[width=0.29\linewidth, height=0.2\linewidth]{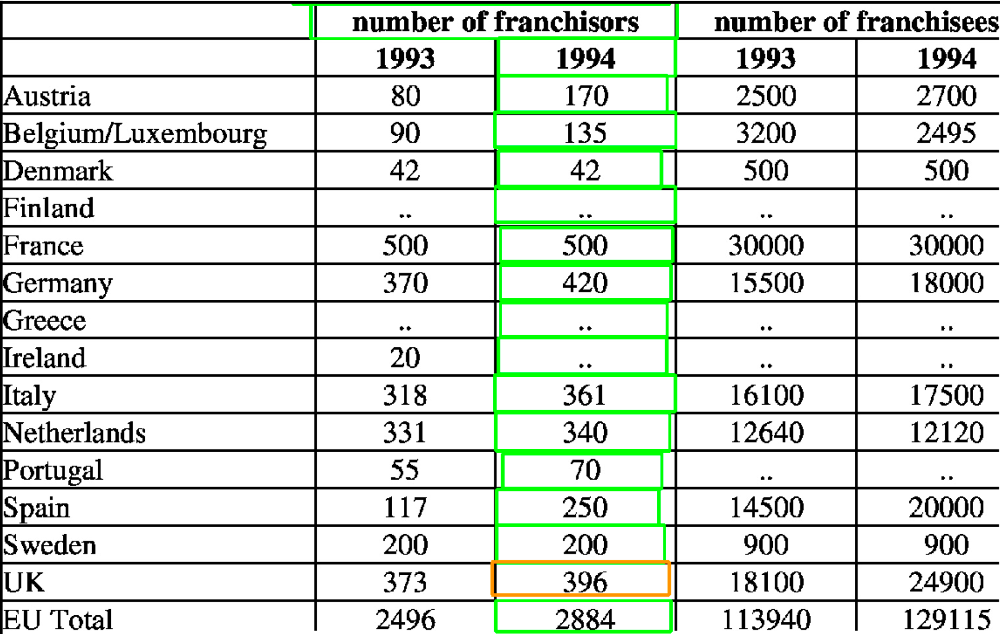}}
\hspace{-0.01\textwidth}
\fbox{
\includegraphics[width=0.29\linewidth, height=0.2\linewidth]{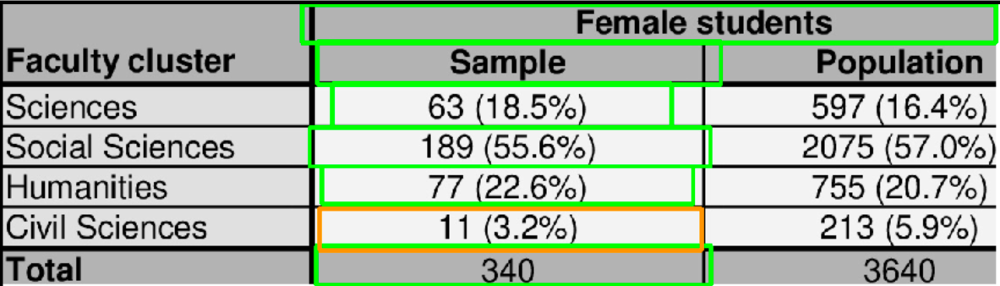}}
\hspace{-0.01\textwidth}
\fbox{
\includegraphics[width=0.29\linewidth, height=0.2\linewidth]{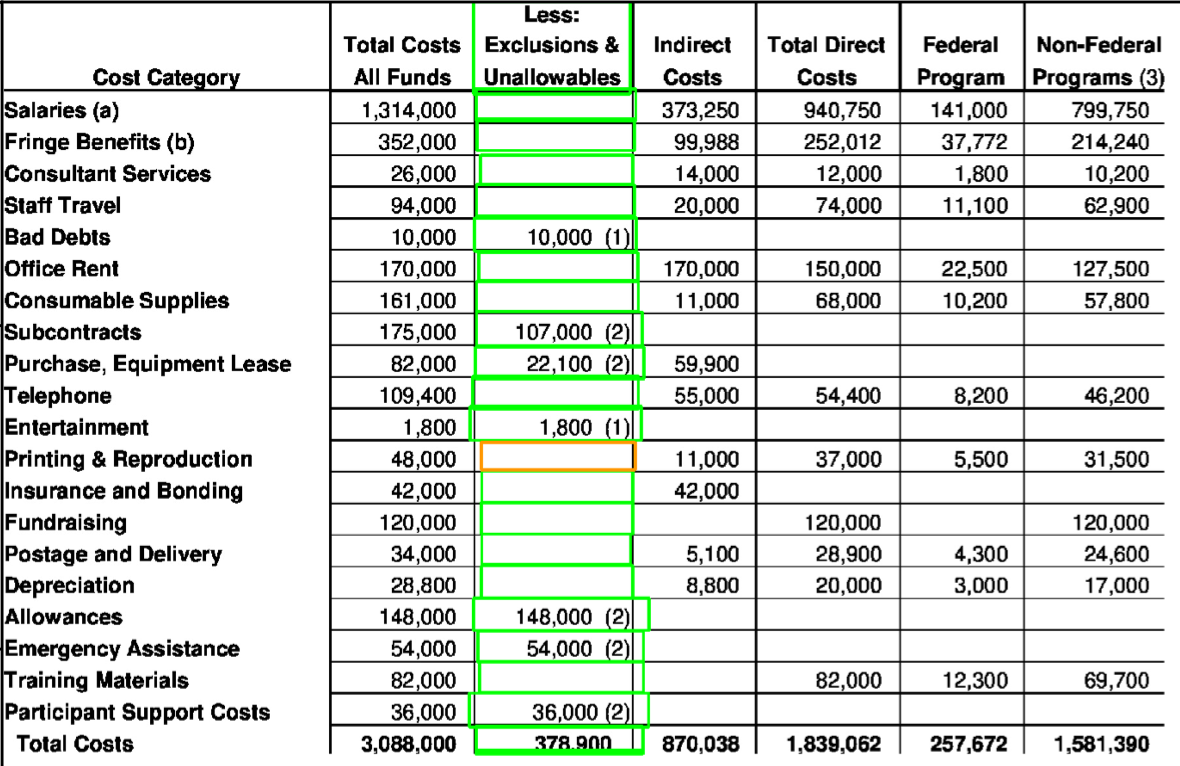}}
\end{center}
\caption{Sample structure recognition output of {\sc t}ab{\sc s}truct-{\sc n}et on table images of {\sc icdar}-2013 dataset. \textbf{First Row:} prediction of cells which belong to the same row. \textbf{Second Row:} prediction of cells which belong to the same column. Cells marked with orange colour represent the examine cells and cells marked with green colour represent those which belong to the same row/column of the examined cell.}
\label{fig_icdar_2013_structure}
\end{figure}

\begin{figure}
\begin{center}
\fbox{
\includegraphics[width=0.29\linewidth, height=0.2\linewidth]{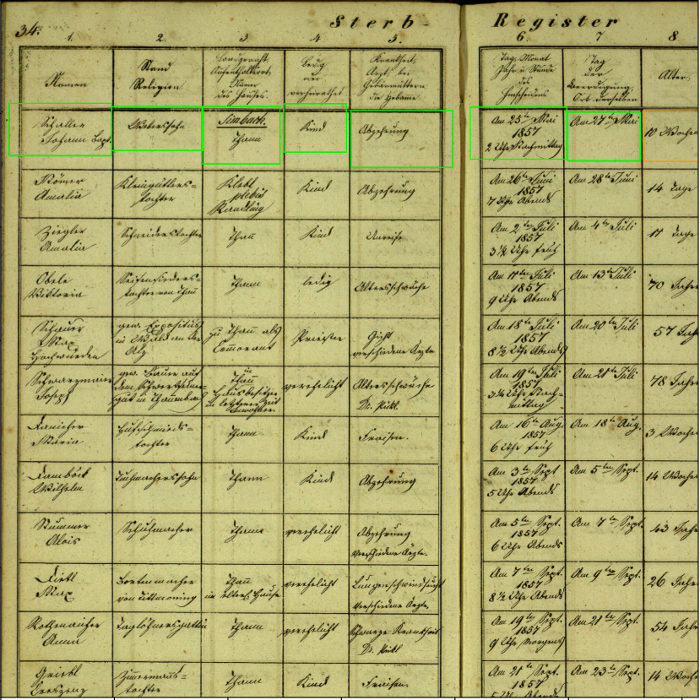}}
\hspace{-0.01\textwidth}
\fbox{
\includegraphics[width=0.29\linewidth, height=0.2\linewidth]{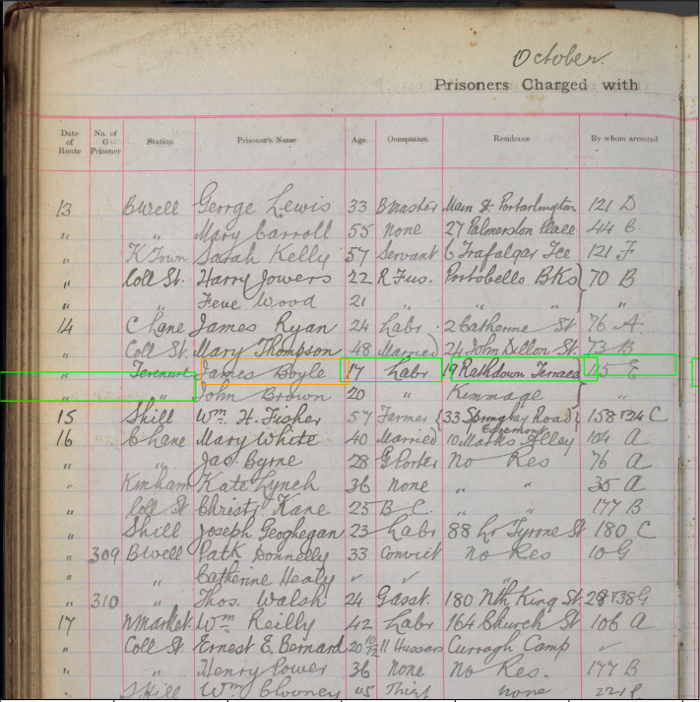}}
\hspace{-0.01\textwidth}
\fbox{
\includegraphics[width=0.29\linewidth, height=0.2\linewidth]{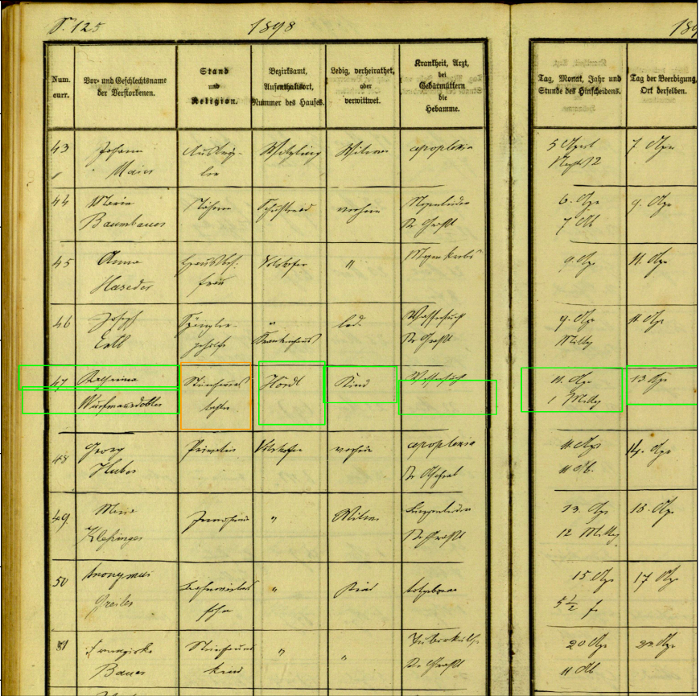}}
\vspace{0.001\textwidth}
\fbox{
\includegraphics[width=0.29\linewidth, height=0.2\linewidth]{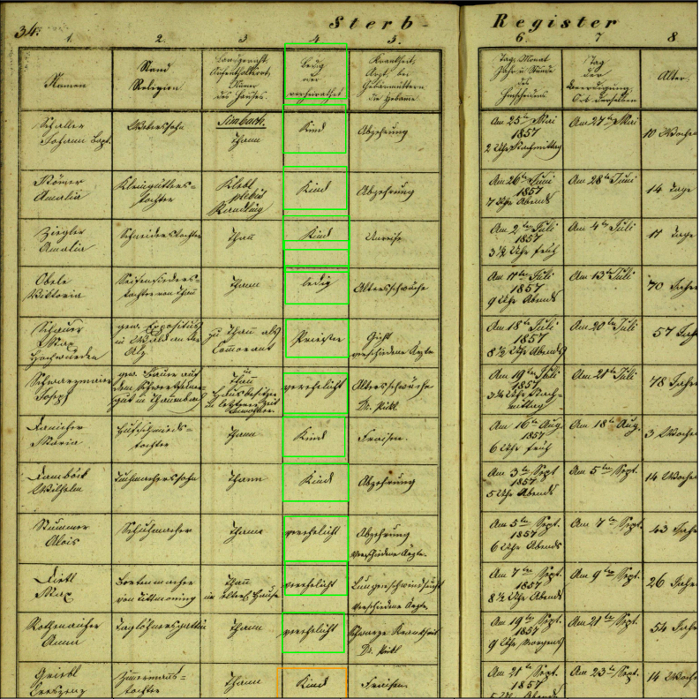}}
\hspace{-0.01\textwidth}
\fbox{
\includegraphics[width=0.29\linewidth, height=0.2\linewidth]{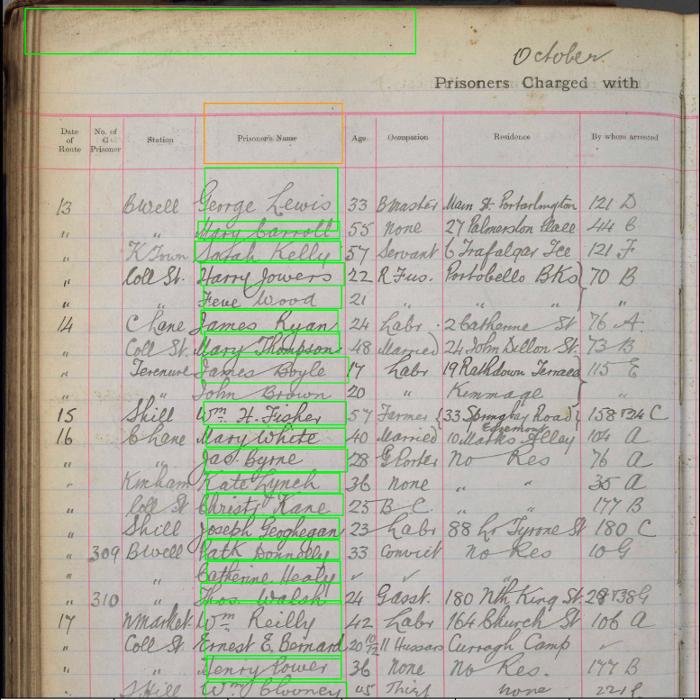}}
\hspace{-0.01\textwidth}
\fbox{
\includegraphics[width=0.29\linewidth, height=0.2\linewidth]{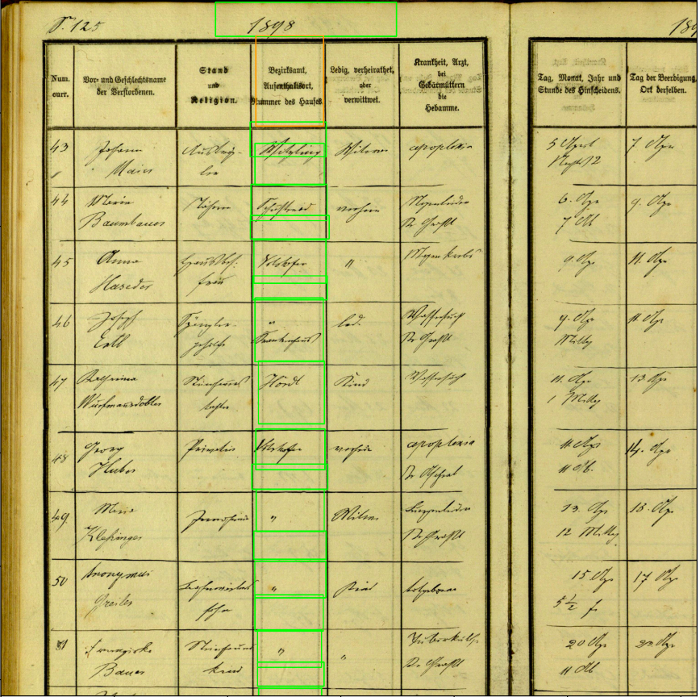}}
\end{center}
\caption{Sample structure recognition output of {\sc t}ab{\sc s}truct-{\sc n}et on table images of {\sc icdar}-2019 dataset. \textbf{First Row:} prediction of cells which belong to the same row. \textbf{Second Row:} prediction of cells which belong to the same column. Cells marked with orange colour represent the examine cells and cells marked with green colour represent those which belong to the same row/column of the examined cell.}
\label{fig_icdar_2019_structure}
\end{figure}

\begin{figure}
\begin{center}
\fbox{
\includegraphics[width=0.29\linewidth, height=0.2\linewidth]{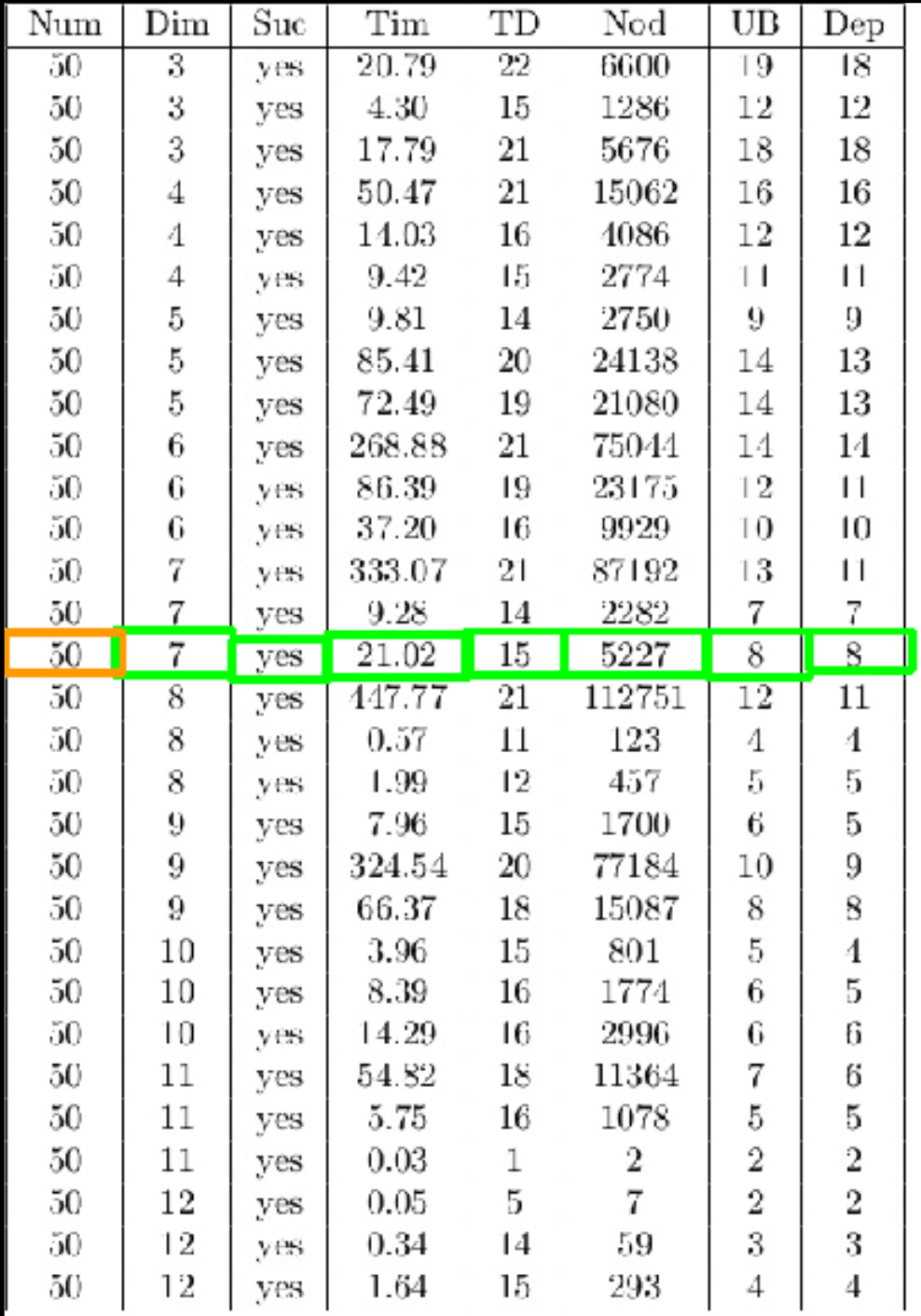}}
\hspace{-0.01\textwidth}
\fbox{
\includegraphics[width=0.29\linewidth, height=0.2\linewidth]{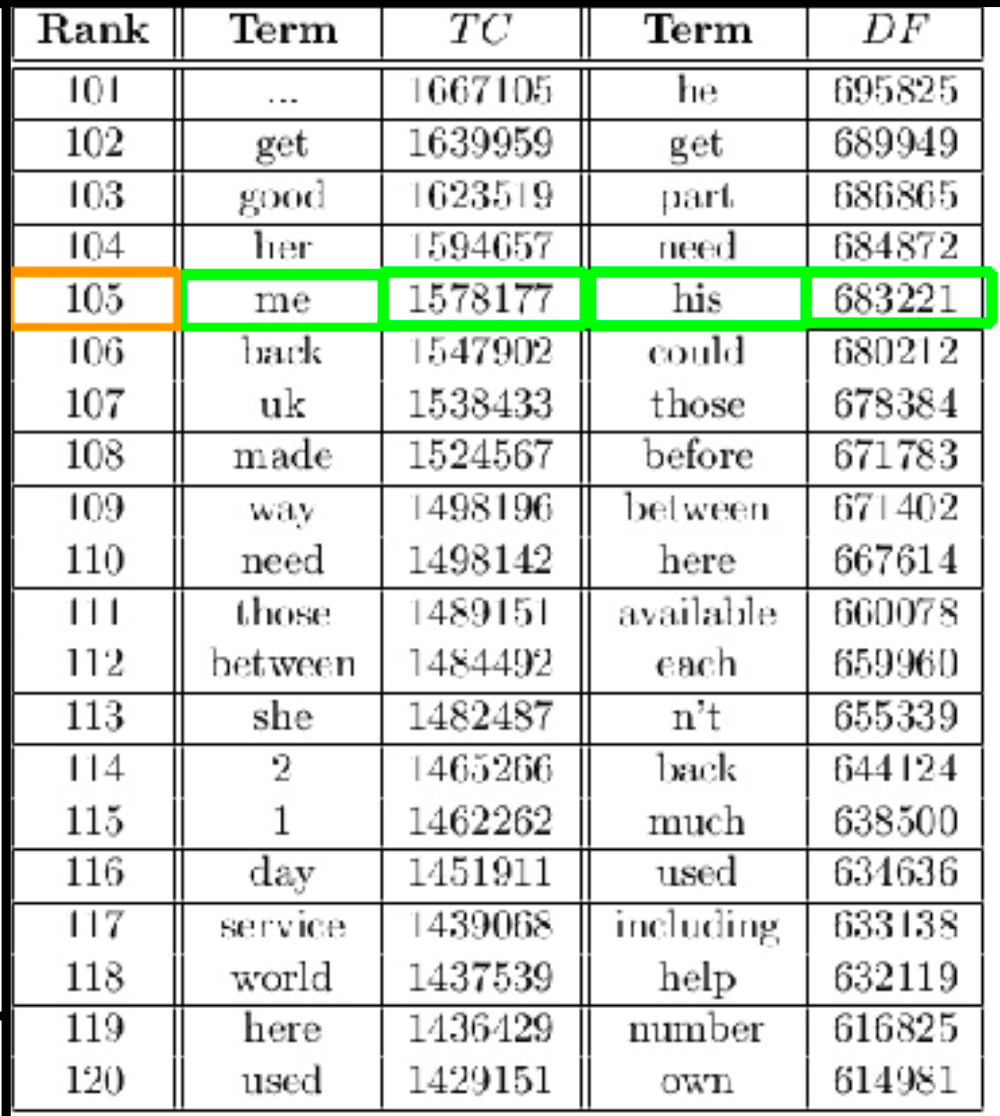}}
\hspace{-0.01\textwidth}
\fbox{
\includegraphics[width=0.29\linewidth, height=0.2\linewidth]{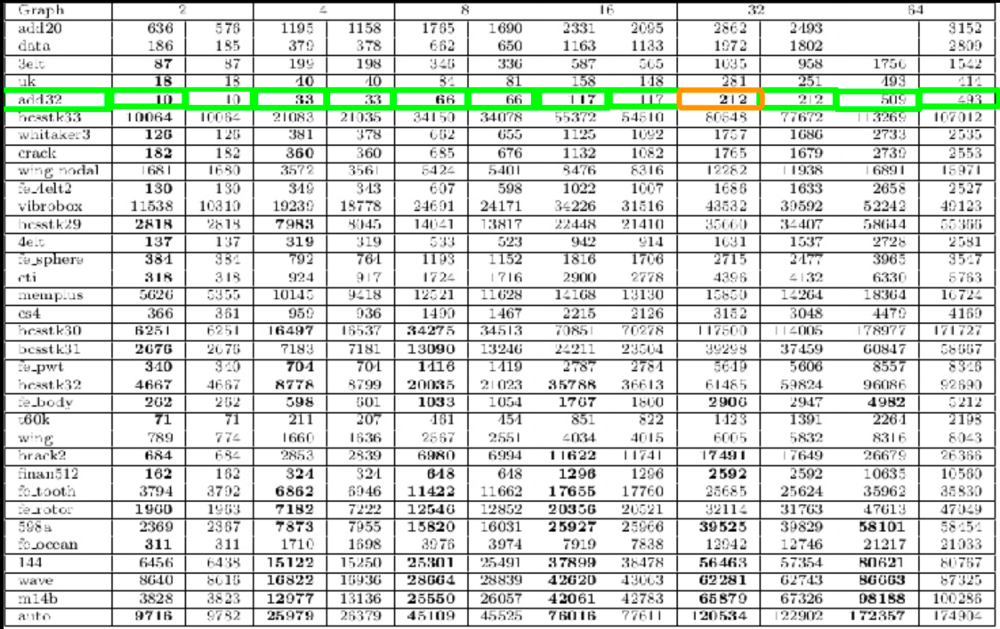}}
\vspace{0.001\textwidth}
\fbox{
\includegraphics[width=0.29\linewidth, height=0.2\linewidth]{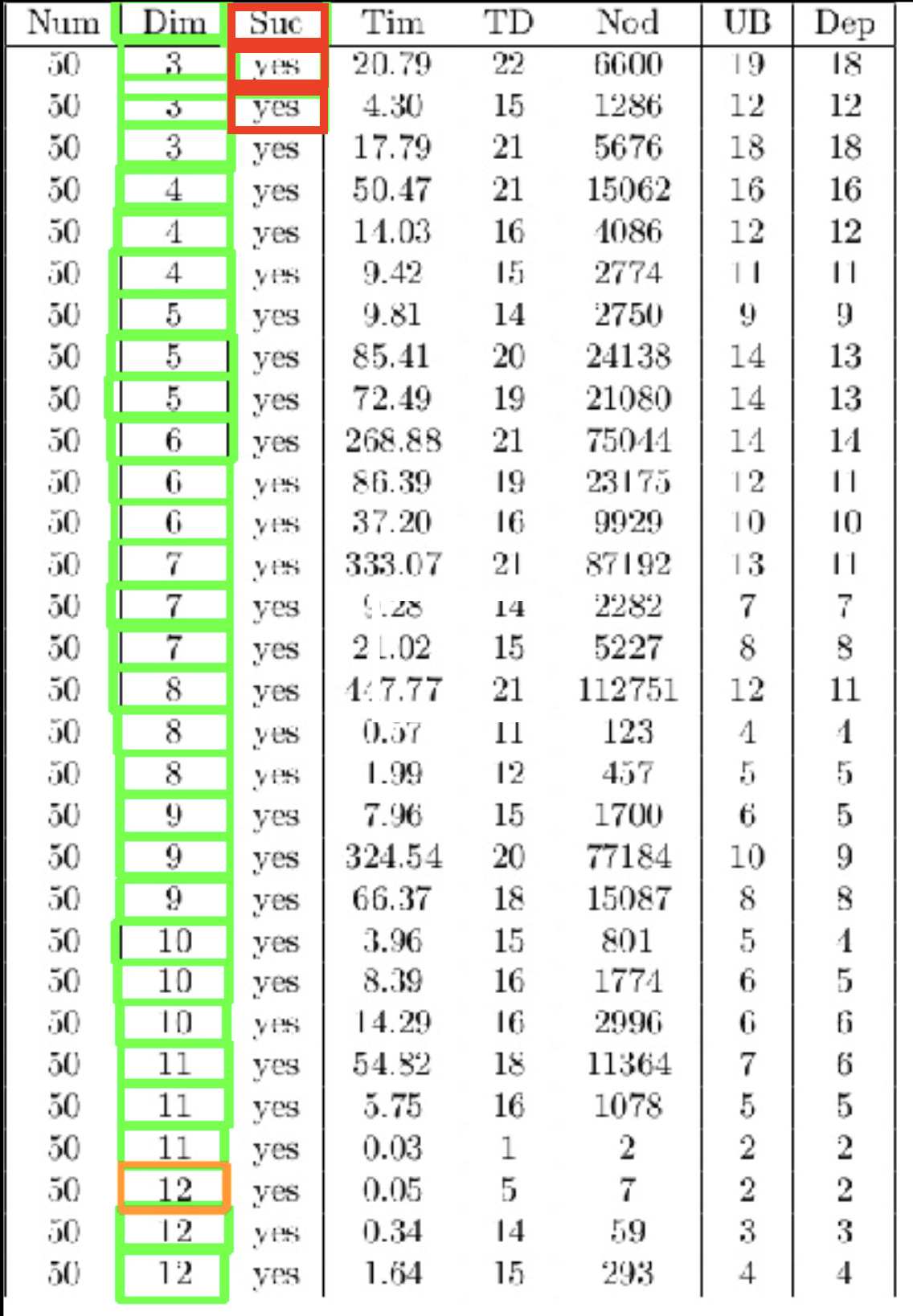}}
\hspace{-0.01\textwidth}
\fbox{
\includegraphics[width=0.29\linewidth, height=0.2\linewidth]{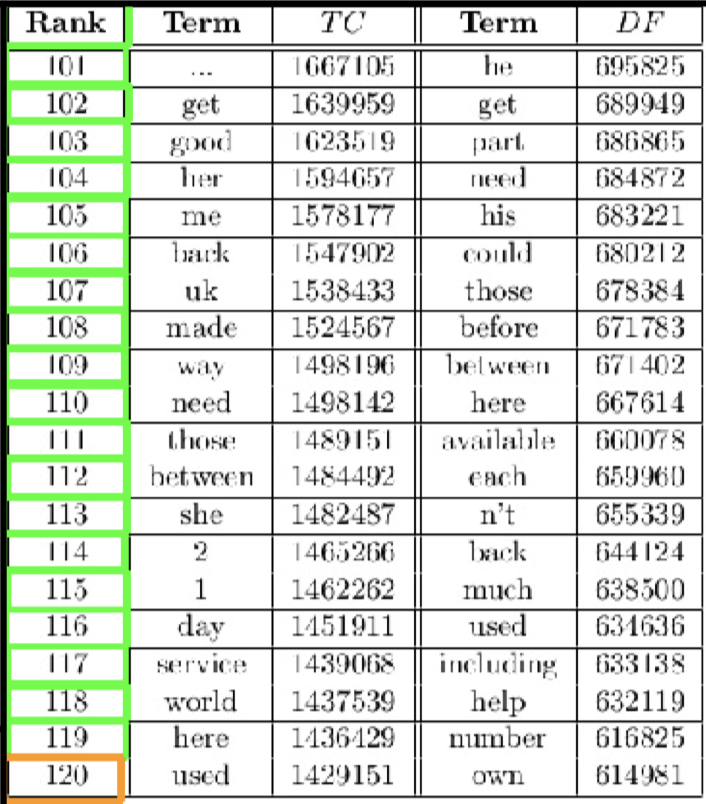}}
\hspace{-0.01\textwidth}
\fbox{
\includegraphics[width=0.29\linewidth, height=0.2\linewidth]{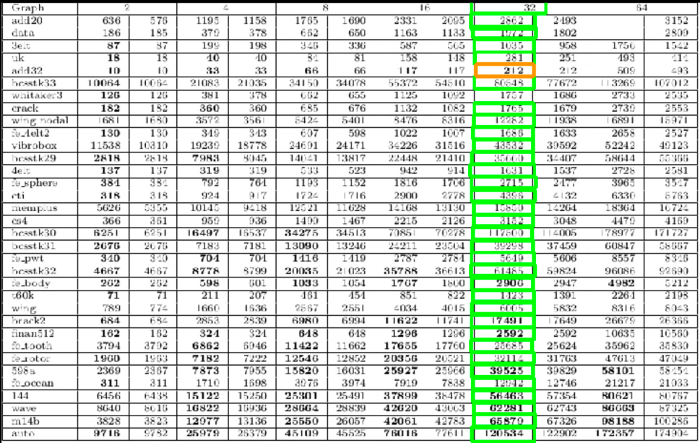}}
\end{center}
\caption{Sample structure recognition output of {\sc t}ab{\sc s}truct-{\sc n}et on table images of {\sc s}ci{\sc tsr} dataset. \textbf{First Row:} prediction of cells which belong to the same row. \textbf{Second Row:} prediction of cells which belong to the same column. Cells marked with orange colour represent the examine cells and cells marked with green colour represent those which belong to the same row/column of the examined cell.}
\label{fig_scitsr_structure}
\end{figure}

\begin{figure}
\begin{center}
\fbox{
\includegraphics[width=0.29\linewidth, height=0.2\linewidth]{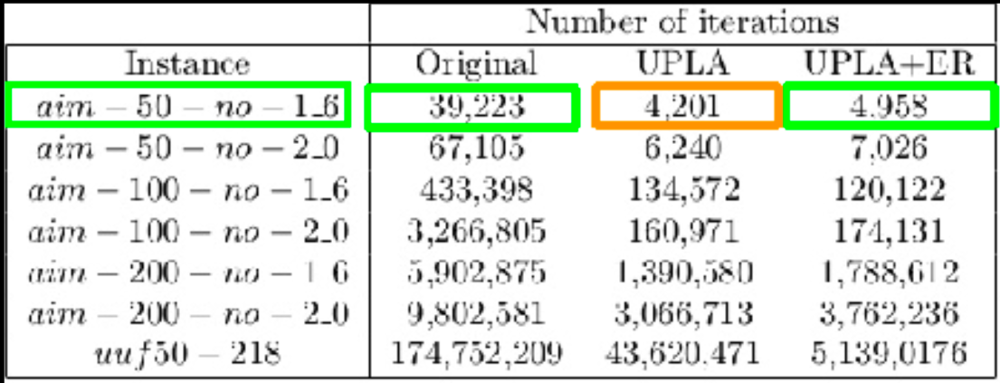}}
\hspace{-0.01\textwidth}
\fbox{
\includegraphics[width=0.29\linewidth, height=0.2\linewidth]{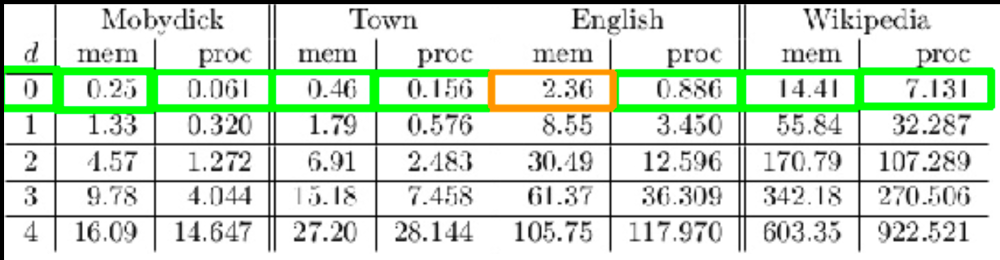}}
\hspace{-0.01\textwidth}
\fbox{
\includegraphics[width=0.29\linewidth, height=0.2\linewidth]{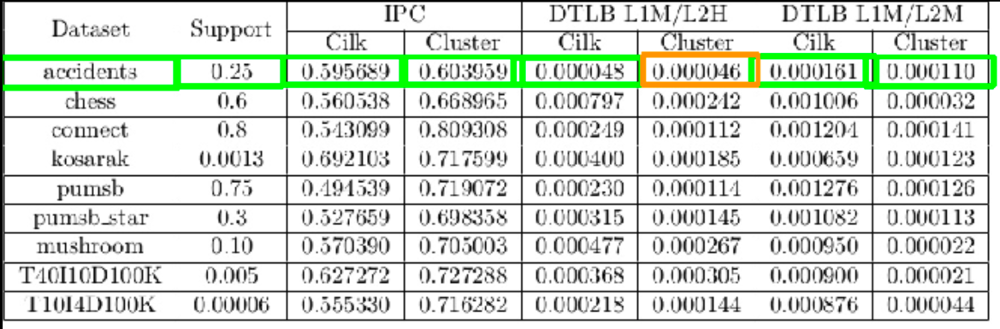}}
\vspace{0.001\textwidth}
\fbox{
\includegraphics[width=0.29\linewidth, height=0.2\linewidth]{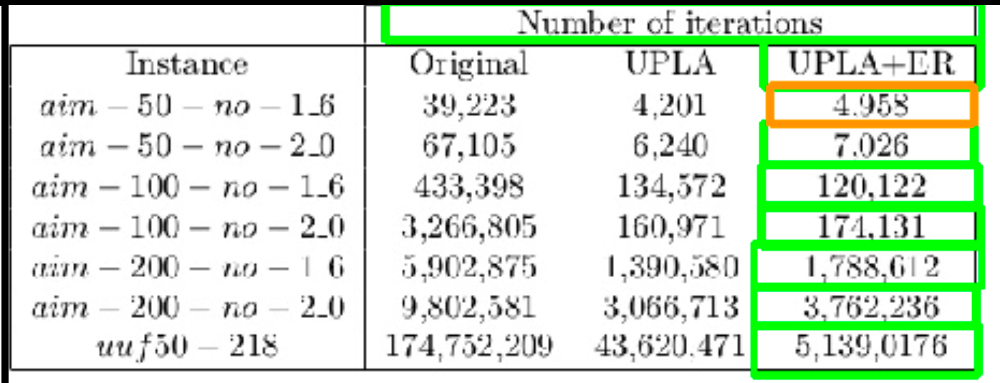}}
\hspace{-0.01\textwidth}
\fbox{
\includegraphics[width=0.29\linewidth, height=0.2\linewidth]{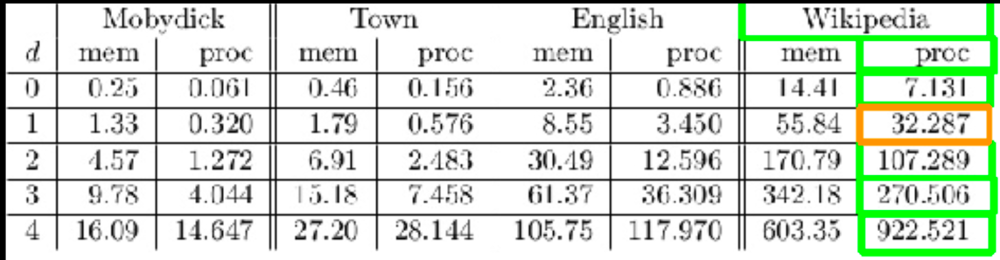}}
\hspace{-0.01\textwidth}
\fbox{
\includegraphics[width=0.29\linewidth, height=0.2\linewidth]{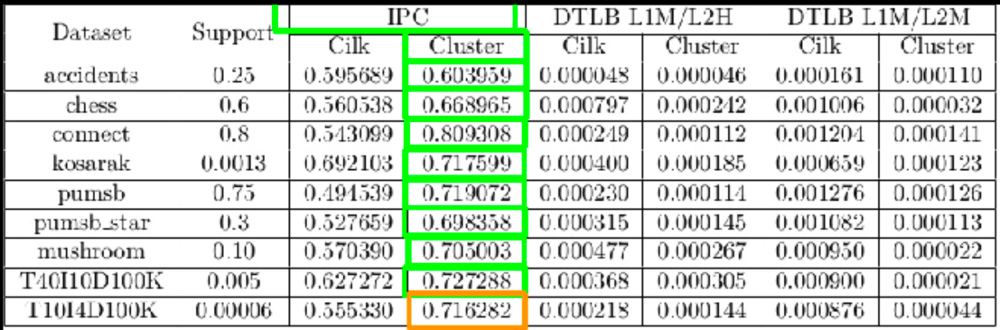}}
\end{center}
\caption{Sample structure recognition output of {\sc t}ab{\sc s}truct-{\sc n}et on table images of {\sc s}ci{\sc tsr-comp} dataset. \textbf{First Row:} prediction of cells which belong to the same row. \textbf{Second Row:} prediction of cells which belong to the same column. Cells marked with orange colour represent the examine cells and cells marked with green colour represent those which belong to the same row/column of the examined cell.}
\label{fig_scitsr_comp_structure}
\end{figure}

\begin{figure}
\begin{center}
\fbox{
\includegraphics[width=0.29\linewidth, height=0.2\linewidth]{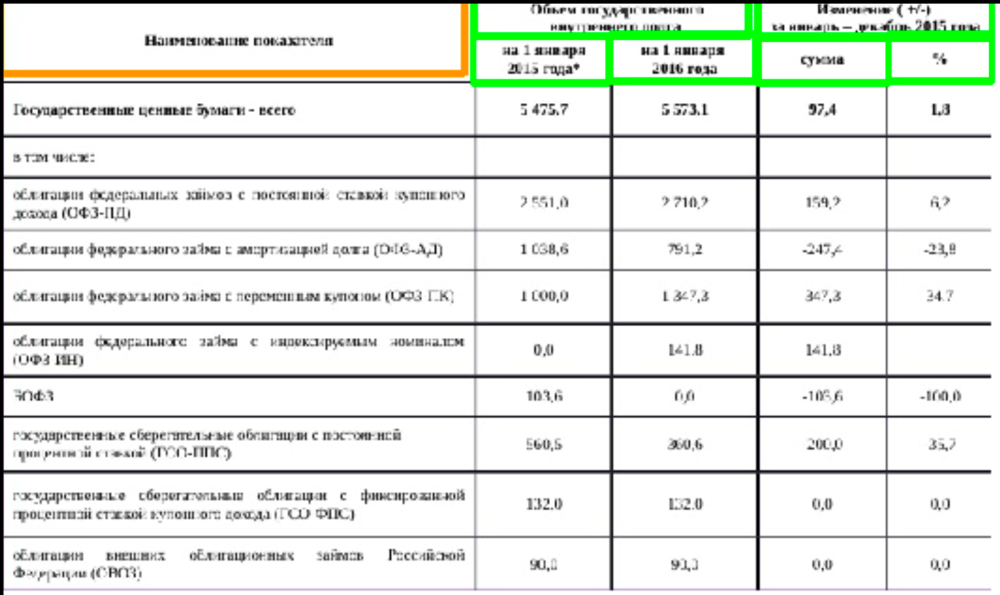}}
\hspace{-0.01\textwidth}
\fbox{
\includegraphics[width=0.29\linewidth, height=0.2\linewidth]{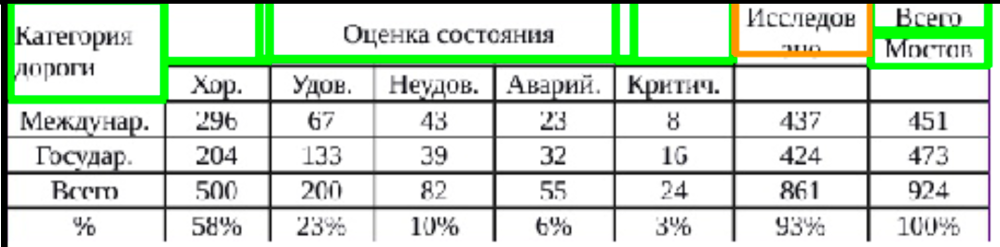}}
\hspace{-0.01\textwidth}
\fbox{
\includegraphics[width=0.29\linewidth, height=0.2\linewidth]{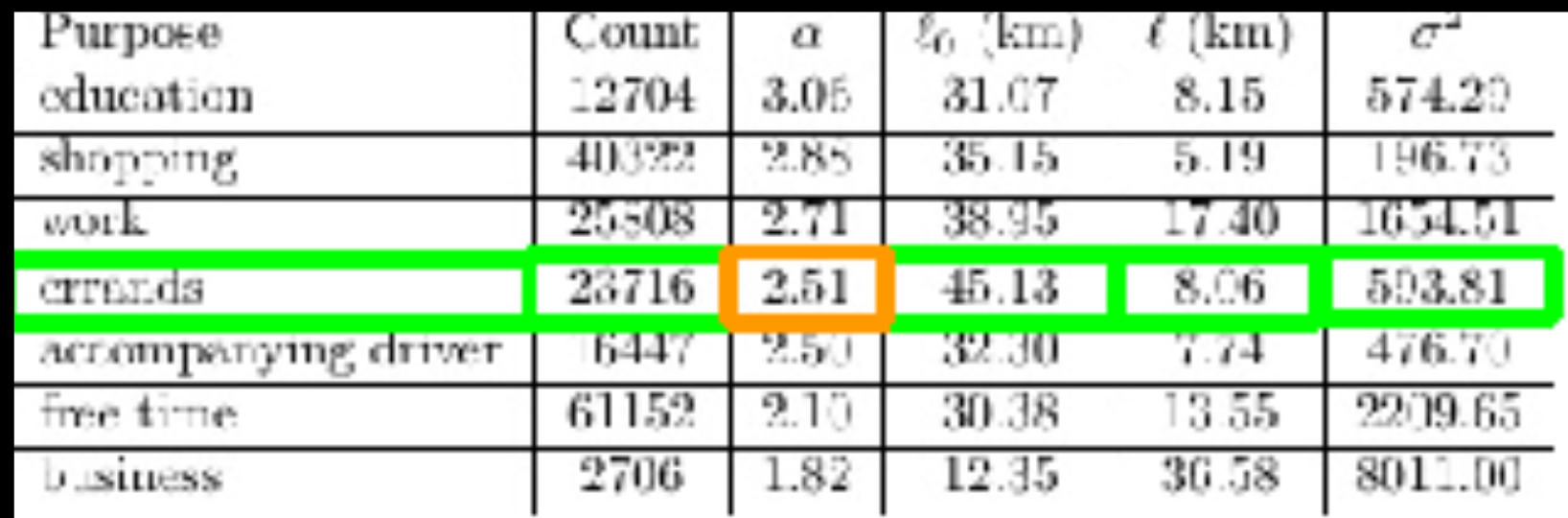}}
\vspace{0.001\textwidth}
\fbox{
\includegraphics[width=0.29\linewidth, height=0.2\linewidth]{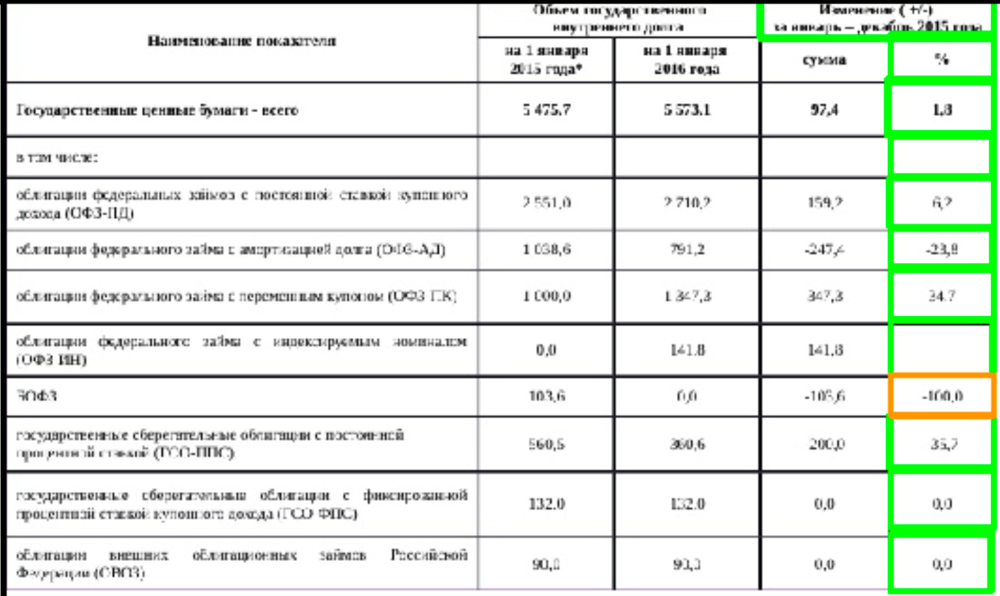}}
\hspace{-0.01\textwidth}
\fbox{
\includegraphics[width=0.29\linewidth, height=0.2\linewidth]{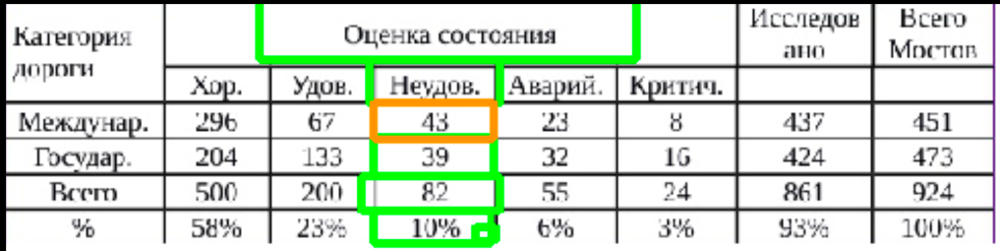}}
\hspace{-0.01\textwidth}
\fbox{
\includegraphics[width=0.29\linewidth, height=0.2\linewidth]{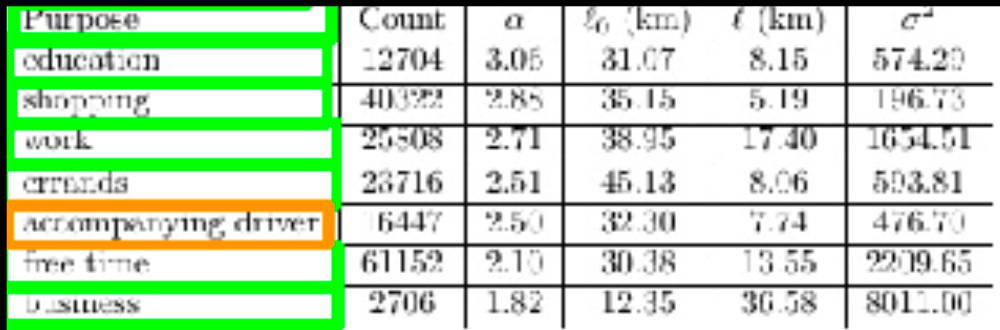}}
\end{center}
\caption{Sample structure recognition output of {\sc t}ab{\sc s}truct-{\sc n}et on table images of {\sc t}able{\sc b}ank dataset. \textbf{First Row:} prediction of cells which belong to the same row. \textbf{Second Row:} prediction of cells which belong to the same column. Cells marked with orange colour represent the examine cells and cells marked with green colour represent those which belong to the same row/column of the examined cell.}
\label{fig_tablebank_structure}
\end{figure}

\begin{figure}
\begin{center}
\fbox{
\includegraphics[width=0.29\linewidth, height=0.2\linewidth]{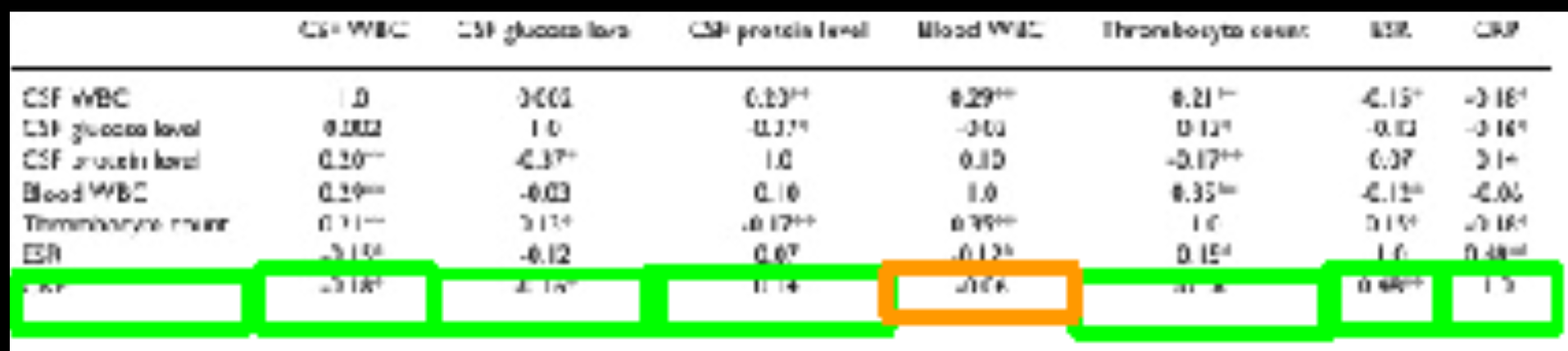}}
\hspace{-0.01\textwidth}
\fbox{
\includegraphics[width=0.29\linewidth, height=0.2\linewidth]{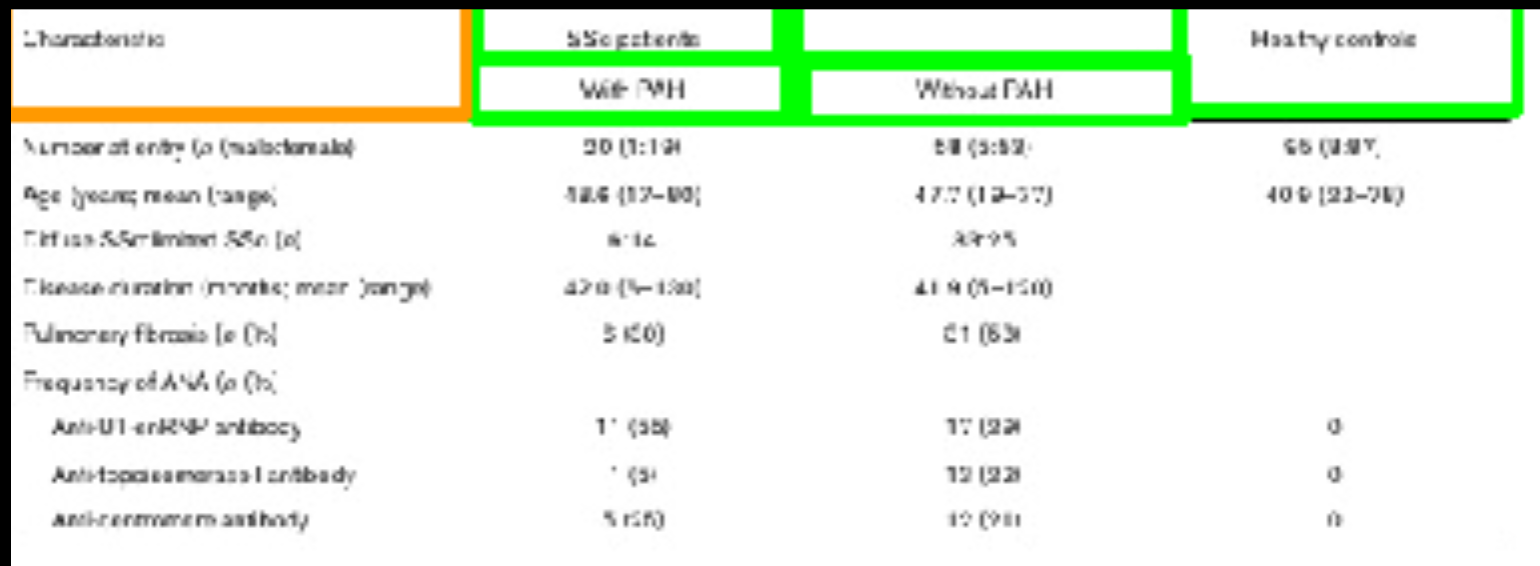}}
\hspace{-0.01\textwidth}
\fbox{
\includegraphics[width=0.29\linewidth, height=0.2\linewidth]{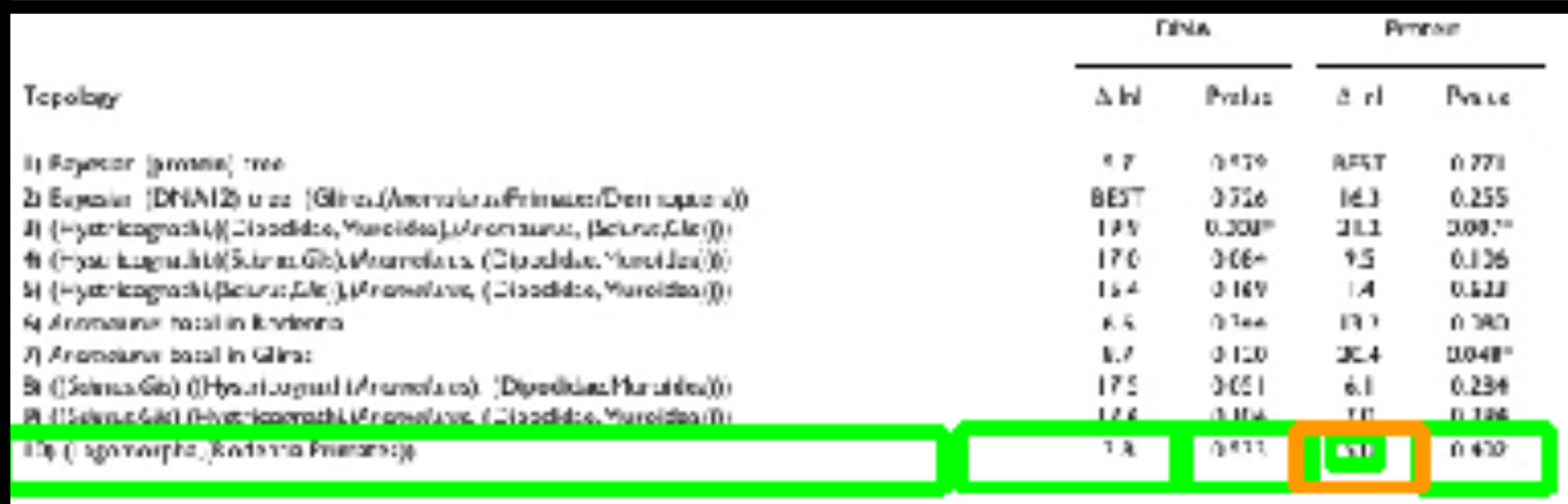}}
\vspace{0.001\textwidth}
\fbox{
\includegraphics[width=0.29\linewidth, height=0.2\linewidth]{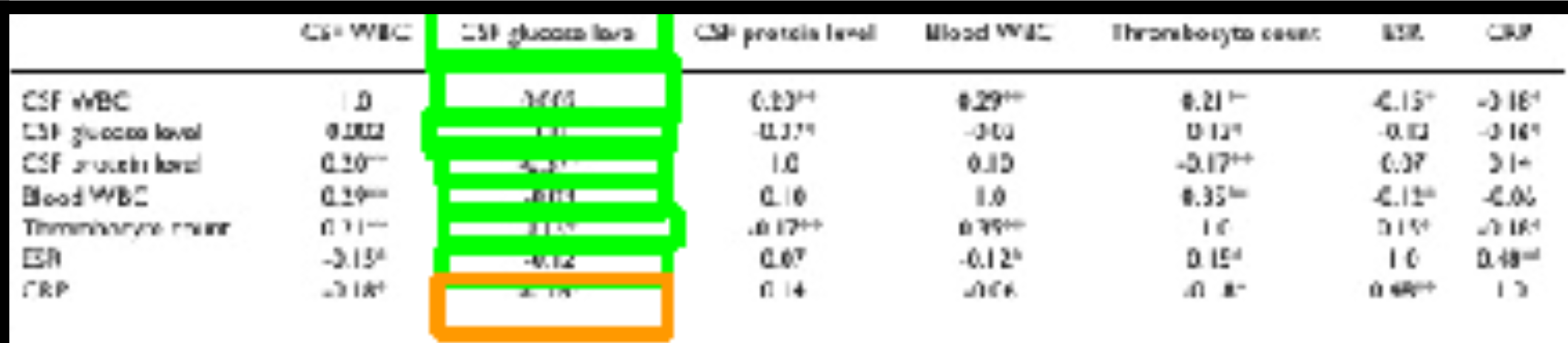}}
\hspace{-0.01\textwidth}
\fbox{
\includegraphics[width=0.29\linewidth, height=0.2\linewidth]{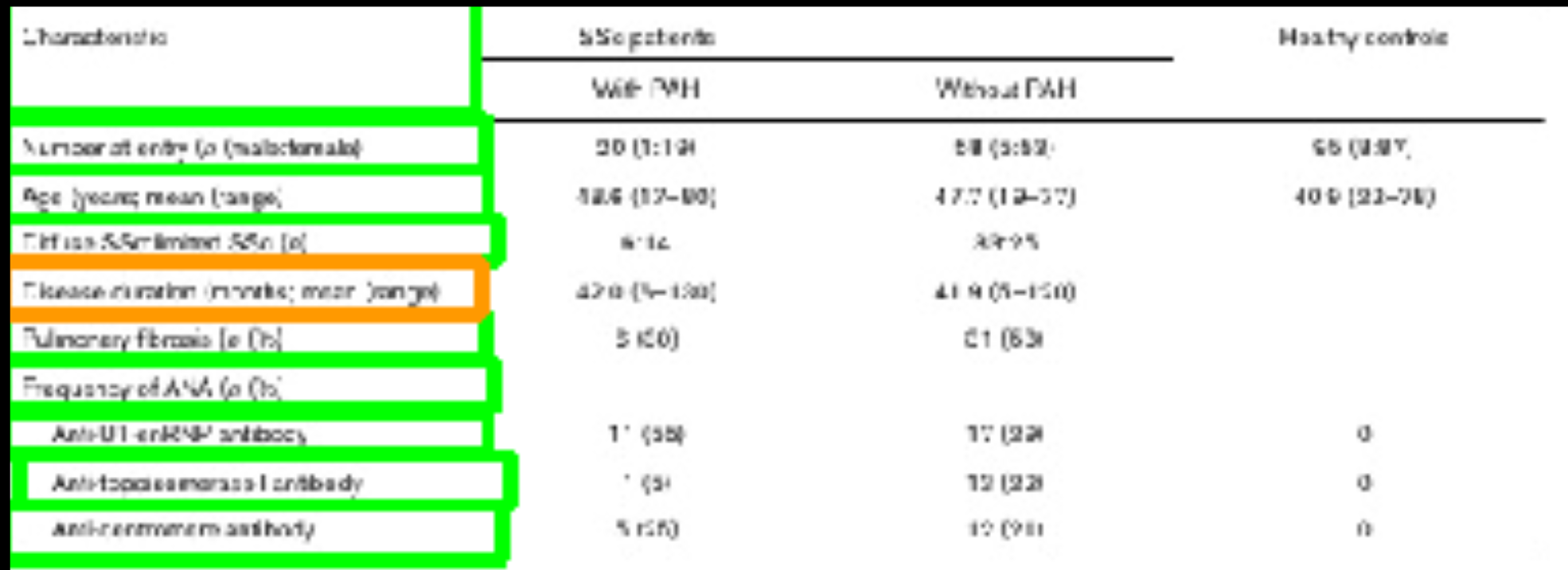}}
\hspace{-0.01\textwidth}
\fbox{
\includegraphics[width=0.29\linewidth, height=0.2\linewidth]{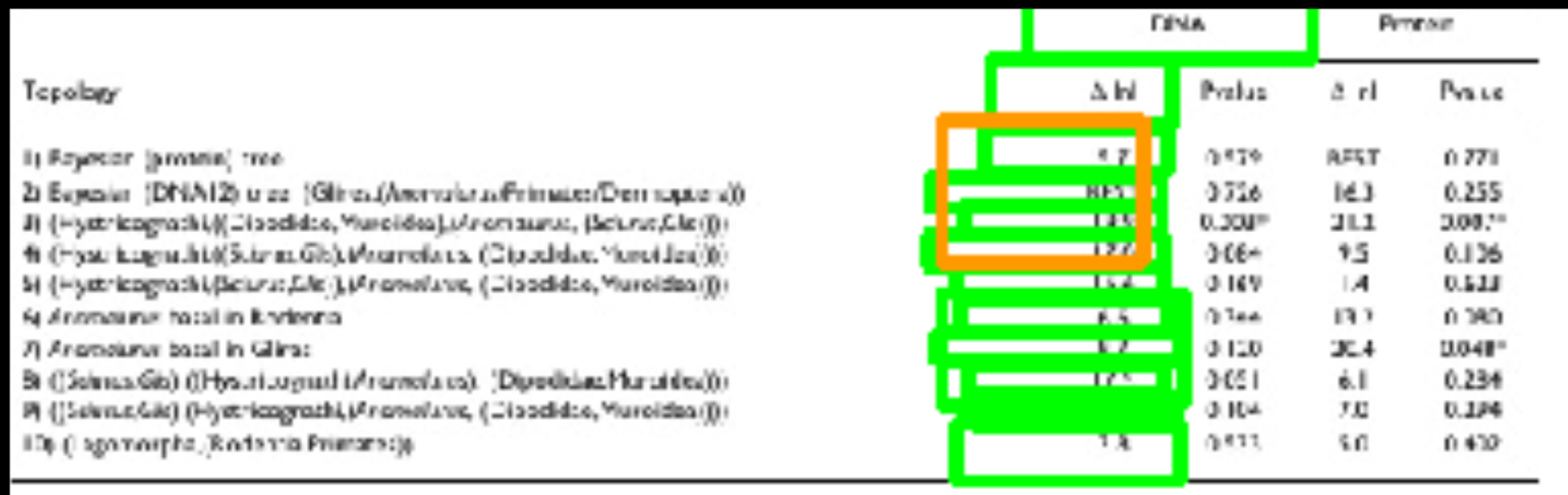}}
\end{center}
\caption{Sample structure recognition output of {\sc t}ab{\sc s}truct-{\sc n}et on table images of {\sc p}ub{\sc t}ab{\sc n}et dataset. \textbf{First Row:} prediction of cells which belong to the same row. \textbf{Second Row:} prediction of cells which belong to the same column. Cells marked with orange colour represent the examine cells and cells marked with green colour represent those which belong to the same row/column of the examined cell.}
\label{fig_pubtabnet_structure}
\end{figure}

\begin{figure}
\begin{center}
\fbox{
\includegraphics[width=0.29\linewidth, height=0.2\linewidth]{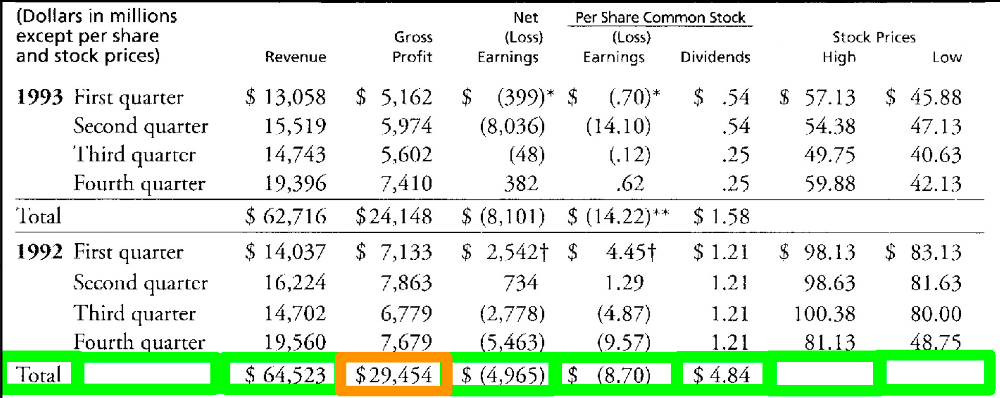}}
\hspace{-0.01\textwidth}
\fbox{
\includegraphics[width=0.29\linewidth, height=0.2\linewidth]{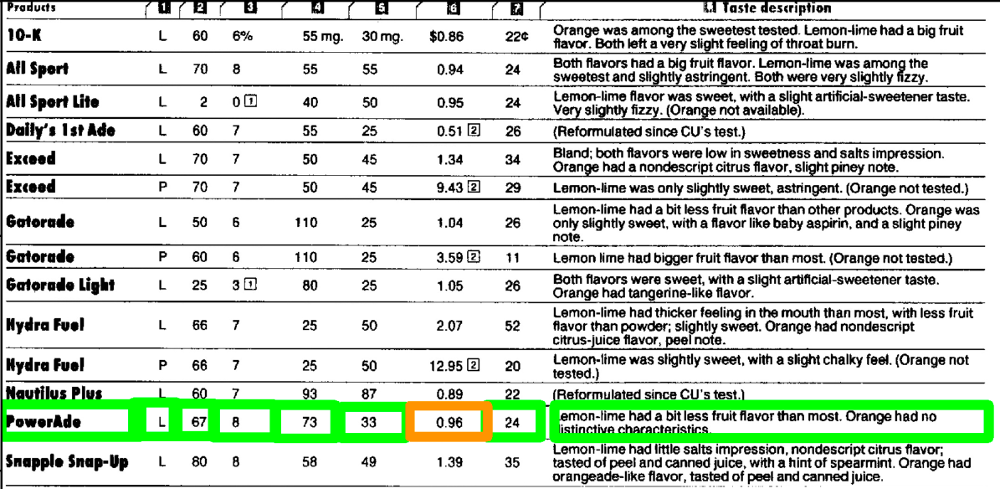}}
\hspace{-0.01\textwidth}
\fbox{
\includegraphics[width=0.29\linewidth, height=0.2\linewidth]{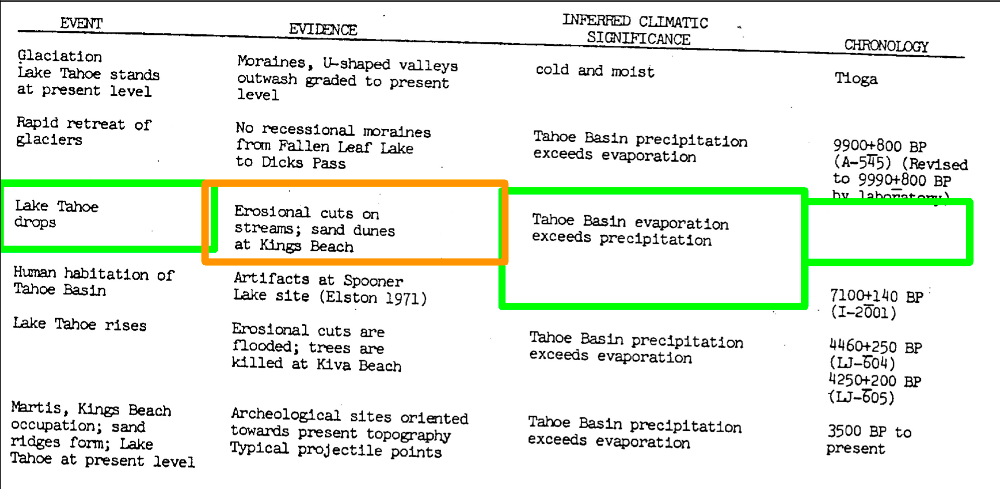}}
\vspace{0.001\textwidth}
\fbox{
\includegraphics[width=0.29\linewidth, height=0.2\linewidth]{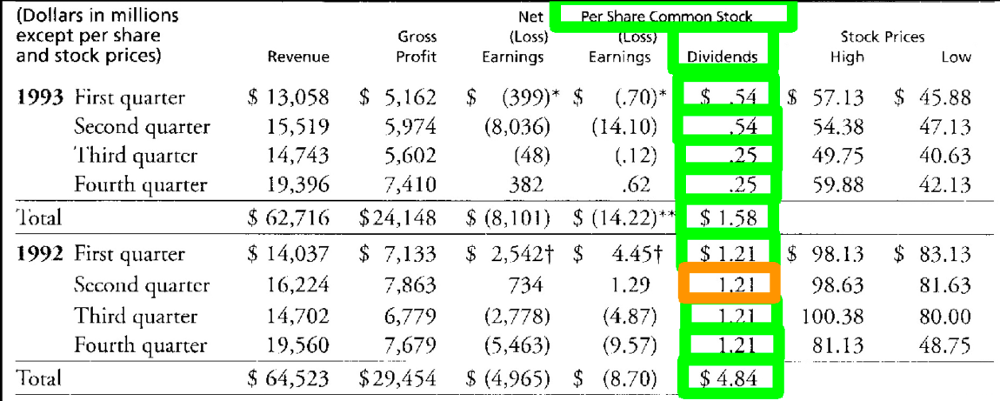}}
\hspace{-0.01\textwidth}
\fbox{
\includegraphics[width=0.29\linewidth, height=0.2\linewidth]{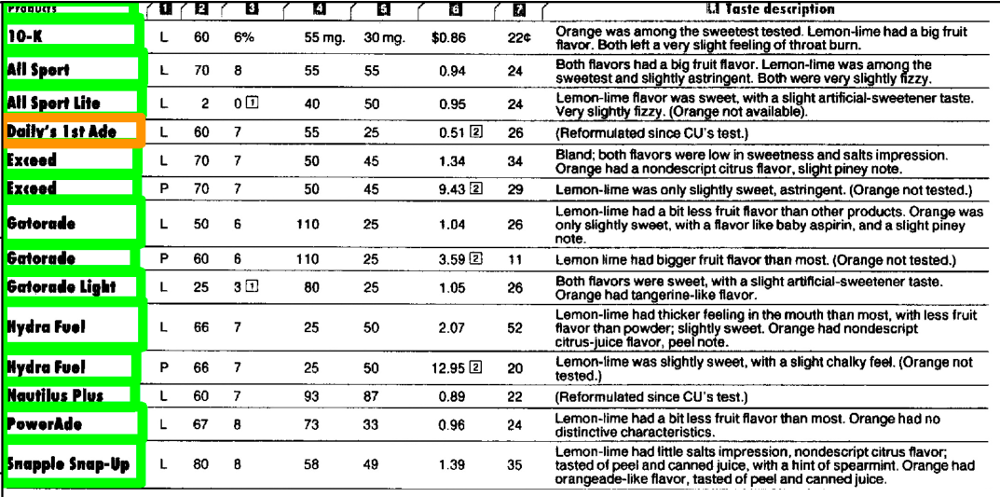}}
\hspace{-0.01\textwidth}
\fbox{
\includegraphics[width=0.29\linewidth, height=0.2\linewidth]{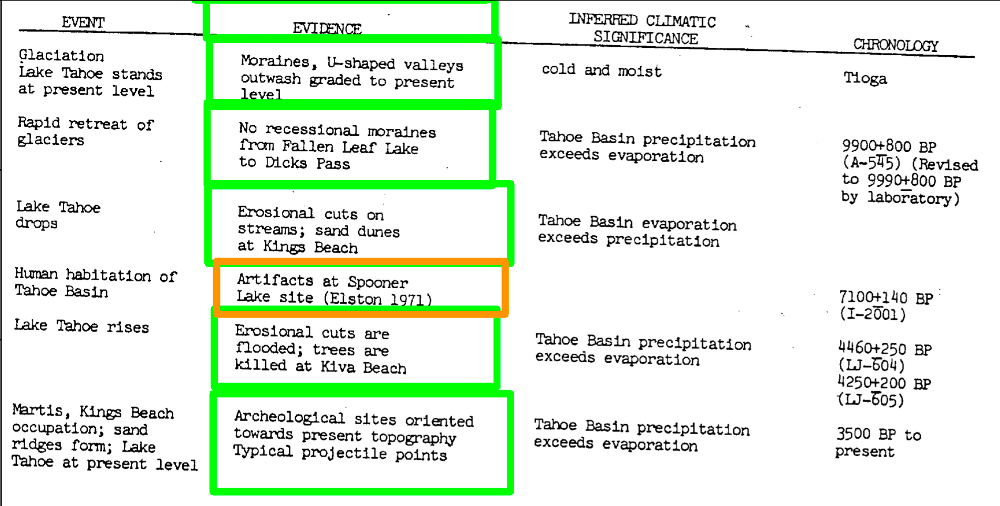}}
\end{center}
\caption{Sample structure recognition output of {\sc t}ab{\sc s}truct-{\sc n}et on table images of {\sc unlv} dataset. \textbf{First Row:} prediction of cells which belong to the same row. \textbf{Second Row:} prediction of cells which belong to the same column. Cells marked with orange colour represent the examine cells and cells marked with green colour represent those which belong to the same row/column of the examined cell.}
\label{fig_unlv_structure}
\end{figure}

Figures~\ref{fig_icdar_2013_structure}-\ref{fig_unlv_structure} demonstrate some qualitative results of structure recognition on all the evaluation datasets. From the figures, it can be seen that our model is able to work in the presence of archival table images, multi row/column spanning cells, varied table layouts and multiple line spanning cells. This indicates the robustness of our method under multiple kind of table images.

%%%%%%%%%%%%%%%%%%%%% ICDAR-2019 %%%%%%%%%%%%%%%%%%%%%%%%%%%%%%%%%%
%%%%%%%%%%%%%%%%%%%%%%% SciTSR %%%%%%%%%%%%%%%%%%%%%%%%%%%%%%%%
%%%%%%%%%%%%%%%%%%%%% SciTSR-COMP %%%%%%%%%%%%%%%%%%%%%%%%%%%%%%%%%%
%%%%%%%%%%%%%%%%%%%% TableBank %%%%%%%%%%%%%%%%%%%%%%%%%%%%%%%%%%%
%%%%%%%%%%%%%%%%%%%%% PubTabNet %%%%%%%%%%%%%%%%%%%%%%%%%
%%%%%%%%%%%%%%%%%%%%%%% UNLV %%%%%%%%%%%%%%%%%%%%%%%%%%%%%%%%

%%%%%%%%%%%%%%%%%%%%%%%%%%%%%%%%%%%%%%%%%%%%%%%%%%%
\newpage
\subsection{Failure Examples}
\begin{figure}
\begin{center}
\fbox{
\includegraphics[width=0.29\linewidth, height=0.2\linewidth]{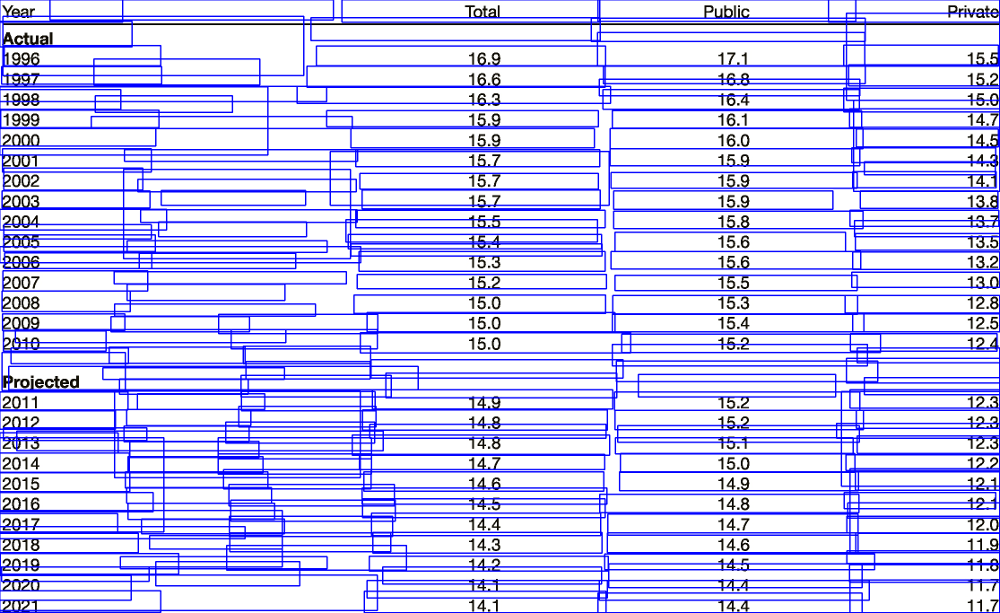}}
\hspace{-0.01\textwidth}
\fbox{
\includegraphics[width=0.29\linewidth, height=0.2\linewidth]{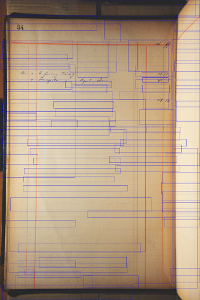}}
\hspace{-0.01\textwidth}
\fbox{
\includegraphics[width=0.29\linewidth, height=0.2\linewidth]{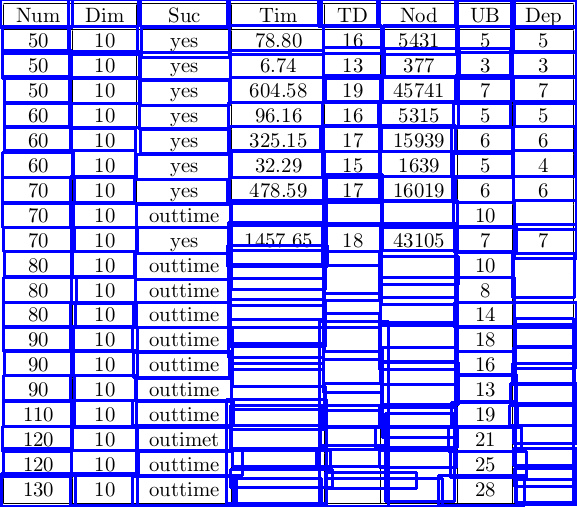}}
\vspace{0.001\textwidth}
\fbox{
\includegraphics[width=0.29\linewidth, height=0.2\linewidth]{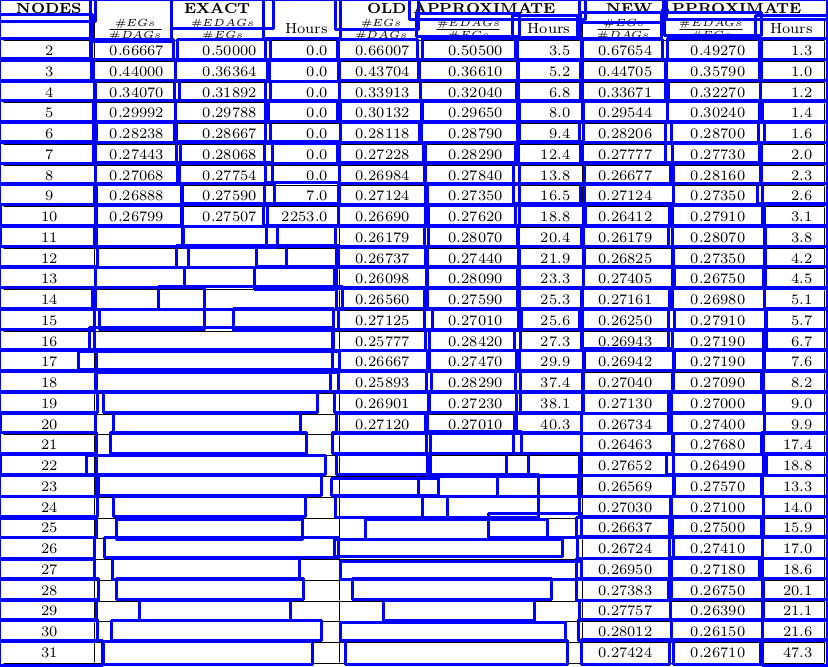}}
\hspace{-0.01\textwidth}
\fbox{
\includegraphics[width=0.29\linewidth, height=0.2\linewidth]{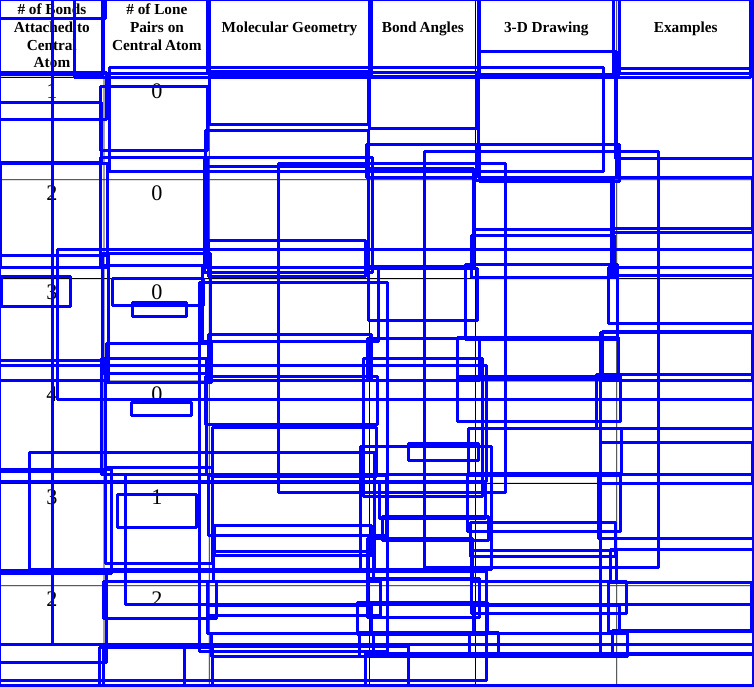}}
\hspace{-0.01\textwidth}
\fbox{
\includegraphics[width=0.29\linewidth, height=0.2\linewidth]{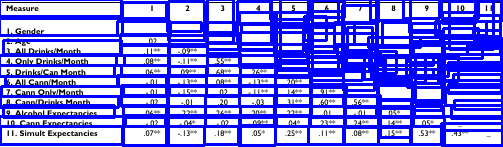}}
\vspace{0.001\textwidth}
\fbox{
\includegraphics[width=0.29\linewidth, height=0.2\linewidth]{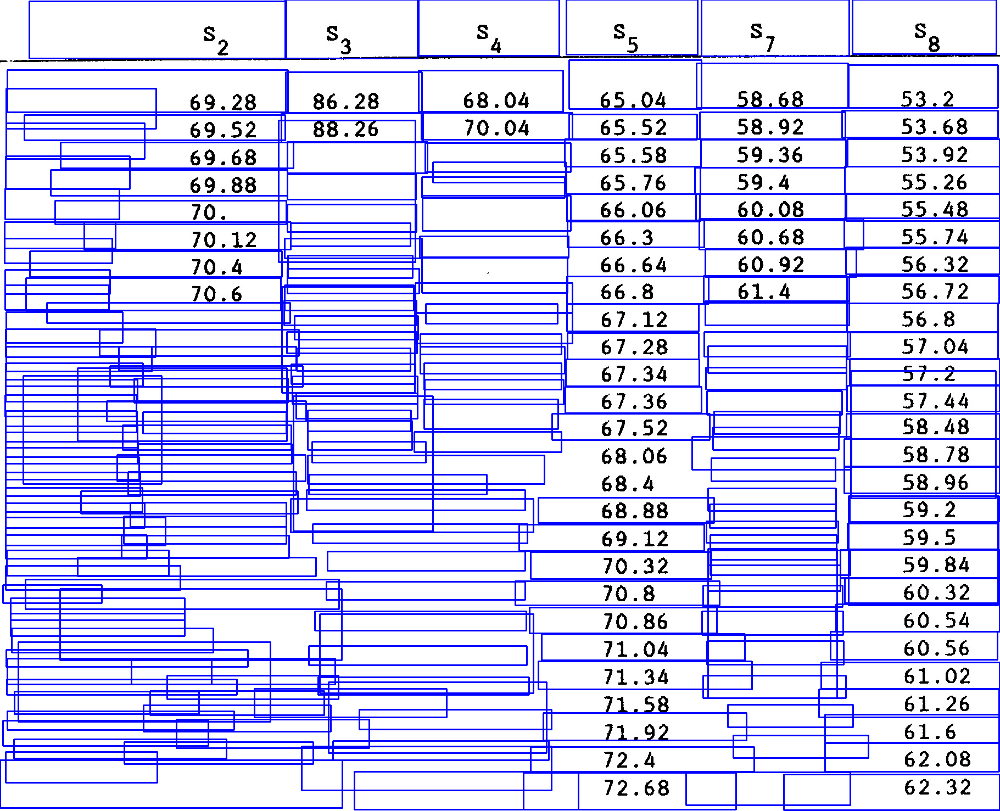}}
\end{center}
\caption{Sample intermediate cell detection results of {\sc t}ab{\sc s}truct-{\sc n}et on table images of {\sc icdar}-2013, {\sc icdar}-2019 c{\sc td}a{\sc r},  {\sc s}ci{\sc tsr}, {\sc s}ci{\sc tsr-comp}, {\sc t}able{\sc b}ank, {\sc p}ub{\sc t}ab{\sc n}et and {\sc unlv} datasets illustrate failure of {\sc t}ab{\sc s}truct-{\sc n}et.} \label{fig_failure_cell}
\end{figure}

Figure~\ref{fig_failure_cell} shows some failure cases of our model in presence of empty spaces along both horizontal and vertical axes.
%%%%%%%%%%%%%%%%%%%%%%%%%%%%%%%%%%%%%%%%%%%%%%%%%%%
%%%%%%%%%%%%%%%%%%%%%%%%%%%%%%%%%%%%%%%%%%%%%%%%%
\newpage
\subsection{Robustness of TabStruct-Net}

%%%%%%%%%%%%%%%%%%%%%%%%%%%%%%%%%%%%%%%%%%%%%
%robustness of method using varying IoU on icdar-2013-partial dataset
\begin{table}
\addtolength{\tabcolsep}{-1.0pt}
\begin{center}
\begin{tabular}{|l |l| l| l l l| l l l|} \hline
\textbf{CD Network} & \textbf{SR Network} &\textbf{IoU} &\multicolumn{3}{l|}{\textbf{CD Scores}} &\multicolumn{3}{l|}{\textbf{SR Scores}} \\ \cline{4-9}
 &  &\textbf{TH} &\textbf{P}$\uparrow$ &\textbf{R}$\uparrow$ &\textbf{F1}$\uparrow$ &\textbf{P}$\uparrow$ &\textbf{R}$\uparrow$ &\textbf{F1}$\uparrow$ \\ \hline
 &  &0.5	&\textbf{0.942}	&\textbf{0.948} &\textbf{0.945} &\textbf{0.933} &\textbf{0.915} &\textbf{0.924} \\  
 &      &0.6 &0.937	&0.941 &0.939 &0.930 &0.908	&0.919 \\
Mask {\sc r-cnn}+{\sc td}+{\sc bu}+{\sc al} &{\sc dgcnn}+{\sc p}2+{\sc lstm}  &0.7 &0.828 &0.831 &0.829 &0.800 &0.791 &0.795  \\
 &      &0.8 &0.651	&0.670 &0.660 &0.638 &0.624	&0.631 \\
 &      &0.9 &0.314	&0.336 &0.325 &0.291 &0.284	&0.287 \\ \hline
 \end{tabular}
\end{center}
\caption{Physical structure recognition results on {\sc icdar}-2013-partial dataset for varying IoU thresholds to demonstrate {\sc t}ab{\sc s}truct-{\sc n}et's robustness. \textbf{ES:} indicates Experimental Setup, \textbf{CD:} indicates Cell Detection, \textbf{TH:} indicates IoU threshold value, \textbf{SR:} indicates Structure Recognition, \textbf{P2:} indicates using visual features from P2 layer of the {\sc fpn} instead of using separate convolution blocks, \textbf{{\sc lstm}:} indicates use of {\sc lstm}s to model visual features along center-horizontal and center-vertical lines for every cell, \textbf{{\sc td}+{\sc bu}:} indicates use of Top-Down and Bottom-Up pathways in the {\sc fpn}, \textbf{AL:} indicates addition of alignment loss as a regularizer to {\sc t}ab{\sc s}truct-{\sc n}et, \textbf{P:} indicates precision, \textbf{R:} indicates recall, \textbf{F1:} indicates F1 Score. \label{table_varying_ious_physical_icdar_2013-partial}}
\vspace{-1em}
\end{table}

%%%%%%%%%%%%%%%%%%%%%%%%%%%%%%%%%%%%%%%%%%%%%
%robustness of method using varying IoU on icdar-2019 dataset
\begin{table}
\addtolength{\tabcolsep}{-1.0pt}
\begin{center}
\begin{tabular}{|l |l| l| l l l| l l l|} \hline
\textbf{CD Network} & \textbf{SR Network} &\textbf{IoU} &\multicolumn{3}{l|}{\textbf{CD Scores}} &\multicolumn{3}{l|}{\textbf{SR Scores}} \\ \cline{4-9}
 &  &\textbf{TH} &\textbf{P}$\uparrow$ &\textbf{R}$\uparrow$ &\textbf{F1}$\uparrow$ &\textbf{P}$\uparrow$ &\textbf{R}$\uparrow$ &\textbf{F1}$\uparrow$ \\ \hline
 &  &0.5	&\textbf{0.865} &\textbf{0.857} &\textbf{0.861} &\textbf{0.864} &\textbf{0.842} &\textbf{0.853} \\  
 &      &0.6 &0.84	&0.836	&0.838	&0.822	&0.787	&0.804 \\
Mask {\sc r-cnn}+{\sc td}+{\sc bu}+{\sc al} &{\sc dgcnn}+{\sc p}2+{\sc lstm}  &0.7 &0.694 &0.681 &0.687 &0.641	&0.625 &0.633  \\
 &      &0.8 &0.454	&0.428 &0.441 &0.404 &0.376	&0.389 \\
 &      &0.9 &0.201	&0.153 &0.174 &0.175 &0.138	&0.154 \\ \hline
 \end{tabular}
\end{center}
\caption{Physical structure recognition results on {\sc icdar}-2019 dataset for varying IoU thresholds to demonstrate {\sc t}ab{\sc s}truct-{\sc n}et's robustness. \textbf{ES:} indicates Experimental Setup, \textbf{CD:} indicates Cell Detection, \textbf{TH:} indicates IoU threshold value, \textbf{SR:} indicates Structure Recognition, \textbf{P2:} indicates using visual features from P2 layer of the {\sc fpn} instead of using separate convolution blocks, \textbf{{\sc lstm}:} indicates use of {\sc lstm}s to model visual features along center-horizontal and center-vertical lines for every cell, \textbf{{\sc td}+{\sc bu}:} indicates use of Top-Down and Bottom-Up pathways in the {\sc fpn}, \textbf{AL:} indicates addition of alignment loss as a regularizer to {\sc t}ab{\sc s}truct-{\sc n}et, \textbf{P:} indicates precision, \textbf{R:} indicates recall, \textbf{F1:} indicates F1 Score. \label{table_varying_ious_physical_icdar_2019}}
\vspace{-1em}
\end{table}

%%%%%%%%%%%%%%%%%%%%%%%%%%%%%%%%%%%%%%%%%%%%%
%robustness of method using varying IoU on UNLV-partial dataset
\begin{table}
\addtolength{\tabcolsep}{-1.0pt}
\begin{center}
\begin{tabular}{|l |l| l| l l l| l l l|} \hline
\textbf{CD Network} & \textbf{SR Network} &\textbf{IoU} &\multicolumn{3}{l|}{\textbf{CD Scores}} &\multicolumn{3}{l|}{\textbf{SR Scores}} \\ \cline{4-9}
 &  &\textbf{TH} &\textbf{P}$\uparrow$ &\textbf{R}$\uparrow$ &\textbf{F1}$\uparrow$ &\textbf{P}$\uparrow$ &\textbf{R}$\uparrow$ &\textbf{F1}$\uparrow$ \\ \hline
 &  &0.5	&\textbf{0.871}	&\textbf{0.879}	&\textbf{0.875}	&\textbf{0.864}	&\textbf{0.842}	&\textbf{0.853} \\  
 &      &0.6 &0.858	&0.864	&0.861	&0.849	&0.828	&0.839 \\
Mask {\sc r-cnn}+{\sc td}+{\sc bu}+{\sc al} &{\sc dgcnn}+{\sc p}2+{\sc lstm}  &0.7 &0.751 &0.773 &0.762	&0.735 &0.711 &0.723 \\
 &      &0.8 &0.595	&0.622 &0.608 &0.558 &0.532	&0.545 \\
 &      &0.9 &0.214	&0.237 &0.225 &0.173 &0.148	&0.160 \\ \hline
 \end{tabular}
\end{center}
\caption{Physical structure recognition results on {\sc unlv}-partial dataset for varying IoU thresholds to demonstrate {\sc t}ab{\sc s}truct-{\sc n}et's robustness. \textbf{ES:} indicates Experimental Setup, \textbf{CD:} indicates Cell Detection, \textbf{TH:} indicates IoU threshold value, \textbf{SR:} indicates Structure Recognition, \textbf{P2:} indicates using visual features from P2 layer of the {\sc fpn} instead of using separate convolution blocks, \textbf{{\sc lstm}:} indicates use of {\sc lstm}s to model visual features along center-horizontal and center-vertical lines for every cell, \textbf{{\sc td}+{\sc bu}:} indicates use of Top-Down and Bottom-Up pathways in the {\sc fpn}, \textbf{AL:} indicates addition of alignment loss as a regularizer to {\sc t}ab{\sc s}truct-{\sc n}et, \textbf{P:} indicates precision, \textbf{R:} indicates recall, \textbf{F1:} indicates F1 Score. \label{table_varying_ious_physical_unlv-partial}}
\vspace{-1em}
\end{table}

%%%%%%%%%%%%%%%%%%%%%%%%%%%%%%%%%%%%%%%%%%%%%
%robustness of method using varying IoU on SciTSR dataset
\begin{table}
\addtolength{\tabcolsep}{-1.0pt}
\begin{center}
\begin{tabular}{|l |l| l| l l l| l l l|} \hline
\textbf{CD Network} & \textbf{SR Network} &\textbf{IoU} &\multicolumn{3}{l|}{\textbf{CD Scores}} &\multicolumn{3}{l|}{\textbf{SR Scores}} \\ \cline{4-9}
 &  &\textbf{TH} &\textbf{P}$\uparrow$ &\textbf{R}$\uparrow$ &\textbf{F1}$\uparrow$ &\textbf{P}$\uparrow$ &\textbf{R}$\uparrow$ &\textbf{F1}$\uparrow$ \\ \hline
 &  &0.5	&\textbf{0.939} &\textbf{0.944} &\textbf{0.941} &\textbf{0.930} &\textbf{0.922} &\textbf{0.926} \\  
 &      &0.6 &0.932	&0.938	&0.935	&0.927	&0.913	&0.920 \\
Mask {\sc r-cnn}+{\sc td}+{\sc bu}+{\sc al} &{\sc dgcnn}+{\sc p}2+{\sc lstm}  &0.7 &0.808 &0.820 &0.814 &0.793 &0.775 &0.784 \\
 &      &0.8 &0.639	&0.652 &0.645 &0.618 &0.594	&0.606 \\
 &      &0.9 &0.297	&0.324 &0.310 &0.271 &0.258	&0.264 \\ \hline
 \end{tabular}
\end{center}
\caption{Physical structure recognition results on {\sc s}ci{\sc tsr} dataset for varying IoU thresholds to demonstrate {\sc t}ab{\sc s}truct-{\sc n}et's robustness. \textbf{ES:} indicates Experimental Setup, \textbf{CD:} indicates Cell Detection, \textbf{TH:} indicates IoU threshold value, \textbf{SR:} indicates Structure Recognition, \textbf{P2:} indicates using visual features from P2 layer of the {\sc fpn} instead of using separate convolution blocks, \textbf{{\sc lstm}:} indicates use of {\sc lstm}s to model visual features along center-horizontal and center-vertical lines for every cell, \textbf{{\sc td}+{\sc bu}:} indicates use of Top-Down and Bottom-Up pathways in the {\sc fpn}, \textbf{AL:} indicates addition of alignment loss as a regularizer to {\sc t}ab{\sc s}truct-{\sc n}et, \textbf{P:} indicates precision, \textbf{R:} indicates recall, \textbf{F1:} indicates F1 Score. \label{table_varying_ious_physical_scitsr}}
\vspace{-1em}
\end{table}

%%%%%%%%%%%%%%%%%%%%%%%%%%%%%%%%%%%%%%%%
\newpage
\subsection{Ablation Study}

%%%%%%%%%%%%%%%%%%%%%%%%%%%%%%%%%%%%%%%%%%%%%
%Ablation Study results on icdar-2013-partial dataset
\begin{table}
\addtolength{\tabcolsep}{-1.0pt}
\begin{center}
\begin{tabular}{|l|l |l |l l l| l l l|} \hline
\textbf{ES} &\textbf{CD Network} & \textbf{SR Network} &\multicolumn{3}{l|}{\textbf{CD Scores}} &\multicolumn{3}{l|}{\textbf{SR Scores}} \\ \cline{4-9}
 &  &   &\textbf{P}$\uparrow$ &\textbf{R}$\uparrow$ &\textbf{F1}$\uparrow$ &\textbf{P}$\uparrow$ &\textbf{R}$\uparrow$ &\textbf{F1}$\uparrow$ \\ \hline
&Mask {\sc r-cnn} &{\sc dgcnn} &0.835 &0.843 &0.839	&0.885 &0.864 &0.874 \\
&Mask {\sc r-cnn} &{\sc dgcnn}+P2 &0.837 &0.846	&0.841 &0.887 &0.865 &0.876 \\
&Mask {\sc r-cnn} &{\sc dgcnn}+P2+{\sc lstm} &0.840	&0.848	&0.844	&0.903	&0.889	&0.896 \\ \cline{2-9}
&Mask {\sc r-cnn}+{\sc td}+{\sc bu} &{\sc dgcnn} &0.898	&0.900	&0.899	&0.889	&0.882	&0.885 \\
S-A &Mask {\sc r-cnn}+{\sc td}+{\sc bu} &{\sc dgcnn}+{\sc p}2 &0.902 &0.905	&0.903 &0.893 &0.885 &0.889 \\ 
&Mask {\sc r-cnn}+{\sc td}+{\sc bu}	&{\sc dgcnn}+{\sc p}2+{\sc lstm} &0.924	&0.928	&0.926	&0.911	&0.896	&0.903 \\ \cline{2-9}
&Mask {\sc r-cnn}+{\sc td}+{\sc bu}+{\sc al} &{\sc dgcnn} &0.920 &0.924	&0.922 &0.915 &0.892 &0.903 \\	
&Mask {\sc r-cnn}+{\sc td}+{\sc bu}+{\sc al} &{\sc dgcnn}+{\sc p}2 &0.924 &0.927 &0.925	&0.918 &0.894 &0.906 \\
&Mask {\sc r-cnn}+{\sc td}+{\sc bu}+{\sc al} &{\sc dgcnn}+{\sc p}2+{\sc lstm} &\textbf{0.937} &\textbf{0.941}	&\textbf{0.939} &\textbf{0.930} &\textbf{0.908} &\textbf{0.919} \\ \hline
 & -{\sc na}- &{\sc dgcnn} & -{\sc na}- & -{\sc na}-  & -{\sc na}- &0.986 &0.990 &0.988 \\ 
S-B & -{\sc na}- &{\sc dgcnn}+{\sc p}2  & -{\sc na}- & -{\sc na}- & -{\sc na}- &0.987 &0.990 &0.989 \\
  & -{\sc na}-  &{\sc dgcnn}+{\sc p}2+{\sc lstm} & -{\sc na}- & -{\sc na}- & -{\sc na}- &\textbf{0.991} &\textbf{0.993} &\textbf{0.992} \\ \hline 
\end{tabular}
\end{center}
\caption{Ablation study for physical structure recognition on {\sc icdar}-2013-partial dataset. \textbf{ES:} indicates Experimental Setup, \textbf{CD:} indicates Cell Detection, \textbf{SR:} indicates Structure Recognition, \textbf{P2:} indicates using visual features from P2 layer of the {\sc fpn} instead of using separate convolution blocks, \textbf{{\sc lstm}:} indicates use of {\sc lstm}s to model visual features along center-horizontal and center-vertical lines for every cell, \textbf{{\sc td}+{\sc bu}:} indicates use of Top-Down and Bottom-Up pathways in the {\sc fpn}, \textbf{AL:} indicates addition of alignment loss as a regularizer to {\sc t}ab{\sc s}truct-{\sc n}et, \textbf{P:} indicates precision, \textbf{R:} indicates recall, \textbf{F1:} indicates F1 Score. \label{table_ablation_study_physical_icdar_2013_partial}}
\vspace{-1em}
\end{table}

%%%%%%%%%%%%%%%%%%%%%%%%%%%%%%%%%%%%%%%%%%%%%
%Ablation Study results on ICDAR-2019 dataset
\begin{table}%[ht!]
\addtolength{\tabcolsep}{-1.0pt}
\begin{center}
\begin{tabular}{|l|l |l |l l l| l l l|} \hline
\textbf{ES} &\textbf{CD Network} & \textbf{SR Network} &\multicolumn{3}{l|}{\textbf{CD Scores}} &\multicolumn{3}{l|}{\textbf{SR Scores}} \\ \cline{4-9}
 &  &   &\textbf{P}$\uparrow$ &\textbf{R}$\uparrow$ &\textbf{F1}$\uparrow$ &\textbf{P}$\uparrow$ &\textbf{R}$\uparrow$ &\textbf{F1}$\uparrow$ \\ \hline
&Mask {\sc r-cnn} &{\sc dgcnn} &0.770 &0.752 &0.761 &0.744 &0.706 &0.725 \\
&Mask {\sc r-cnn} &{\sc dgcnn}+P2 &0.774 &0.761	&0.767 &0.751 &0.718 &0.734 \\
&Mask {\sc r-cnn} &{\sc dgcnn}+P2+{\sc lstm} &0.797	&0.785 &0.791 &0.775 &0.750 &0.762 \\ \cline{2-9}
&Mask {\sc r-cnn}+{\sc td}+{\sc bu} &{\sc dgcnn} &0.775	&0.761 &0.768 &0.751 &0.713	&0.732 \\
S-A &Mask {\sc r-cnn}+{\sc td}+{\sc bu} &{\sc dgcnn}+{\sc p}2 &0.781 &0.768	&0.774 &0.756 &0.721 &0.738 \\ 
&Mask {\sc r-cnn}+{\sc td}+{\sc bu}	&{\sc dgcnn}+{\sc p}2+{\sc lstm} &0.803	&0.790 &0.796 &0.782 &0.754 &0.768 \\ \cline{2-9}
&Mask {\sc r-cnn}+{\sc td}+{\sc bu}+{\sc al} &{\sc dgcnn} &0.821 &0.814	&0.817 &0.797 &0.748 &0.772\\	
&Mask {\sc r-cnn}+{\sc td}+{\sc bu}+{\sc al} &{\sc dgcnn}+{\sc p}2 &0.823 &0.818 &0.820 &0.800 &0.753	&0.776 \\
&Mask {\sc r-cnn}+{\sc td}+{\sc bu}+{\sc al} &{\sc dgcnn}+{\sc p}2+{\sc lstm} &\textbf{0.840} &\textbf{0.836}	&\textbf{0.838} &\textbf{0.822}	&\textbf{0.787} &\textbf{0.804} \\ \hline
 & -{\sc na}- &{\sc dgcnn} & -{\sc na}- & -{\sc na}-  & -{\sc na}- &0.904 &0.889 &0.896 \\ 
S-B & -{\sc na}- &{\sc dgcnn}+{\sc p}2  & -{\sc na}- & -{\sc na}- & -{\sc na}- &0.932 &0.921 &0.927 \\
  & -{\sc na}-  &{\sc dgcnn}+{\sc p}2+{\sc lstm} & -{\sc na}- & -{\sc na}- & -{\sc na}- &\textbf{0.975} &\textbf{0.958} &\textbf{0.966} \\ \hline 
\end{tabular}
\end{center}
\caption{Ablation study for physical structure recognition on {\sc icdar-2019} dataset. \textbf{ES:} indicates Experimental Setup, \textbf{CD:} indicates Cell Detection, \textbf{SR:} indicates Structure Recognition, \textbf{P2:} indicates using visual features from P2 layer of the {\sc fpn} instead of using separate convolution blocks, \textbf{{\sc lstm}:} indicates use of {\sc lstm}s to model visual features along center-horizontal and center-vertical lines for every cell, \textbf{{\sc td}+{\sc bu}:} indicates use of Top-Down and Bottom-Up pathways in the {\sc fpn}, \textbf{AL:} indicates addition of alignment loss as a regularizer to {\sc t}ab{\sc s}truct-{\sc n}et, \textbf{P:} indicates precision, \textbf{R:} indicates recall, \textbf{F1:} indicates F1 Score. \label{table_ablation_study_physical_icdar_2019}}
\vspace{-1em}
\end{table}

%%%%%%%%%%%%%%%%%%%%%%%%%%%%%%%%%%%%%%%%%%%%%
%Ablation Study results on UNLV-partial dataset
\begin{table}%[ht!]
\addtolength{\tabcolsep}{-1.0pt}
\begin{center}
\begin{tabular}{|l|l |l |l l l| l l l|} \hline
\textbf{ES} &\textbf{CD Network} & \textbf{SR Network} &\multicolumn{3}{l|}{\textbf{CD Scores}} &\multicolumn{3}{l|}{\textbf{SR Scores}} \\ \cline{4-9}
 &  &   &\textbf{P}$\uparrow$ &\textbf{R}$\uparrow$ &\textbf{F1}$\uparrow$ &\textbf{P}$\uparrow$ &\textbf{R}$\uparrow$ &\textbf{F1}$\uparrow$ \\ \hline
&Mask {\sc r-cnn} &{\sc dgcnn} &0.835 &0.843 &0.839	&0.795 &0.764 &0.779 \\
&Mask {\sc r-cnn} &{\sc dgcnn}+P2 &0.837 &0.846	&0.841 &0.812 &0.788 &0.800 \\
&Mask {\sc r-cnn} &{\sc dgcnn}+P2+{\sc lstm} &0.840	&0.848 &0.844 &0.838 &0.821	&0.829 \\ \cline{2-9}
&Mask {\sc r-cnn}+{\sc td}+{\sc bu} &{\sc dgcnn} &0.837	&0.845 &0.841 &0.797 &0.766	&0.781 \\
S-A &Mask {\sc r-cnn}+{\sc td}+{\sc bu} &{\sc dgcnn}+{\sc p}2 &0.840 &0.849	&0.844 &0.815 &0.790 &0.802 \\ 
&Mask {\sc r-cnn}+{\sc td}+{\sc bu}	&{\sc dgcnn}+{\sc p}2+{\sc lstm} &0.844	&0.851 &0.847 &0.841 &0.823	&0.832 \\ \cline{2-9}
&Mask {\sc r-cnn}+{\sc td}+{\sc bu}+{\sc al} &{\sc dgcnn} &0.847 &0.855	&0.851 &0.802 &0.775 &0.788 \\	
&Mask {\sc r-cnn}+{\sc td}+{\sc bu}+{\sc al} &{\sc dgcnn}+{\sc p}2 &0.853 &0.860 &0.856	&0.823 &0.797 &0.810 \\
&Mask {\sc r-cnn}+{\sc td}+{\sc bu}+{\sc al} &{\sc dgcnn}+{\sc p}2+{\sc lstm} &\textbf{0.858} &\textbf{0.864}	&\textbf{0.861}	&\textbf{0.849} &\textbf{0.828}	&\textbf{0.839} \\ \hline
 & -{\sc na}- &{\sc dgcnn} & -{\sc na}- & -{\sc na}-  & -{\sc na}- &0.921 &0.898 &0.909 \\ 
S-B & -{\sc na}- &{\sc dgcnn}+{\sc p}2  & -{\sc na}- & -{\sc na}- & -{\sc na}- &0.950 &0.935 &0.942 \\
  & -{\sc na}-  &{\sc dgcnn}+{\sc p}2+{\sc lstm} & -{\sc na}- & -{\sc na}- & -{\sc na}- &\textbf{0.992} &\textbf{0.994}	&\textbf{0.993} \\ \hline 
\end{tabular}
\end{center}
\caption{Ablation study for physical structure recognition on {\sc unlv}-partial dataset. \textbf{ES:} indicates Experimental Setup, \textbf{CD:} indicates Cell Detection, \textbf{SR:} indicates Structure Recognition, \textbf{P2:} indicates using visual features from P2 layer of the {\sc fpn} instead of using separate convolution blocks, \textbf{{\sc lstm}:} indicates use of {\sc lstm}s to model visual features along center-horizontal and center-vertical lines for every cell, \textbf{{\sc td}+{\sc bu}:} indicates use of Top-Down and Bottom-Up pathways in the {\sc fpn}, \textbf{AL:} indicates addition of alignment loss as a regularizer to {\sc t}ab{\sc s}truct-{\sc n}et, \textbf{P:} indicates precision, \textbf{R:} indicates recall, \textbf{F1:} indicates F1 Score. \label{table_ablation_study_physical_unlv_partial}}
\vspace{-1em}
\end{table}

%%%%%%%%%%%%%%%%%%%%%%%%%%%%%%%%%%%%%%%%%%%%%
%Ablation Study results on SciTSR dataset
\begin{table}%[ht!]
\addtolength{\tabcolsep}{-1.0pt}
\begin{center}
\begin{tabular}{|l|l |l |l l l| l l l|} \hline
\textbf{ES} &\textbf{CD Network} & \textbf{SR Network} &\multicolumn{3}{l|}{\textbf{CD Scores}} &\multicolumn{3}{l|}{\textbf{SR Scores}} \\ \cline{4-9}
 &  &   &\textbf{P}$\uparrow$ &\textbf{R}$\uparrow$ &\textbf{F1}$\uparrow$ &\textbf{P}$\uparrow$ &\textbf{R}$\uparrow$ &\textbf{F1}$\uparrow$ \\ \hline
&Mask {\sc r-cnn} &{\sc dgcnn} &0.896 &0.900 &0.898	&0.888 &0.874 &0.881 \\
&Mask {\sc r-cnn} &{\sc dgcnn}+P2 &0.904 &0.907	&0.905 &0.892 &0.879 &0.885 \\
&Mask {\sc r-cnn} &{\sc dgcnn}+P2+{\sc lstm} &0.911	&0.915 &0.913 &0.903 &0.894	&0.898 \\ \cline{2-9}
&Mask {\sc r-cnn}+{\sc td}+{\sc bu} &{\sc dgcnn} &0.901	&0.909 &0.905 &0.893 &0.880	&0.886 \\
S-A &Mask {\sc r-cnn}+{\sc td}+{\sc bu} &{\sc dgcnn}+{\sc p}2 &0.905 &0.917 &0.911 &0.896 &0.882 &0.889 \\ 
&Mask {\sc r-cnn}+{\sc td}+{\sc bu}	&{\sc dgcnn}+{\sc p}2+{\sc lstm} &0.918	&0.924 &0.921 &0.905 &0.898	&0.902 \\ \cline{2-9}
&Mask {\sc r-cnn}+{\sc td}+{\sc bu}+{\sc al} &{\sc dgcnn} &0.908 &0.919	&0.913 &0.908 &0.894 &0.901 \\	
&Mask {\sc r-cnn}+{\sc td}+{\sc bu}+{\sc al} &{\sc dgcnn}+{\sc p}2 &0.921 &0.926 &0.923	&0.913 &0.901 &0.907 \\
&Mask {\sc r-cnn}+{\sc td}+{\sc bu}+{\sc al} &{\sc dgcnn}+{\sc p}2+{\sc lstm} &\textbf{0.932} &\textbf{0.938}	&\textbf{0.935}	&\textbf{0.927}	&\textbf{0.913}	&\textbf{0.920} \\ \hline
 & -{\sc na}- &{\sc dgcnn} & -{\sc na}- & -{\sc na}-  & -{\sc na}- &0.970 &0.981 &0.976 \\ 
S-B & -{\sc na}- &{\sc dgcnn}+{\sc p}2  & -{\sc na}- & -{\sc na}- & -{\sc na}- &0.973 &0.986 &0.979 \\
  & -{\sc na}-  &{\sc dgcnn}+{\sc p}2+{\sc lstm} & -{\sc na}- & -{\sc na}- & -{\sc na}- &\textbf{0.989}	&\textbf{0.993}	&\textbf{0.991} \\ \hline 
\end{tabular}
\end{center}
\caption{Ablation study for physical structure recognition on {\sc s}ci{\sc tsr} dataset. \textbf{ES:} indicates Experimental Setup, \textbf{CD:} indicates Cell Detection, \textbf{SR:} indicates Structure Recognition, \textbf{P2:} indicates using visual features from P2 layer of the {\sc fpn} instead of using separate convolution blocks, \textbf{{\sc lstm}:} indicates use of {\sc lstm}s to model visual features along center-horizontal and center-vertical lines for every cell, \textbf{{\sc td}+{\sc bu}:} indicates use of Top-Down and Bottom-Up pathways in the {\sc fpn}, \textbf{AL:} indicates addition of alignment loss as a regularizer to {\sc t}ab{\sc s}truct-{\sc n}et, \textbf{P:} indicates precision, \textbf{R:} indicates recall, \textbf{F1:} indicates F1 Score. \label{table_ablation_study_physical_scitsr}}
\vspace{-1em}
\end{table}

\end{document}